%% file: main.tex
\documentclass{article}
\usepackage{graphicx} 
\usepackage{caption}
\usepackage{subcaption}
\usepackage{geometry}
\usepackage{blindtext}
\usepackage{amsmath}
\usepackage{amsfonts}
\usepackage{amssymb}
\usepackage{tokcycle}
\usepackage{xcolor}
\usepackage{wrapfig}
\usepackage{xifthen}
\usepackage{hyperref}
\usepackage{authblk}
\usepackage{algorithm2e}
\usepackage[numbers]{natbib}
\usepackage{tikz}
\usepackage{pgfplots}
\usetikzlibrary{positioning, decorations.pathreplacing,calligraphy, calc, shapes.geometric, intersections, arrows}
\usepgfplotslibrary{fillbetween}
\pgfplotsset{compat=1.18}

\SetKwComment{Comment}{/* }{ */}

\RestyleAlgo{ruled}

\title{The Role of Recurrency in Image Segmentation\\ for Noisy and Limited Sample Settings}
\author[1]{David Calhas}
\author[1]{João Marques}
\author[1]{Arlindo L. Oliveira}
\affil[1]{INESC-ID \\Instituto Superior Tecnico}
\date{}

\begin{document}

\maketitle

\begin{abstract}
    The biological brain has inspired multiple advances in machine learning. However, most state-of-the-art models in computer vision do not operate like the human brain, simply because they are not capable of changing or improving their decisions/outputs based on a deeper analysis. The brain is recurrent, while these models are not. It is therefore relevant to explore what would be the impact of adding recurrent mechanisms to existing state-of-the-art architectures and to answer the question of whether recurrency can improve existing architectures. To this end, we build on a feed-forward segmentation model and explore multiple types of recurrency for image segmentation. We explore self-organizing, relational, and memory retrieval types of recurrency that minimize a specific energy function. In our experiments, we tested these models on artificial and medical imaging data, while analyzing the impact of high levels of noise and few-shot learning settings. Our results do not validate our initial hypothesis that recurrent models should perform better in these settings, suggesting that these recurrent architectures, by themselves, are not sufficient to surpass state-of-the-art feed-forward versions and that additional work needs to be done on the topic.
\end{abstract}

\input{introduction/introduction}

\input{problem/problem}

\input{recurrent/recurrent}

\input{setting/setting}

\input{results/results}
\input{discussion/discussion}

\input{related_work/related_work}

\input{conclusions/conclusions}

\bibliographystyle{unsrtnat}
\bibliography{bibliography}

\clearpage
\newpage

\appendix


\input{results/noise/artificial}


\input{results/examples/artificial}

\input{results/noise/cad}

\input{results/examples/cad}

\end{document}

%% file: introduction/introduction.tex
\section{Introduction}\label{section:intro}

In \textit{semantic segmentation}, given an input, the corresponding segmentation mask associates a class to each pixel of the image. The U-Net architecture is widely used in image segmentation \cite{ronneberger2015u,ho2020denoising,zhou2018unet} and can obtain good predictions in many different settings. Although the U-Net architecture has some similarities to the known structure of the visual cortex \cite{serre2014hierarchical}, it is critically different in that it is unable to use feedback from higher levels to improve its prediction. This is also the case for many state-of-the-art models that do not have a \textit{feedback cycle} in their computational graph \cite{spoerer2020recurrent}. It is believed that the brain can be best described as a dynamical system, containing an internal state that represents its belief about the world \cite{flavell2022emergence}. When the stimuli are different from what is expected given its internal state, it processes the input and corrects its expectation \cite{rao1999predictive}.
\begin{figure}[t]
    \centering
    \input{introduction/figures/back_forward}
    \caption{This figure illustrates the fact that a system with feedback connections is able to change its decision with time. For that, at least one cycle is needed in the computational graph.}
    \label{fig:forward_vs_feedback}
\end{figure}
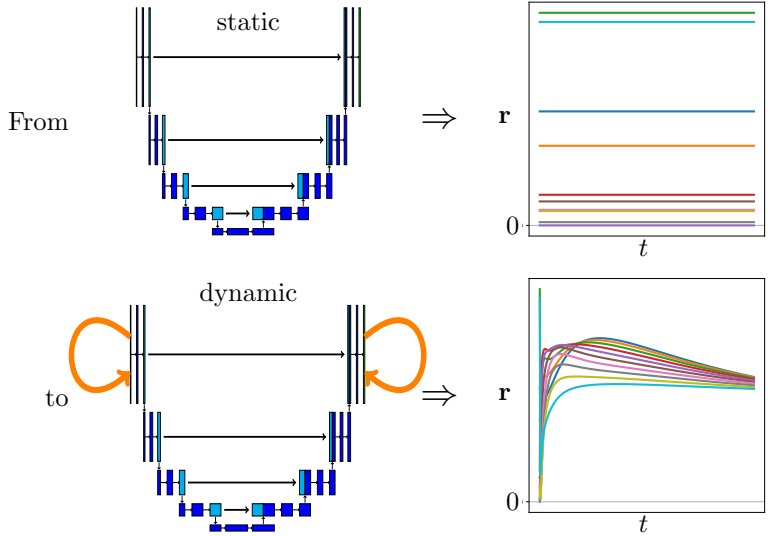
In this work, we study the hypothesis that there is a fundamental limitation on current machine learning models because they are not able to process their decisions and use intermediate decisions to move towards a more correct/certain one. 
To this end, we apply recurrency to a state-of-the-art segmentation model and explore its application on artificial and medical images. Our contributions are the following:
\begin{itemize}
    \item we describe how three recurrent types of paradigms used in this work (self-organizing maps, conditional random fields, and modern Hopfield networks, described in section \ref{section:recurrent_models}) can be added or appended at any level of the U-Net. Specifically, we use the EfficientUnet++ \cite{lourencco2021encoder} backbone as the base model, which is based on the U-Net++ architecture, with an EfficientNet \cite{tan2019efficientnet} backend;
    \item we propose and make available a new dataset with irregular polygon shapes (see section \ref{dataset:shapes}) and define two experimental settings, the noisy and limited sample size cases, to analyze feed-forward and recurrent models (see sections \ref{subsection:noise} and \ref{subsection:examples});
    \item using artificially generated data, we show that self-organizing recurrency is the best type of model for noisy settings (see section \ref{section:results}) and recurrency is preferred to feed forward in limited sample size settings (see sections \ref{section:results} and \ref{section:discussion});
    \item we present experimental results showing that none of the models used has significant success in a real world medical imaging dataset (see sections \ref{dataset:cad}, \ref{section:results},  and \ref{section:discussion}).
\end{itemize}

%% file: introduction/figures/back_forward.tex
\begin{tikzpicture}
    \node (center) at (0,0) {};

    \node (forward) [draw, rectangle, white, minimum width=100, minimum height=90] at ($(center)+(0,0)$) {};
    \node (backward) [anchor=north, draw,  white,rectangle, minimum width=100, minimum height=90] at ($(forward.south)+(0,-0.5)$) {};
    
    \node (from) [anchor=east] at ($(forward.west)+(-0.5,0)$) {From};
    
    \node (versus) [anchor=east] at ($(backward.west)+(-0.5,0)$) {to};

    \node (static) [anchor=south] at ($(forward.north)+(0.,-0.5)$) {static};
    \node (dynamic) [anchor=south] at ($(backward.north)+(0.,-0.5)$) {dynamic};

    \node (unet) at ($(forward)+(0,0)$) {\resizebox{0.2\textwidth}{0.2\textwidth}{\input{introduction/figures/unet}}};
    \node (unet_dynamic) at ($(backward)+(0,0)$) {\resizebox{0.35\textwidth}{0.235\textwidth}{\input{introduction/figures/unet_dynamic}}};
    
    \node (response_dynamic) at ($(backward.east)+(1.75,0.05)$) [anchor=west] {\includegraphics[width=0.24\textwidth, clip, trim={1.3cm 1.3cm 0cm 1.3cm}]{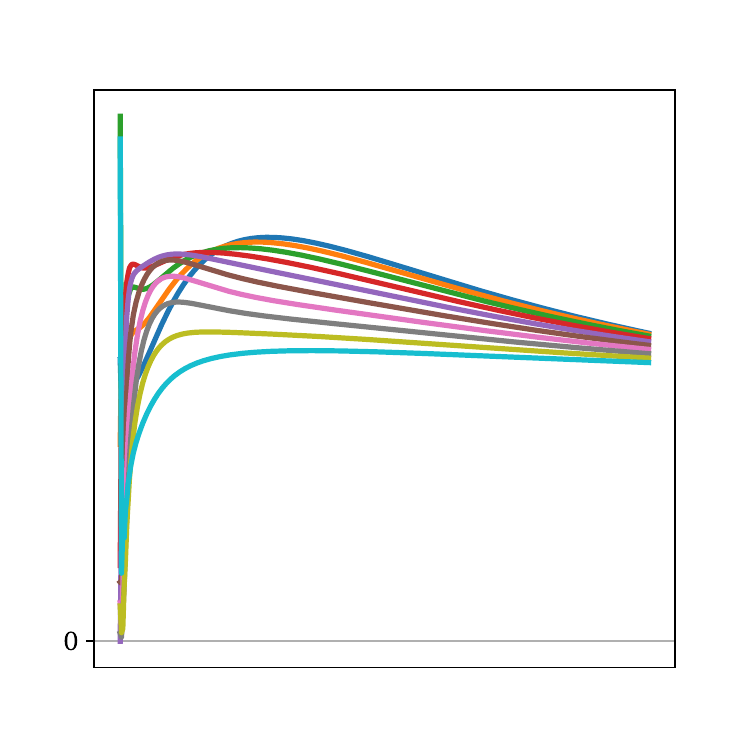}};

    \node (response_static) at ($(forward.east)+(1.75,0.05)$) [anchor=west] {\includegraphics[width=0.24\textwidth, clip, trim={1.3cm 1.3cm 0cm 1.3cm}]{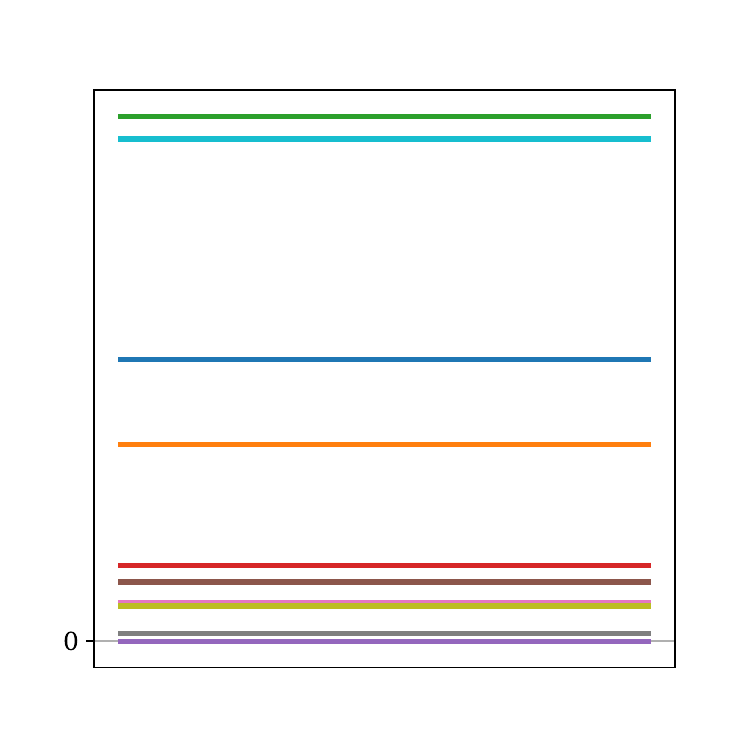}};

    \node (implies) at ($(backward.east)+(0.4,0)$) [anchor=west] {\Large{$\Rightarrow$}};
    \node (implies) at ($(forward.east)+(0.4,0)$) [anchor=west] {\Large{$\Rightarrow$}};

    \node (legend) at ($(response_dynamic.west)+(0.1,0)$) [anchor=east] {\normalsize{$\mathbf{r}$}};
    \node (legend) at ($(response_static.west)+(0.1,0)$) [anchor=east] {\normalsize{$\mathbf{r}$}};

    \node (legend) at ($(response_dynamic.south)+(-0.2,-0.0)$) [anchor=center] {\normalsize{$t$}};
    \node (legend) at ($(response_static.south)+(-0.2,-0.0)$) [anchor=center] {\normalsize{$t$}};

    \node (legend) at ($(response_dynamic.south west)+(-0.,0.3)$) [anchor=center] {\normalsize{$0$}};
    \node (legend) at ($(response_static.south west)+(-0.,0.3)$) [anchor=center] {\normalsize{$0$}};

\end{tikzpicture}

%% file: problem/problem.tex
\section{Problem description}

Let $\mathbf{X} = \{\mathbf{x}_1, \dots, \mathbf{x}_N\}$ be a set of images (two-dimensional representations) and $\mathbf{Y} = \{\mathbf{y}_1, \dots, \mathbf{y}_N\}$ the corresponding set of pixelwise labels, with $\mathbf{x}_i \in \mathbb{R}^{H \times W\times C} \wedge \mathbf{y}_i \in \mathbb{R}^{H \times W\times L}, \forall i \in \{1, \dots, N\}$, where $H$ and $W$ define the dimension of the image, $C$ the number of channels of the original representation ($C=3$ for RGB), and $L$ the number of classes. Our goal is to compute the parameters $\theta$ of a function $f$, such that
\begin{equation}
    \theta = \mbox{argmin}_\theta \frac{1}{N} \sum_i^N \mathcal{L}\left( f(\mathbf{x}_i; \theta), \mathbf{y}_i \right),
\end{equation}
where $\mathcal{L}$ is an objective function, e.g. cross-entropy.

%% file: recurrent/recurrent.tex
\section{Recurrent Models}\label{section:recurrent_models}

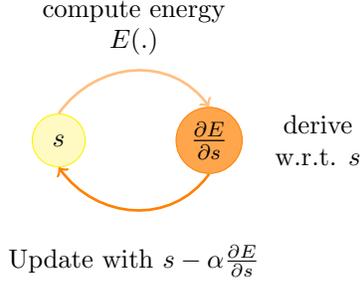
\begin{figure}[ht]
    \centering
    \input{recurrent/figures/recurrency}
    \caption{An illustration of recurrency, where the dynamics are governed by gradient descent, which implies the minimization of an energy function $E$ w.r.t. a state $s$.}
    \label{fig:recurrency}
\end{figure}

Gradient descent is a form of recurrency, where we updates are performed recurrently as illustrated in figure \ref{fig:recurrency}.

\noindent Consider a state $s$ that evolves in time according to the dynamics
\begin{equation}
    \tau \frac{ds}{dt}=- \frac{\partial E}{\partial s(t)},
\end{equation}
where the evolution is governed by an energy function $E$ that is being minimized. Then we can update a state $s$ with gradient descent as
\begin{equation}
    s(t+1) = s(t) - \alpha \frac{\partial E}{\partial s(t)} ,
\end{equation}
where $\alpha$ is the learning rate. The goal of this study is to explore multiple forms of energy-based recurrency and bridge its dynamics with human neuronal processing, such as self-organizing maps \cite{kohonen1982self}, conditional random fields \cite{zheng2015conditional}, and Hopfield networks \cite{hoover2024energy}.

\input{recurrent/som/som}

\input{recurrent/crf/crf}

\input{recurrent/hopfield/hopfield}

%% file: recurrent/figures/recurrency.tex
\begin{tikzpicture}
    \node (center) at (0,0) {};

    \node (state) [draw=yellow, circle, minimum size=20, fill=yellow!30, inner sep=0] at ($(center)+(0,0)$) {$s$};

    \node (gradient) [draw=orange, circle, minimum size=25, fill=orange!70, inner sep=0]  at ($(state)+(2,0)$) {\large{$\frac{\partial E}{\partial s}$}};

    \draw[->, line width=1, orange!50] ($(state.north)+(0,0)$) .. controls ($(state.north)+(0.5,0.7)$) and ($(gradient.north)+(-0.5,0.7)$) .. ($(gradient.north)+(0,0)$);

    \draw[->, line width=1, orange] ($(gradient.south)+(0,0)$) .. controls ($(gradient.south)+(-0.5,-0.7)$) and ($(state.south)+(0.5,-0.7)$) .. ($(state.south)+(0,0)$);

    \node (legend_compute) [align=center] at ($(state)+(1,1.5)$) {compute energy\\$E(.)$};
    \node (derive) [align=center] at ($(gradient.east)+(1,0)$) {derive\\w.r.t. $s$};
    \node (update) [align=center] at ($(state)+(1,-1.6)$) {Update with $s-\alpha\frac{\partial E}{\partial s}$};
\end{tikzpicture}

%% file: recurrent/som/som.tex
\subsection{Self-organizing map}\label{section:som}

Originally proposed by \citet{kohonen1982self}, a self-organizing map is a neural network architecture that learns the topology of a space and is inspired in the way some neuronal cells respond to orientation \cite{von1973self}. A self-organizing map is defined by a set of nodes, $\mathcal{V} \in \mathbb{R}^{M\times M \times L}$, that define a square lattice. The lattice can be rectangular if we define $\mathcal{V} \in \mathbb{R}^{H \times W \times L} with H \neq W$. Each node $\mathbf{v} \in \mathbb{R}^{L}$ is a feature vector to be optimized according to the algorithm. 
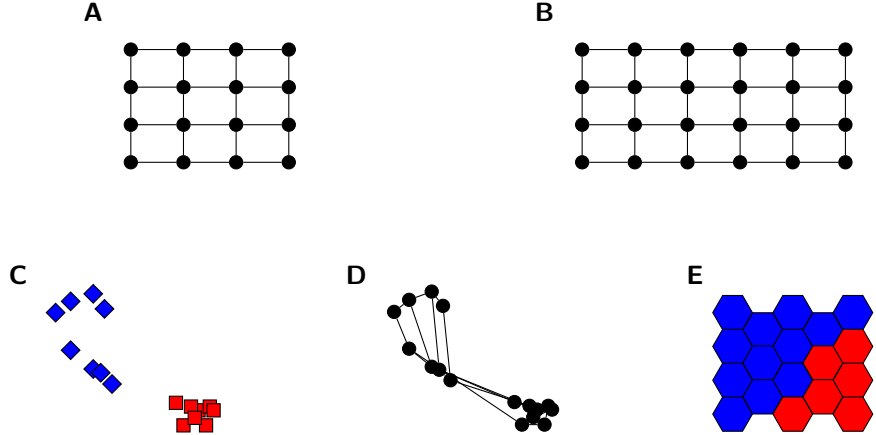
\begin{figure}[ht]
    \centering
    \input{recurrent/som/figures/som_grid}
    \caption{A self-organizing map square lattice.}
    \label{fig:som_grid}
\end{figure}
Given a point $\mathbf{x}$, we first compute the distance to all nodes and select the closest node, $\mathbf{v}^* = argmin_{\mathbf{v} \in \mathcal{V}} ||\mathbf{x}-\mathbf{v}||^2_2$. The algorithm proceeds to update the closest node as
\begin{equation}\label{equation:som}
    \mathbf{v}^* = \mathbf{v}^* + \alpha \left( \mathbf{x} -\mathbf{v}^* \right),
\end{equation}
where $\alpha$ is the learning rate. The same update, $\mathbf{v}' = \mathbf{v}' + \alpha \left( \mathbf{x} -\mathbf{v}' \right)$, is done for any node $\mathbf{v}'$ that is a neighbour of $\mathbf{v}^*$. For a more compact notation, let $\mathcal{N}: \mathbb{R}^2 \to \mathbb{R}^{5\times L}$ be the neighbor function of a node $\mathbf{v}$, that given its lattice coordinates $i$, returns the $4$ direct neighbors and $\mathbf{v}$ itself. For the sake of simplicity, we interchange between coordinates giving $i$ and the representation of the node $\mathbf{v}$ as input to $\mathcal{N}$, such that the neighbors $\mathbf{v}$ are $\mathcal{N}(i)=\mathcal{N}(\mathbf{v})$. Then the update rule for the set of neighbors is
\begin{equation}\label{equation:som_neighbours}
    \mathcal{N}(\mathbf{v}^*) = \mathcal{N}(\mathbf{v}^*) + \alpha \left( \mathbf{x} - \mathcal{N}(\mathbf{v}^*) \right) .
\end{equation}
Note that this imposes a competitive dynamic for the nodes when iterating through the points of the space we are learning from. Space regions with more density are going to have more nodes nearby, while regions with less density are going to have fewer nodes. Regions with no density will not have nodes.

\textbf{Self-organizing map energy.}  The energy of a map $\mathcal{V} \in \mathbb{R}^{M\times M \times C}$ given a set of instances $\mathbf{Y}$ is defined as
\begin{equation}\label{equation:som_energy_original}
    E^{\mbox{SOM}} = \frac{1}{M\times M} \sum_j^{M\times M} \frac{1}{5} \left(\sum \left( \mathcal{N}\left(\mbox{argmin}_{\mathbf{v}}\sum(\mathbf{v}-\mathbf{y})^2\right) - \mathbf{y}_j^\top \right)^2 \right),
\end{equation}
where $\mbox{argmin}_{\mathbf{v}}\sum(\mathbf{v}-\mathbf{y})^2$ determines $\mathbf{y}$, the best matching node and $\mathcal{N}\left(\mbox{argmin}_{\mathbf{v}}\sum(\mathbf{v}-\mathbf{y})^2\right) - \mathbf{y}_j^\top$ computes the distance matrix of $\mathbf{y}$ with the best matching node and its neighbors. The division by $5$ refers to the normalization of the $5$ nodes taken into account.

\input{recurrent/som/som_inspired}

%% file: recurrent/som/figures/som_grid.tex
\begin{tikzpicture}
    \node (center) at (0,0) {};

    \node (x1y1) [draw, circle, minimum size=5, inner sep=0., fill=black] at ($(center)+(0,0)$) {};
    \node (x2y1) [draw, circle, minimum size=5, inner sep=0., fill=black] at ($(x1y1)+(0.7,0)$) {};
    \node (x3y1) [draw, circle, minimum size=5, inner sep=0., fill=black] at ($(x2y1)+(0.7,0)$) {};
    \node (x4y1) [draw, circle, minimum size=5, inner sep=0., fill=black] at ($(x3y1)+(0.7,0)$) {};

    \node (x1y2) [draw, circle, minimum size=5, inner sep=0., fill=black] at ($(x1y1)+(0,-0.5)$) {};
    \node (x2y2) [draw, circle, minimum size=5, inner sep=0., fill=black] at ($(x1y2)+(0.7,0)$) {};
    \node (x3y2) [draw, circle, minimum size=5, inner sep=0., fill=black] at ($(x2y2)+(0.7,0)$) {};
    \node (x4y2) [draw, circle, minimum size=5, inner sep=0., fill=black] at ($(x3y2)+(0.7,0)$) {};

    \node (x1y3) [draw, circle, minimum size=5, inner sep=0., fill=black] at ($(x1y2)+(0,-0.5)$) {};
    \node (x2y3) [draw, circle, minimum size=5, inner sep=0., fill=black] at ($(x1y3)+(0.7,0)$) {};
    \node (x3y3) [draw, circle, minimum size=5, inner sep=0., fill=black] at ($(x2y3)+(0.7,0)$) {};
    \node (x4y3) [draw, circle, minimum size=5, inner sep=0., fill=black] at ($(x3y3)+(0.7,0)$) {};

    \node (x1y4) [draw, circle, minimum size=5, inner sep=0., fill=black] at ($(x1y3)+(0,-0.5)$) {};
    \node (x2y4) [draw, circle, minimum size=5, inner sep=0., fill=black] at ($(x1y4)+(0.7,0)$) {};
    \node (x3y4) [draw, circle, minimum size=5, inner sep=0., fill=black] at ($(x2y4)+(0.7,0)$) {};
    \node (x4y4) [draw, circle, minimum size=5, inner sep=0., fill=black] at ($(x3y4)+(0.7,0)$) {};

    \draw[black] (x1y1) -- (x2y1);
    \draw[black] (x2y1) -- (x3y1);
    \draw[black] (x3y1) -- (x4y1);

    \draw[black] (x1y2) -- (x2y2);
    \draw[black] (x2y2) -- (x3y2);
    \draw[black] (x3y2) -- (x4y2);

    \draw[black] (x1y3) -- (x2y3);
    \draw[black] (x2y3) -- (x3y3);
    \draw[black] (x3y3) -- (x4y3);

    \draw[black] (x1y4) -- (x2y4);
    \draw[black] (x2y4) -- (x3y4);
    \draw[black] (x3y4) -- (x4y4);

    \draw[black] (x1y1) -- (x1y2);
    \draw[black] (x1y2) -- (x1y3);
    \draw[black] (x1y3) -- (x1y4);

    \draw[black] (x2y1) -- (x2y2);
    \draw[black] (x2y2) -- (x2y3);
    \draw[black] (x2y3) -- (x2y4);

    \draw[black] (x3y1) -- (x3y2);
    \draw[black] (x3y2) -- (x3y3);
    \draw[black] (x3y3) -- (x3y4);

    \draw[black] (x4y1) -- (x4y2);
    \draw[black] (x4y2) -- (x4y3);
    \draw[black] (x4y3) -- (x4y4);

    \node (legend) at ($(center)+(-0.5, 0.5)$) {\sffamily{\textbf{A}}};

    \node (center) at (6,0) {};

    \node (x1y1) [draw, circle, minimum size=5, inner sep=0., fill=black] at ($(center)+(0,0)$) {};
    \node (x2y1) [draw, circle, minimum size=5, inner sep=0., fill=black] at ($(x1y1)+(0.7,0)$) {};
    \node (x3y1) [draw, circle, minimum size=5, inner sep=0., fill=black] at ($(x2y1)+(0.7,0)$) {};
    \node (x4y1) [draw, circle, minimum size=5, inner sep=0., fill=black] at ($(x3y1)+(0.7,0)$) {};
    \node (x5y1) [draw, circle, minimum size=5, inner sep=0., fill=black] at ($(x4y1)+(0.7,0)$) {};
    \node (x6y1) [draw, circle, minimum size=5, inner sep=0., fill=black] at ($(x5y1)+(0.7,0)$) {};

    \node (x1y2) [draw, circle, minimum size=5, inner sep=0., fill=black] at ($(x1y1)+(0,-0.5)$) {};
    \node (x2y2) [draw, circle, minimum size=5, inner sep=0., fill=black] at ($(x1y2)+(0.7,0)$) {};
    \node (x3y2) [draw, circle, minimum size=5, inner sep=0., fill=black] at ($(x2y2)+(0.7,0)$) {};
    \node (x4y2) [draw, circle, minimum size=5, inner sep=0., fill=black] at ($(x3y2)+(0.7,0)$) {};
    \node (x5y2) [draw, circle, minimum size=5, inner sep=0., fill=black] at ($(x4y2)+(0.7,0)$) {};
    \node (x6y2) [draw, circle, minimum size=5, inner sep=0., fill=black] at ($(x5y2)+(0.7,0)$) {};

    \node (x1y3) [draw, circle, minimum size=5, inner sep=0., fill=black] at ($(x1y2)+(0,-0.5)$) {};
    \node (x2y3) [draw, circle, minimum size=5, inner sep=0., fill=black] at ($(x1y3)+(0.7,0)$) {};
    \node (x3y3) [draw, circle, minimum size=5, inner sep=0., fill=black] at ($(x2y3)+(0.7,0)$) {};
    \node (x4y3) [draw, circle, minimum size=5, inner sep=0., fill=black] at ($(x3y3)+(0.7,0)$) {};
    \node (x5y3) [draw, circle, minimum size=5, inner sep=0., fill=black] at ($(x4y3)+(0.7,0)$) {};
    \node (x6y3) [draw, circle, minimum size=5, inner sep=0., fill=black] at ($(x5y3)+(0.7,0)$) {};

    \node (x1y4) [draw, circle, minimum size=5, inner sep=0., fill=black] at ($(x1y3)+(0,-0.5)$) {};
    \node (x2y4) [draw, circle, minimum size=5, inner sep=0., fill=black] at ($(x1y4)+(0.7,0)$) {};
    \node (x3y4) [draw, circle, minimum size=5, inner sep=0., fill=black] at ($(x2y4)+(0.7,0)$) {};
    \node (x4y4) [draw, circle, minimum size=5, inner sep=0., fill=black] at ($(x3y4)+(0.7,0)$) {};
    \node (x5y4) [draw, circle, minimum size=5, inner sep=0., fill=black] at ($(x4y4)+(0.7,0)$) {};
    \node (x6y4) [draw, circle, minimum size=5, inner sep=0., fill=black] at ($(x5y4)+(0.7,0)$) {};

    \draw[black] (x1y1) -- (x2y1);
    \draw[black] (x2y1) -- (x3y1);
    \draw[black] (x3y1) -- (x4y1);
    \draw[black] (x4y1) -- (x5y1);
    \draw[black] (x5y1) -- (x6y1);

    \draw[black] (x1y2) -- (x2y2);
    \draw[black] (x2y2) -- (x3y2);
    \draw[black] (x3y2) -- (x4y2);
    \draw[black] (x4y2) -- (x5y2);
    \draw[black] (x5y2) -- (x6y2);

    \draw[black] (x1y3) -- (x2y3);
    \draw[black] (x2y3) -- (x3y3);
    \draw[black] (x3y3) -- (x4y3);
    \draw[black] (x4y3) -- (x5y3);
    \draw[black] (x5y3) -- (x6y3);

    \draw[black] (x1y4) -- (x2y4);
    \draw[black] (x2y4) -- (x3y4);
    \draw[black] (x3y4) -- (x4y4);
    \draw[black] (x4y4) -- (x5y4);
    \draw[black] (x5y4) -- (x6y4);

    \draw[black] (x1y1) -- (x1y2);
    \draw[black] (x1y2) -- (x1y3);
    \draw[black] (x1y3) -- (x1y4);

    \draw[black] (x2y1) -- (x2y2);
    \draw[black] (x2y2) -- (x2y3);
    \draw[black] (x2y3) -- (x2y4);

    \draw[black] (x3y1) -- (x3y2);
    \draw[black] (x3y2) -- (x3y3);
    \draw[black] (x3y3) -- (x3y4);

    \draw[black] (x4y1) -- (x4y2);
    \draw[black] (x4y2) -- (x4y3);
    \draw[black] (x4y3) -- (x4y4);

    \draw[black] (x5y1) -- (x5y2);
    \draw[black] (x5y2) -- (x5y3);
    \draw[black] (x5y3) -- (x5y4);

    \draw[black] (x6y1) -- (x6y2);
    \draw[black] (x6y2) -- (x6y3);
    \draw[black] (x6y3) -- (x6y4);

    \node (legend) at ($(center)+(-0.5, 0.5)$) {\sffamily{\textbf{B}}};

    \node (center) at (-1,-3.5) {};

    \node (x1y1) [draw, rectangle, align=center, rotate=45, minimum width=5, minimum height=5, inner sep=0., fill=blue] at ($(center)+(0,0)$) {};
    \node (x2y1) [draw, rectangle, align=center, rotate=45, minimum width=5, minimum height=5, inner sep=0., fill=blue] at ($(x1y1)+(0.2,0.15)$) {};
    \node (x3y1) [draw, rectangle, align=center, rotate=45, minimum width=5, minimum height=5, inner sep=0., fill=blue] at ($(x2y1)+(0.3,0.1)$) {};
    \node (x4y1) [draw, rectangle, align=center, rotate=45, minimum width=5, minimum height=5, inner sep=0., fill=blue] at ($(x3y1)+(0.15,-0.2)$) {};

    \node (x1y2) [draw, rectangle, align=center, rotate=45, minimum width=5, minimum height=5, inner sep=0., fill=blue] at ($(x1y1)+(0.2,-0.5)$) {};
    \node (x2y2) [draw, rectangle, align=center, rotate=45, minimum width=5, minimum height=5, inner sep=0., fill=blue] at ($(x1y2)+(0.3,-0.25)$) {};
    \node (x3y2) [draw, rectangle, align=center, rotate=45, minimum width=5, minimum height=5, inner sep=0., fill=blue] at ($(x2y2)+(0.1,-0.05)$) {};
    \node (x4y2) [draw, rectangle, align=center, rotate=45, minimum width=5, minimum height=5, inner sep=0., fill=blue] at ($(x3y2)+(0.15,-0.15)$) {};

    \node (x1y3) [draw, rectangle, align=center, rotate=0, minimum width=5, minimum height=5, inner sep=0., fill=red] at ($(center)+(1.7,-1.5)$) {};
    \node (x2y3) [draw, rectangle, align=center, rotate=0, minimum width=5, minimum height=5, inner sep=0., fill=red] at ($(center)+(1.9,-1.3)$) {};
    \node (x3y3) [draw, rectangle, align=center, rotate=0, minimum width=5, minimum height=5, inner sep=0., fill=red] at ($(center)+(1.6,-1.2)$) {};
    \node (x4y3) [draw, rectangle, align=center, rotate=0, minimum width=5, minimum height=5, inner sep=0., fill=red] at ($(center)+(1.8,-1.25)$) {};
    
    \node (x1y4) [draw, rectangle, align=center, rotate=0, minimum width=5, minimum height=5, inner sep=0., fill=red] at ($(center)+(2.0,-1.5)$) {};
    \node (x2y4) [draw, rectangle, align=center, rotate=0, minimum width=5, minimum height=5, inner sep=0., fill=red] at ($(center)+(2.05,-1.25)$) {};
    \node (x3y4) [draw, rectangle, align=center, rotate=0, minimum width=5, minimum height=5, inner sep=0., fill=red] at ($(center)+(2.1,-1.3)$) {};
    \node (x4y4) [draw, rectangle, align=center, rotate=0, minimum width=5, minimum height=5, inner sep=0., fill=red] at ($(center)+(1.85,-1.4)$) {};
    
    \node (legend) at ($(center)+(-0.5, 0.5)$) {\sffamily{\textbf{C}}};

    \node (center) at ($(center)+(4.5,0)$) {};
    
    \node (x1y1) [draw, circle, minimum size=5, inner sep=0., fill=black] at ($(center)+(0,0+0.01)$) {};
    \node (x2y1) [draw, circle, minimum size=5, inner sep=0., fill=black] at ($(x1y1)+(0.2,0.15+0.01)$) {};
    \node (x3y1) [draw, circle, minimum size=5, inner sep=0., fill=black] at ($(x2y1)+(0.3,0.1+0.01)$) {};
    \node (x4y1) [draw, circle, minimum size=5, inner sep=0., fill=black] at ($(x3y1)+(0.15,-0.2+0.01)$) {};
    
    \node (x1y2) [draw, circle, minimum size=5, inner sep=0., fill=black] at ($(x1y1)+(0.2,-0.5+0.01)$) {};
    \node (x2y2) [draw, circle, minimum size=5, inner sep=0., fill=black] at ($(x1y2)+(0.3,-0.25+0.01)$) {};
    \node (x3y2) [draw, circle, minimum size=5, inner sep=0., fill=black] at ($(x2y2)+(0.1,-0.05+0.01)$) {};
    \node (x4y2) [draw, circle, minimum size=5, inner sep=0., fill=black] at ($(x3y2)+(0.15,-0.15+0.01)$) {};
    
    \node (x1y3) [draw, circle, minimum size=5, inner sep=0., fill=black] at ($(center)+(1.7,-1.5+0.01)$) {};
    \node (x2y3) [draw, circle, minimum size=5, inner sep=0., fill=black] at ($(center)+(1.9,-1.3+0.01)$) {};
    \node (x3y3) [draw, circle, minimum size=5, inner sep=0., fill=black] at ($(center)+(1.6,-1.2+0.01)$) {};
    \node (x4y3) [draw, circle, minimum size=5, inner sep=0., fill=black] at ($(center)+(1.8,-1.25+0.01)$) {};
    
    \node (x1y4) [draw, circle, minimum size=5, inner sep=0., fill=black] at ($(center)+(2.0,-1.5+0.01)$) {};
    \node (x2y4) [draw, circle, minimum size=5, inner sep=0., fill=black] at ($(center)+(2.05,-1.25+0.01)$) {};
    \node (x3y4) [draw, circle, minimum size=5, inner sep=0., fill=black] at ($(center)+(2.1,-1.3+0.01)$) {};
    \node (x4y4) [draw, circle, minimum size=5, inner sep=0., fill=black] at ($(center)+(1.85,-1.4+0.01)$) {};

    \draw[black] (x1y1) -- (x2y1);
    \draw[black] (x2y1) -- (x3y1);
    \draw[black] (x3y1) -- (x4y1);

    \draw[black] (x1y2) -- (x2y2);
    \draw[black] (x2y2) -- (x3y2);
    \draw[black] (x3y2) -- (x4y2);

    \draw[black] (x1y3) -- (x2y3);
    \draw[black] (x2y3) -- (x3y3);
    \draw[black] (x3y3) -- (x4y3);

    \draw[black] (x1y4) -- (x2y4);
    \draw[black] (x2y4) -- (x3y4);
    \draw[black] (x3y4) -- (x4y4);

    \draw[black] (x1y1) -- (x1y2);
    \draw[black] (x1y2) -- (x1y3);
    \draw[black] (x1y3) -- (x1y4);

    \draw[black] (x2y1) -- (x2y2);
    \draw[black] (x2y2) -- (x2y3);
    \draw[black] (x2y3) -- (x2y4);

    \draw[black] (x3y1) -- (x3y2);
    \draw[black] (x3y2) -- (x3y3);
    \draw[black] (x3y3) -- (x3y4);

    \draw[black] (x4y1) -- (x4y2);
    \draw[black] (x4y2) -- (x4y3);
    \draw[black] (x4y3) -- (x4y4);

    \node (legend) at ($(center)+(-0.5, 0.5)$) {\sffamily{\textbf{D}}};

    \node (center) at ($(center)+(4.5,0)$) {};
    
    \node (x1y1) [draw, regular polygon, regular polygon sides=6, align=center, rotate=0, minimum width=15, minimum height=15, inner sep=0., fill=blue] at ($(center)+(0,0)$) {};
    \node (x2y1) [draw, regular polygon, regular polygon sides=6, align=center, rotate=0, minimum width=15, minimum height=15, inner sep=0., fill=blue] at ($(x1y1)+(0.4,-0.23)$) {};
    \node (x3y1) [draw, regular polygon, regular polygon sides=6, align=center, rotate=0, minimum width=15, minimum height=15, inner sep=0., fill=blue] at ($(x2y1)+(0.4,0.23)$) {};
    \node (x4y1) [draw, regular polygon, regular polygon sides=6, align=center, rotate=0, minimum width=15, minimum height=15, inner sep=0., fill=blue] at ($(x3y1)+(0.4,-0.23)$) {};
    \node (x5y1) [draw, regular polygon, regular polygon sides=6, align=center, rotate=0, minimum width=15, minimum height=15, inner sep=0., fill=blue] at ($(x4y1)+(0.4,0.23)$) {};
    
    \node (x1y2) [draw, regular polygon, regular polygon sides=6, align=center, rotate=0, minimum width=15, minimum height=15, inner sep=0., fill=blue] at ($(x1y1)+(0,-0.45)$) {};
    \node (x2y2) [draw, regular polygon, regular polygon sides=6, align=center, rotate=0, minimum width=15, minimum height=15, inner sep=0., fill=blue] at ($(x1y2)+(0.4,-0.23)$) {};
    \node (x3y2) [draw, regular polygon, regular polygon sides=6, align=center, rotate=0, minimum width=15, minimum height=15, inner sep=0., fill=blue] at ($(x2y2)+(0.4,0.23)$) {};
    \node (x4y2) [draw, regular polygon, regular polygon sides=6, align=center, rotate=0, minimum width=15, minimum height=15, inner sep=0., fill=blue] at ($(x3y2)+(0.4,-0.23)$) {};
    \node (x5y2) [draw, regular polygon, regular polygon sides=6, align=center, rotate=0, minimum width=15, minimum height=15, inner sep=0., fill=red] at ($(x4y2)+(0.4,0.23)$) {};
    
    \node (x1y3) [draw, regular polygon, regular polygon sides=6, align=center, rotate=0, minimum width=15, minimum height=15, inner sep=0., fill=blue] at ($(x1y2)+(0,-0.45)$) {};
    \node (x2y3) [draw, regular polygon, regular polygon sides=6, align=center, rotate=0, minimum width=15, minimum height=15, inner sep=0., fill=blue] at ($(x1y3)+(0.4,0.23)$) {};
    \node (x3y3) [draw, regular polygon, regular polygon sides=6, align=center, rotate=0, minimum width=15, minimum height=15, inner sep=0., fill=blue] at ($(x2y3)+(0.4,-0.23)$) {};
    \node (x4y3) [draw, regular polygon, regular polygon sides=6, align=center, rotate=0, minimum width=15, minimum height=15, inner sep=0., fill=red] at ($(x3y3)+(0.4,0.23)$) {};
    \node (x5y3) [draw, regular polygon, regular polygon sides=6, align=center, rotate=0, minimum width=15, minimum height=15, inner sep=0., fill=red] at ($(x4y3)+(0.4,-0.23)$) {};
    
    \node (x1y4) [draw, regular polygon, regular polygon sides=6, align=center, rotate=0, minimum width=15, minimum height=15, inner sep=0., fill=blue] at ($(x1y3)+(0,-0.45)$) {};
    \node (x2y4) [draw, regular polygon, regular polygon sides=6, align=center, rotate=0, minimum width=15, minimum height=15, inner sep=0., fill=blue] at ($(x1y4)+(0.4,0.23)$) {};
    \node (x3y4) [draw, regular polygon, regular polygon sides=6, align=center, rotate=0, minimum width=15, minimum height=15, inner sep=0., fill=red] at ($(x2y4)+(0.4,-0.23)$) {};
    \node (x4y4) [draw, regular polygon, regular polygon sides=6, align=center, rotate=0, minimum width=15, minimum height=15, inner sep=0., fill=red] at ($(x3y4)+(0.4,0.23)$) {};
    \node (x5y4) [draw, regular polygon, regular polygon sides=6, align=center, rotate=0, minimum width=15, minimum height=15, inner sep=0., fill=red]at ($(x4y4)+(0.4,-0.23)$) {};

    \node (legend) at ($(center)+(-0.5, 0.5)$) {\sffamily{\textbf{E}}};
\end{tikzpicture}

%% file: recurrent/som/som_inspired.tex
\subsubsection{Self organizing segmentation}\label{section:som_segmentation}

In the segmentation task, we can apply a type of self organizing recurrency given the belief segmentation map of an instance. We will introduce a different notation where a representation of node $i$ is denoted as $\mathbf{v}_i$. Consider that the nodes are initialized using $\mathbf{v}_i = f(\mathbf{x};\theta)_i \in \mathbb{R}^{H \times W \times L}, \forall i \in \{1, \dots, H\times W\}$, which corresponds to a segmentation belief map. If we run the algorithm on the set of nodes itself, we will be updating the map representation recurrently to reduce energy.

\textbf{Self-organizing segmentation.} The energy of a segmentation belief map is computed from neighbor nodes as
\begin{equation}\label{equation:som_energy}
    E^{\mbox{SOM}} = \sum_i^{H\times W} \frac{1}{5} \left(\sum \frac{1}{2}\left( \mathcal{N}(i) - \mathcal{N}(i)^\top \right)^2 \right),
\end{equation}
with $\mathcal{N}(i) - \mathcal{N}(i)^\top$ representing the distance matrix of the set of neighbors of node $\mathbf{v}_i$.

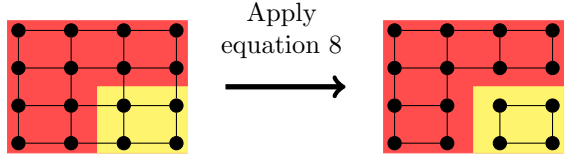
\begin{figure}[ht]
    \centering
    \input{recurrent/som/figures/component}
    \caption{Given a filter, we can produce a grid-like graph with more than one connected component, where each component will have pixels from only one class.}
    \label{fig:connected_component}
\end{figure}

\textbf{Component-based graph.} To help the correct belief propagation through the square lattice, the edges that connect two pixels with a gradient are dropped. Let $F(\mathbf{x})\in \mathbb{R}^{H\times W}$ be a filter, for instance, the Laplacian operator, Hessian filter \cite{frangi1998multiscale}, or even a neural-based filter. We drop edges from the square lattice (figure \ref{fig:som_grid} A) with a message passing of the form
\begin{equation}\label{equation:gradient_component}
    \phi_{i \to j}(\mathbf{x}) = \left(|F(\mathbf{x})_i - \mathcal{N}\left(F(\mathbf{x})_i\right)_j|\leq\epsilon\right),
\end{equation}
where ${i \to j}$ specifies a message from node $i$ to node $j$. If $\phi_{i \to j}(\mathbf{x})$ holds true, then there is an edge connecting node $j$ to the respective neighbor node $i$. Do this for all nodes and we should end up with an image with well-defined components around the semantic representations of interest, see figure \ref{fig:connected_component}.


\textbf{Connected component equilibrium.} To propagate the beliefs of each node throughout the lattice, we are going to take advantage of message passing. The update of a node $j$ is influenced by a neighbour $i$ with a message defined as $\phi_{i\to j}(\mathbf{x})\times\left( \mathbf{v}_i - \mathbf{v}_j \right)$, which is the gradient to approximate $\mathbf{v}_j$ to $\mathbf{v}_i$. The use of $\phi_{i\to j}$ cuts the communication between nodes that belong to different components. This type of propagation cuts off the mixing of different beliefs. The update of a node is given by
\begin{equation}\label{equation:uniform_som_update}
    \mathbf{v}_j = \mathbf{v}_j + \alpha \left( \frac{1}{\sum_{i \in \mathcal{N}(j)} \phi_{i \to j}(\mathbf{x}) } \sum_{i \in \mathcal{N}(j)} \phi_{i\to j}(\mathbf{x})\times\left( \mathbf{v}_i - \mathbf{v}_j \right) \right),
\end{equation}
which corresponds to the gradient descent update of node $\mathbf{v}_j$ according to the energy defined in function \ref{equation:som_energy}.

\textbf{Propagate high responses.} Although the self-organizing algorithm has a competitive nature, in the segmentation task we want the network to make its prediction based on the neurons that have a high response and propagate sparse beliefs. It has been shown that using sparse activations that respond strongly to stimuli improves the performance of models \cite{li2024emergence}. To do this, we define the response of a neuron $i$ as $\mathbf{r}_i = \sum \frac{1}{2} (\mathbf{v}_i - \mathbf{v}_i^\top)^2$. The update rule becomes
\begin{equation}\label{equation:high_response_message_passing}
    \mathbf{v}_j = \mathbf{v}_j + \alpha \left( \sum_{i \in \mathcal{N}(j)}  \frac{\mbox{exp}\left(\mathbf{r}_i \times \phi_{i \to j}(\mathbf{x}) \right)}{\mbox{exp}(\mathbf{r}_j) + \sum_{k \in \mathcal{N}(j)} \mbox{exp}\left(\mathbf{r}_i \times \phi_{i \to j}(\mathbf{x}) \right) } \times\left( \mathbf{v}_i - \mathbf{v}_j \right) \right),
\end{equation}
where the main difference from the update of equation \ref{equation:uniform_som_update} is the softmax normalization based on neuron response. This allows a neuron with low response, i.e. low norm, to approximate a neuron from the same component that has a high norm. In addition, a neuron with a high response will not be nudged by neighbors with a lower response because we are adding $\mbox{exp}(\mathbf{r}_j)$ to the denominator of the softmax. This type of update is particularly interesting because it propagates high responses, such that when an attractor state is reached, neurons from the same component will have equal (likely high) responses.

%% file: recurrent/som/figures/component.tex
\begin{tikzpicture}
    \node (center) at (0,0) {};

    \node (rectangle1) [draw, rectangle, fill=red!70, red!70, minimum width=68, minimum height=50] at ($(center)+(1.05,-0.75)$) {};
    \node (rectangle2) [draw, rectangle, fill=yellow!70, yellow!70, minimum width=34, minimum height=25] at ($(center)+(1.65,-1.19)$) {};

    \node (x1y1) [draw, circle, minimum size=5, inner sep=0., fill=black] at ($(center)+(0,0)$) {};
    \node (x2y1) [draw, circle, minimum size=5, inner sep=0., fill=black] at ($(x1y1)+(0.7,0)$) {};
    \node (x3y1) [draw, circle, minimum size=5, inner sep=0., fill=black] at ($(x2y1)+(0.7,0)$) {};
    \node (x4y1) [draw, circle, minimum size=5, inner sep=0., fill=black] at ($(x3y1)+(0.7,0)$) {};

    \node (x1y2) [draw, circle, minimum size=5, inner sep=0., fill=black] at ($(x1y1)+(0,-0.5)$) {};
    \node (x2y2) [draw, circle, minimum size=5, inner sep=0., fill=black] at ($(x1y2)+(0.7,0)$) {};
    \node (x3y2) [draw, circle, minimum size=5, inner sep=0., fill=black] at ($(x2y2)+(0.7,0)$) {};
    \node (x4y2) [draw, circle, minimum size=5, inner sep=0., fill=black] at ($(x3y2)+(0.7,0)$) {};

    \node (x1y3) [draw, circle, minimum size=5, inner sep=0., fill=black] at ($(x1y2)+(0,-0.5)$) {};
    \node (x2y3) [draw, circle, minimum size=5, inner sep=0., fill=black] at ($(x1y3)+(0.7,0)$) {};
    \node (x3y3) [draw, circle, minimum size=5, inner sep=0., fill=black] at ($(x2y3)+(0.7,0)$) {};
    \node (x4y3) [draw, circle, minimum size=5, inner sep=0., fill=black] at ($(x3y3)+(0.7,0)$) {};

    \node (x1y4) [draw, circle, minimum size=5, inner sep=0., fill=black] at ($(x1y3)+(0,-0.5)$) {};
    \node (x2y4) [draw, circle, minimum size=5, inner sep=0., fill=black] at ($(x1y4)+(0.7,0)$) {};
    \node (x3y4) [draw, circle, minimum size=5, inner sep=0., fill=black] at ($(x2y4)+(0.7,0)$) {};
    \node (x4y4) [draw, circle, minimum size=5, inner sep=0., fill=black] at ($(x3y4)+(0.7,0)$) {};

    \draw[black] (x1y1) -- (x2y1);
    \draw[black] (x2y1) -- (x3y1);
    \draw[black] (x3y1) -- (x4y1);

    \draw[black] (x1y2) -- (x2y2);
    \draw[black] (x2y2) -- (x3y2);
    \draw[black] (x3y2) -- (x4y2);

    \draw[black] (x1y3) -- (x2y3);
    \draw[black] (x2y3) -- (x3y3);
    \draw[black] (x3y3) -- (x4y3);

    \draw[black] (x1y4) -- (x2y4);
    \draw[black] (x2y4) -- (x3y4);
    \draw[black] (x3y4) -- (x4y4);

    \draw[black] (x1y1) -- (x1y2);
    \draw[black] (x1y2) -- (x1y3);
    \draw[black] (x1y3) -- (x1y4);

    \draw[black] (x2y1) -- (x2y2);
    \draw[black] (x2y2) -- (x2y3);
    \draw[black] (x2y3) -- (x2y4);

    \draw[black] (x3y1) -- (x3y2);
    \draw[black] (x3y2) -- (x3y3);
    \draw[black] (x3y3) -- (x3y4);

    \draw[black] (x4y1) -- (x4y2);
    \draw[black] (x4y2) -- (x4y3);
    \draw[black] (x4y3) -- (x4y4);

    \node (center) at (5,0) {};

    \node (rectangle3) [draw, rectangle, fill=red!70, red!70, minimum width=68, minimum height=50] at ($(center)+(1.05,-0.75)$) {};
    \node (rectangle4) [draw, rectangle, fill=yellow!70, yellow!70, minimum width=34, minimum height=25] at ($(center)+(1.65,-1.19)$) {};

    \node (x1y1) [draw, circle, minimum size=5, inner sep=0., fill=black] at ($(center)+(0,0)$) {};
    \node (x2y1) [draw, circle, minimum size=5, inner sep=0., fill=black] at ($(x1y1)+(0.7,0)$) {};
    \node (x3y1) [draw, circle, minimum size=5, inner sep=0., fill=black] at ($(x2y1)+(0.7,0)$) {};
    \node (x4y1) [draw, circle, minimum size=5, inner sep=0., fill=black] at ($(x3y1)+(0.7,0)$) {};

    \node (x1y2) [draw, circle, minimum size=5, inner sep=0., fill=black] at ($(x1y1)+(0,-0.5)$) {};
    \node (x2y2) [draw, circle, minimum size=5, inner sep=0., fill=black] at ($(x1y2)+(0.7,0)$) {};
    \node (x3y2) [draw, circle, minimum size=5, inner sep=0., fill=black] at ($(x2y2)+(0.7,0)$) {};
    \node (x4y2) [draw, circle, minimum size=5, inner sep=0., fill=black] at ($(x3y2)+(0.7,0)$) {};

    \node (x1y3) [draw, circle, minimum size=5, inner sep=0., fill=black] at ($(x1y2)+(0,-0.5)$) {};
    \node (x2y3) [draw, circle, minimum size=5, inner sep=0., fill=black] at ($(x1y3)+(0.7,0)$) {};
    \node (x3y3) [draw, circle, minimum size=5, inner sep=0., fill=black] at ($(x2y3)+(0.7,0)$) {};
    \node (x4y3) [draw, circle, minimum size=5, inner sep=0., fill=black] at ($(x3y3)+(0.7,0)$) {};

    \node (x1y4) [draw, circle, minimum size=5, inner sep=0., fill=black] at ($(x1y3)+(0,-0.5)$) {};
    \node (x2y4) [draw, circle, minimum size=5, inner sep=0., fill=black] at ($(x1y4)+(0.7,0)$) {};
    \node (x3y4) [draw, circle, minimum size=5, inner sep=0., fill=black] at ($(x2y4)+(0.7,0)$) {};
    \node (x4y4) [draw, circle, minimum size=5, inner sep=0., fill=black] at ($(x3y4)+(0.7,0)$) {};

    \draw[black] (x1y1) -- (x2y1);
    \draw[black] (x2y1) -- (x3y1);
    \draw[black] (x3y1) -- (x4y1);

    \draw[black] (x1y2) -- (x2y2);
    \draw[black] (x2y2) -- (x3y2);
    \draw[black] (x3y2) -- (x4y2);

    \draw[black] (x1y3) -- (x2y3);
    \draw[black] (x3y3) -- (x4y3);

    \draw[black] (x1y4) -- (x2y4);
    \draw[black] (x3y4) -- (x4y4);

    \draw[black] (x1y1) -- (x1y2);
    \draw[black] (x1y2) -- (x1y3);
    \draw[black] (x1y3) -- (x1y4);

    \draw[black] (x2y1) -- (x2y2);
    \draw[black] (x2y2) -- (x2y3);
    \draw[black] (x2y3) -- (x2y4);

    \draw[black] (x3y1) -- (x3y2);
    \draw[black] (x3y3) -- (x3y4);

    \draw[black] (x4y1) -- (x4y2);
    \draw[black] (x4y3) -- (x4y4);

    \node [align=center] (legend) at (3.5,0.) {Apply\\equation \ref{equation:gradient_component}};
    \draw[->, line width=2] ($(rectangle1.east)+(0.5,0)$) -- ($(rectangle3.west)+(-0.5,0)$);
\end{tikzpicture}

%% file: recurrent/crf/crf.tex

\subsection{Conditional random field for segmentation}

This method is parametrized by $\mathbf{w}$ which encodes the compatibility between two labels. We define the energy as
\begin{equation}
    E^{\mbox{CRF}}= - \sum_i \sum_{j \neq i} \mathbf{w}\cdot \left( k_G(\mathbf{x}_i,\mathbf{x}_j) \cdot (\mathbf{v}_j^\top \cdot \mathbf{v}_i) \right),
\end{equation}
where $k_G$ computes a probability mass function that encodes the similarity between the features of pixels $i$ and $j$ as
\begin{equation}
    k_G(\mathbf{a},\mathbf{b}) = \frac{\mbox{exp}\left(-(\mathbf{a} - \mathbf{b})^2\right)}{\sum_l \mbox{exp}\left(-(\mathbf{a}_l-\mathbf{b}_l)^2\right)}.
\end{equation}
By deriving this energy w.r.t. $\mathbf{v}_i$, we get
\begin{equation}
    \frac{\partial E^{\mbox{CRF}}}{\partial \mathbf{v}_i} = - \sum_i \sum_{j \in \mathcal{N}(i)} \mathbf{w} \cdot \left( k_G(\mathbf{x}_i,\mathbf{x}_j) \cdot (\mathbf{v}_j) \right),
\end{equation}
which results in the following update rule
\begin{equation}
    \mathbf{v}_i = \mathbf{v}_i + \alpha \left( \sum_{j \in \mathcal{N}(i)} \mathbf{w} \cdot \left( k_G(\mathbf{x}_i,\mathbf{x}_j) \cdot (\mathbf{v}_j) \right) \right).
\end{equation}

%% file: recurrent/hopfield/hopfield.tex
\subsection{Associative memories}


\citet{hopfield1982neural} first introduced the notion of associative memories. It consists of a network, whose weights connect features. Given a set of memories $\xi$ and a state pattern $\mathbf{v}$, the dynamical system evolves according to an energy function
\begin{equation}
    E = - G(\xi \cdot \mathbf{v}) + \mathbf{v}^\top \cdot \mathbf{v}.
\end{equation}
This type of energy results in the dynamics
\begin{equation}
    \tau \frac{d \mathbf{v}}{ dt} = \xi^\top \cdot g(\xi \cdot \mathbf{v}) - \mathbf{v},
\end{equation}
where $g=G'$ is the derivative of function $G$.

In this section, we treat $\mathbf{v}_i$ as a patch of an image representation, instead of a single voxel (as in the previous sections). According to what was done by \citet{hoover2024energy}, we focus on the associative memory component of their study and use patch representation as $\mathbf{v}_i \in \mathbb{R}^{P \times C}$ (see figure \ref{fig:hopfield_patches}). $P$ denotes the dimension of the patch, which is $\ll M$.

\subsubsection{Hopfield networks for segmentation}

Given a set of learnable memories $\xi \in \mathbb{R}^{I_1 \times I_0}$, where $I_0 = P \times C $, $P$ is the size of a flattened image patch, and $I_1$ is the size of the memory dimension; we define the energy of Hopfield lattice network as
\begin{equation}
    E^{HN} = - \sum_i^{I_1} log\left(  \sum_{j}^{I_0} \mbox{exp}(\xi \cdot \mathbf{v}_i)  \right) + \frac{1}{2} \mathbf{v}_i^\top \cdot \mathbf{v}_i.
\end{equation}

In this setting we chose $G= LogSumExp$ function, whose derivative is the softmax function. The update rule for a node $i$ is
\begin{equation}
    \mathbf{v}_i = \mathbf{v}_i + \alpha \left( \sum_j \xi^\top \cdot \mbox{softmax}\left( \xi \cdot \mathbf{v}_i \right) - \mathbf{v}_i \right).
\end{equation}

\input{recurrent/hopfield/figures/diagram}

In addition to the softmax, one may have more functions to select the \textit{memory} that won for the state $\mathbf{v}$. For instance, choosing $G= x^2$ results in representing $g = ReLU$ \cite{krotov2016dense}. The ReLU function has the advantage of suppressing memories when $x<0$, but it may result in a collection of too many memories for $x>0$. Functions with sparser activations \cite{martins2016softmax} may be a good choice. Typically, researchers work with $G = LogSumExp$ and tweak the temperature parameter. However, one can manipulate the energy landscape to force states to reach fixed point attractors instead of metastable states \cite{martins2023sparse}.



%% file: recurrent/hopfield/figures/diagram.tex
\begin{wrapfigure}{r}{0.5\textwidth}
    \begin{center}
        \vspace{-0.5cm}
        \begin{tikzpicture}

            \node (center) at (0,0) {};
            
            \node (original) at ($(center)+(0,0)$) {\includegraphics[width=0.2\textwidth]{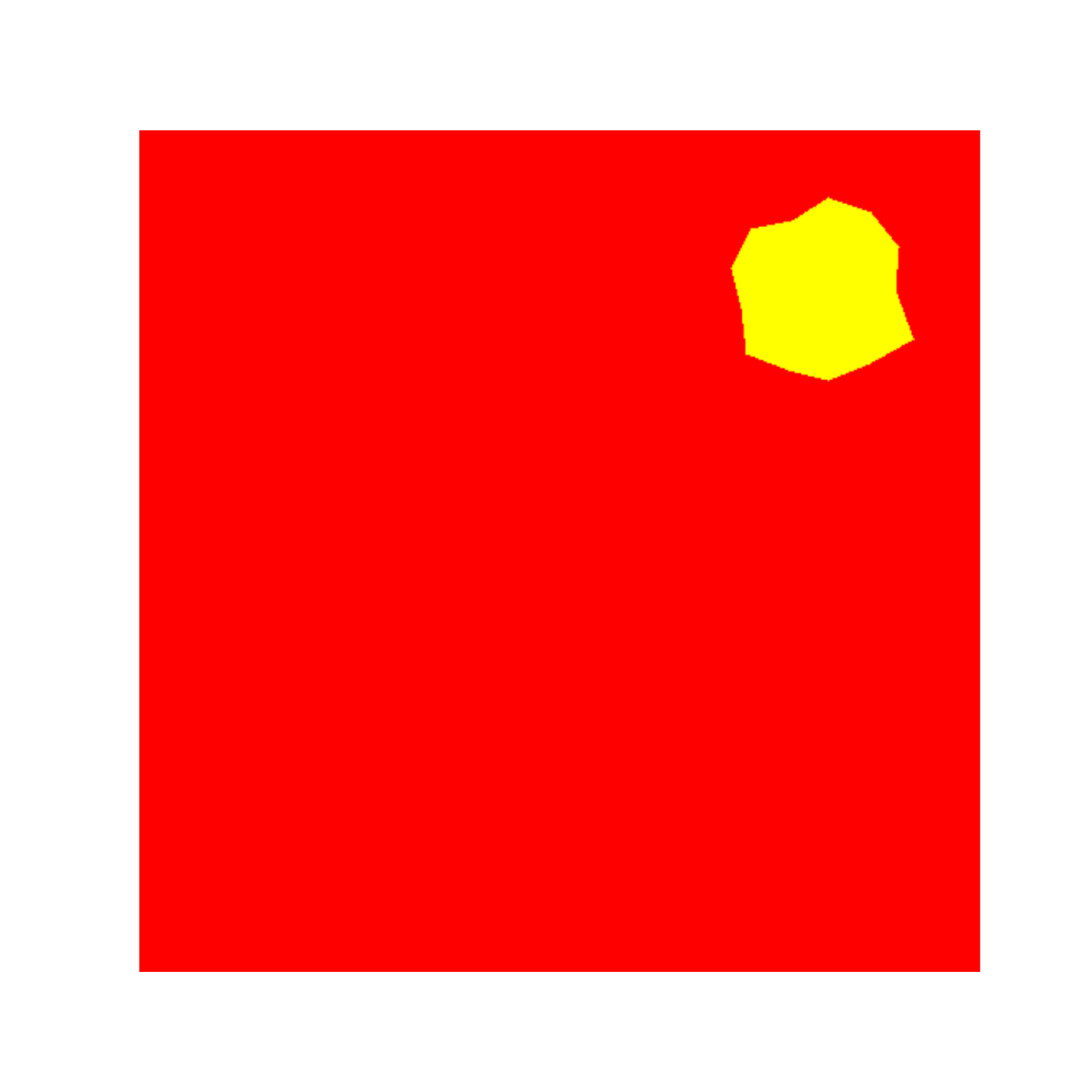}};
            \node (legend) at ($(original.west)+(0., 1.6)$) {\sffamily{\textbf{A}}};

            \node [anchor=west] (patchify) at ($(original.east)+(0.1,0)$) {\includegraphics[width=0.2\textwidth]{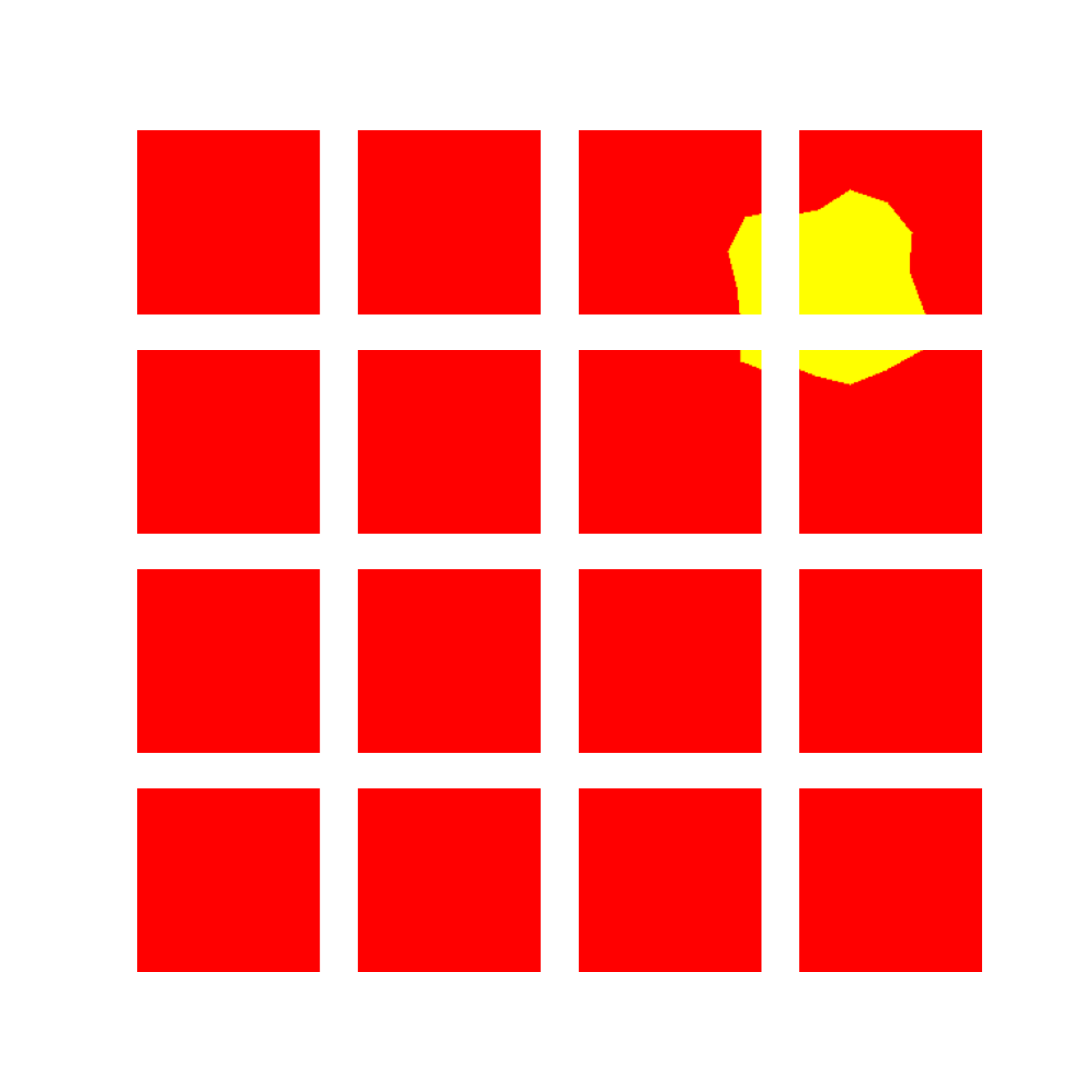}};
            \node (legend) at ($(patchify.west)+(0., 1.6)$) {\sffamily{\textbf{B}}};
        \end{tikzpicture}    
    \end{center}
    \vspace{-0.5cm}
    \caption{The original image, figure {\sffamily{\textbf{A}}}, is split into equal sized patches, figure {\sffamily{\textbf{A}}}. These patches are then used as tokens for the memory retrieval process of the modern Hopfield network.}
    \label{fig:hopfield_patches}
\end{wrapfigure}
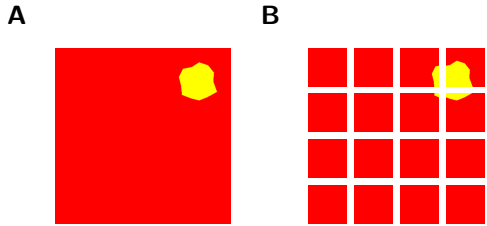

%% file: setting/setting.tex
\section{Experimental setting}\label{section:setting}

We proceed to describe the experimental setting for the results presented in this manuscript. The parameters used in the learning session were: Adam \cite{kingma2014adam} optimizer with a learning rate set to $1e-2$; batch size of $1$; $10$ epochs. All the results, shown in section \ref{section:results}, are averaged through $5$ simulations with different seeds.

\subsection{Datasets}\label{subsection:datasets}

\input{setting/dataset/datasets}

\subsection{Noise corrupted instances}\label{subsection:noise}

\input{setting/noise/noise}

\subsection{Learning from limited examples}\label{subsection:examples}

\input{setting/number_examples/number_examples}

\subsection{Loss function and evaluation metrics}\label{subsection:metrics}

\input{setting/metrics/metrics}

%% file: setting/dataset/datasets.tex
In this section, two datasets are described: Shapes and Catheter Artery Segmentation (CAD) \cite{lourencco2021encoder}. The first is an artificial dataset generated randomly and the second is a medical imaging dataset. 

\subsubsection{Artificial shapes data}\label{dataset:shapes}

\input{setting/dataset/artificial}

\subsubsection{Catheter artery segmentation data}\label{dataset:cad}

\input{setting/dataset/cad}

%% file: setting/dataset/artificial.tex
We define a dataset to test the segmentation performance of the models under noisy and limited sample settings. For this, we take into consideration polygons and circles (a circle being a polygon with $\infty$ sides). Each instance contains two classes: a background and a shape class. Please refer to algorithm \ref{algorithm:shapes_dataset} for a detailed explanation of this dataset.

\begin{algorithm}[ht]
    \caption{Artificial data generation algorithm}\label{algorithm:shapes_dataset}
    $N$ \Comment*[r]{Argument for size of dataset}
    $L \gets \{ 3, 4, 5, 6, 7, 8, 9 , 10, 11, 12, 13, \infty \}$ \Comment*[r]{15 classes in total}
    $\mathcal{D} = \{  \} $\;
    \While{$N > 0$}{
        $l \sim \mathcal{U}(L)$ \Comment*[r]{Chosse a class}
        $\mathbf{x} \gets generate(l)$ \Comment*[r]{function that generates a shape based on a class}
        $\mathbf{y} \gets one\_hot(l)$\;
        $\mathcal{D} = \mathcal{D} \cup \left( \mathbf{x}, \mathbf{y} \right)$\;
        $N \gets N-1$\;
    }
    $\mathcal{D}$\;
\end{algorithm}


%% file: setting/dataset/cad.tex
This dataset contains X-ray images of arteries and their corresponding masks. This dataset has a total of 515 X-ray coronary angiography scan images, with a $512 \times 512$ resolution. Multiple cardiology experts manually labeled the associated segmentation masks. The classes considered for segmentation were: background, vessel, and catheter. Patients gave consent to have their X-ray data recorded and included in this \textit{private dataset}. For this dataset, we perform the same types of data augmentation operations done in \citet{lourencco2021encoder}: image rotation with a $\pm20$ degree angle; shifts of $10\%$; zoom of $\pm 10 \%$; and brightness change of $\pm 10\%$. 

%% file: setting/noise/noise.tex
We corrupt instances with Gaussian noise. This process takes the form
\begin{equation}
    \mathbf{\tilde{x}} = \mathbf{x} + n: n \sim \mathcal{N}(0, \epsilon),
\end{equation}
with the new dataset taking the form $\mathcal{D} = \bigcup (\mathbf{\tilde{x}}, \mathbf{y})$, such that only the inputs are corrupted.

The datasets described in section \ref{subsection:datasets} have different distributions and, as a consequence, different levels of noise impact the performance differently. To assess the impact of noise, we want to choose a set of noise values where the performance decreases for at least one of the models. With that in mind, we ran preliminary experiments, and the chosen sets are:
\begin{itemize}
    \item \textbf{Artificial shapes dataset}: $\epsilon \in \{ 10, 20, 30, 40, 50, 60, 70, 80, 90, 100 \}$;
    \item \textbf{Catheter artery segmentation data}: $\epsilon \in \{ 0, 1, 2, 3, 4, 5, 6, 7, 8, 9, 10 \}$.
\end{itemize}
We include $\epsilon=0$ in the CAD data experiments because we hypothesize that the complexity is much greater for this dataset. While we have irregular shapes present in the artificial shapes dataset, the nature of the data somehow follows a known distribution, whereas the CAD data is closer to a truly random, but unknown, distribution. The vessels of different individuals do not have the same shape and given the context of a vessel, one is not able to infer the true shape of a given location. We hypothesize memory retrieval recurrency will not have as good a performance as in the artificial shapes data.

%% file: setting/number_examples/number_examples.tex
In the limited sample setting, we want to study the impact of the size of $\mathcal{D}$. For this, we limit the size of the training set and leave the test set size\footnote{\textbf{Test size}: we set the test size to $200$ instances for the artificial shapes data and we consider $20\%$ of the whole CAD dataset as the test set} the same.

Consider $\mathcal{D}$, the training set, defined by 
\begin{equation}
    \mathcal{D} = \bigcup_{i=1}^{N} \left( \mathbf{x}_i, \mathbf{y}_i \right),
\end{equation}
where $N$ is the size, also referred to as $|\mathcal{D}|$. The values for size are $N = \{ 10, 20, 40, 80, 160, 320 \}$, for both datasets. We also test $N=2$ for the CAD dataset.

%% file: setting/metrics/metrics.tex
We use the same loss \citet{lourencco2021encoder}  used, for both feed-forward and recurrent versions. The loss function is defined by two components: \textit{generalized dice loss} and a \textit{focal loss}. The generalized dice loss is the conjugate of the score version of this metric, defined as
\begin{equation}
    GDL = 1- GDS = 1 - 2 \frac{\sum_{l=1}^{L} w_l \sum_{i=1}^{H\times W} g_{li} p_{li}}{\sum_{l=1}^{L} w_l \left(\sum_{i=1}^{H\times W} g_{li} + p_{li}\right)},
\end{equation}
which is translated to the final form of this loss component as
\begin{equation}
    \mathcal{L}_G = \frac{GDL}{1+ k (1-GDL)},
\end{equation}
where $k$ is set to $0.75$. The $g_{li}$ is the ground truth and $p_{li}$ is the probability given to pixel $i$ for class $l$. Class weights, $w_l$, are set to 
\begin{equation}
w_l = \frac{1}{\left( \sum_{i=1}^{H\times W} g_{li} \right)^2}
\end{equation}
to address class imbalance. This is particularly useful for medical data, i.e. in the CAD dataset.

The second component is the focal loss that, as the name says, allows the model to focus more on a class that appears with a certain density. Let 
\begin{equation}
    p'_{li} = \begin{cases}
        p_{li} & \mbox{if } g_{li}=1,\\
        1-p_{li} & \mbox{if } g_{li}=0.\\
    \end{cases}
\end{equation}
Then the loss takes the form
\begin{equation}
    \mathcal{L}_F = - \alpha' (1-p'_{li})^\gamma log(p'_{li}),
\end{equation}
where $\alpha$ balances positive and negative samples and $\alpha'$ is derived similarly to $p'_{li}$, $\gamma$ forces the model to learn classes that may appear with more or less density. We set the value of $\gamma = 2$ and $\alpha= 0.25$ \cite{lourencco2021encoder}.

The loss given to the optimization problem is 
\begin{equation}
    \mathcal{L} = \mathcal{L}_G + \mathcal{L}_F.
\end{equation}

To evaluate the quality of the segmentation belief maps we use the intersection over union (IoU). We also look into the precision, the amount of hits for a class $i$ divided by all of the misses it made predicting $i$ when in fact it was another class $j$, and recall, $i$ is the hits divided by the misses predicting a class $j$ when it was $i$
\begin{align}
    IoU = \frac{\sum_i T_i}{\sum_i \sum_j T_i+F_{i|j}+F_{j|i}}, \\ \mbox{precision}_i = \frac{T_i}{\sum_j F_{i|j}}, \\ \mbox{recall}_i = \frac{T_i}{\sum_j F_{j|i}}.
\end{align}

%% file: results/results.tex
\section{Results}\label{section:results}

\input{results/figures/artificial/iou}

The results on the artificial dataset experiments are illustrated in figures \ref{fig:iou_artificial}.  Regarding the impact of increasing noise, we observe that all models deteriorate their segmentation as the noise level increases. The model that performs best in high-noise settings is the self-organizing recurrency, showing a more stable segmentation quality as the number of iterations increases. Right after, the feed-forward version had the second-best performance, decreasing at the same rate as the SOM. The CRF recurrency seems to begin with a good segmentation, but as the number of iterations increases the segmentation quality decreases. The latter is a quality we do not want to see in recurrent models. This happens at all noise values tested. Though the Hopfield recurrency had the worst performance out of all the baselines studied, it showed an increase as the number of iterations increased. This shows that the initial segmentation belief was flawed, but the memory retrieval mechanism improves the quality and enables the system to reach an equilibrium quickly. In the limited sample setting, the results do not show as much as large a discrepancy as in the noise experiments. The feed-forward and the self-organizing recurrency are the best models in the limited sample test. They appear to have the same performance for all values, with the self-organizing recurrency having a slightly better performance in the smallest sample experiment for $|\mathcal{D}|=10$ (the most interesting number of samples, since we want to assess which algorithm performs best in small sample dataset). The Hopfield recurrency performed third best overall. It showed the same performance for all the values tested and it decreased its segmentation quality after $20$ iterations and stabilized after. The CRF was the worst of all baselines in this setting, with high variability with the number of iterations. Regarding the CRF dynamics, it was similar to the noise experiments, where it increased its segmentation quality after the first $20$ iterations, peaking at $40$ iterations, and stabilizing at $60$ iterations in a worse segmentation quality. Nonetheless, the segmentation quality at $40$ iterations is competitive with the other baselines. 

\input{results/figures/artificial/pvalues_artificial}

In order to better assess, for each setting, which baselines were the best, we ran statistical $t$-tests (illustrated in figure \ref{fig:pvalues_artificial}). These tests were performed with the average of the other dimensions. For instance, the statistical tests for the limited sample values, shown in figure \ref{fig:pvalues_artificial} {\sffamily{C}}, were done with the iterations, classes, and seed dimensions averaged. We decided to illustrate the superiority of a model with statistical significance with the same color scheme of the bar plot (figure \ref{fig:iou_artificial}), but with the difference that if there is no statistical significance of the best model, i.e. $p>0.05$, we represent it with the color black. The color bar has four dimensions, one for each model, and it is illustrated in figure \ref{fig:pvalues_artificial} {\sffamily{A}}. We start by assessing which classes had the most impact on the performance of the models (figure \ref{fig:pvalues_artificial} {\sffamily{B}}). For instance, it is hard to distinguish a polygon with $40$ sides from a circle. Overall there is no statistical significance in the precision, both for the noise and limited sample experiments. Regarding the recall of each class, the self-organizing recurrency had statistical significance for most classes in the noise setting, which is in accordance with previous observations. The Hopfield recurrency had statistical significance in most classes, for the number of examples tests, which shows that the superiority of the feed-forward and self-organizing recurrency in the IoU metric did not translate to the recall metric. When it comes to the number of limited samples (figure \ref{fig:pvalues_artificial} {\sffamily{C}}), we observe different phenomena: Hopfield had statistical significance in \textit{precision} for all values, and the self-organizing recurrency had statistical significance in \textit{recall} for all values. This shows the importance of looking from different perspectives/dimensions. Regarding the noise setting (figure \ref{fig:pvalues_artificial} {\sffamily{D}}), the self-organizing recurrency had statistical significance in precision for some values, which is in accordance with the bar plots (figure \ref{fig:iou_artificial} {\sffamily{B}}).

\input{results/figures/cad/iou}

Figure \ref{fig:iou_cad} shows the performance of the models in a medical imaging setting (CAD dataset). Though the IoU plots show a reasonable performance for all the models, the precision for each class of interest (\textit{catheter} and \textit{artery}), was low. Regarding the noise setting, it seems that even small amounts of noise decrease the performance very fast. The feed-forward version had the best performance in low-noise settings. On the other hand, recurrency appeared to be better than feed-forward in high-noise settings. As the number of iterations increases, the precision for the \textit{catheter} and \textit{artery} classes deteriorates rapidly for the SOM and CRF models. The Hopfield recurrency showed again stability, meaning that it reached an equilibrium state. However, in low noise values, the Hopfield recurrency had the worst precision. In general, the low precision and high recall phenomena tell us that the models were mostly predicting background. Regarding the size of the training set, there is no clear consensus of the best model in terms of IoU. The feed-forward and self-organizing recurrency are the best models. The CRF and Hopfield recurrency have trouble handling limited sample settings, with the worst performance when $|\mathcal{D}|=2$. The performance of all models increases when the training set size increases as well. However, when we look at precision and recall, of the classes of interest, recurrency seems to handle better $|\mathcal{D}|=\{2,10\}$ settings.

\input{results/figures/cad/pvalues_cad}

Statistical $t$-tests, see figure \ref{fig:pvalues_cad}, do not have a general consensus on which is the best model for each class. In terms of noise, the feed-forward version is the best one with statistical significance for the recall metric. For the set size, Hopfield had statistical significance in the precision metric.

%% file: results/figures/artificial/iou.tex
\begin{figure}[t]
    \centering
    \begin{tikzpicture}
        \node (center) at (0,0) {};

        \node (noise) at ($(center)+(0,0)$) {\includegraphics[clip, trim={0.cm, 0cm, 0cm, 3cm}, width=0.45\textwidth]{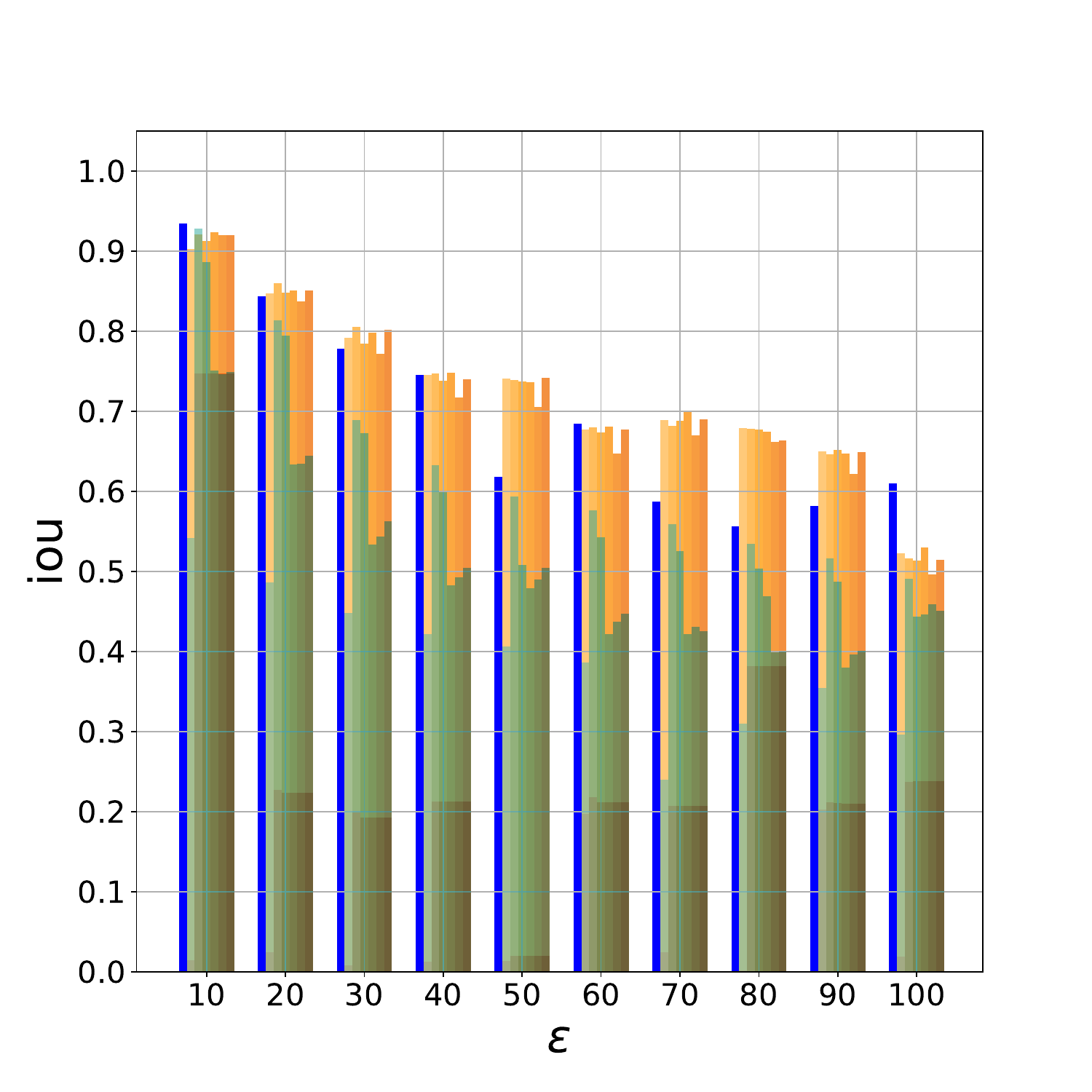}}; 
        \node (examples) [anchor=west] at ($(noise.east)+(0.,0)$) {\includegraphics[clip, trim={0.cm, 0cm, 0cm, 3cm}, width=0.45\textwidth]{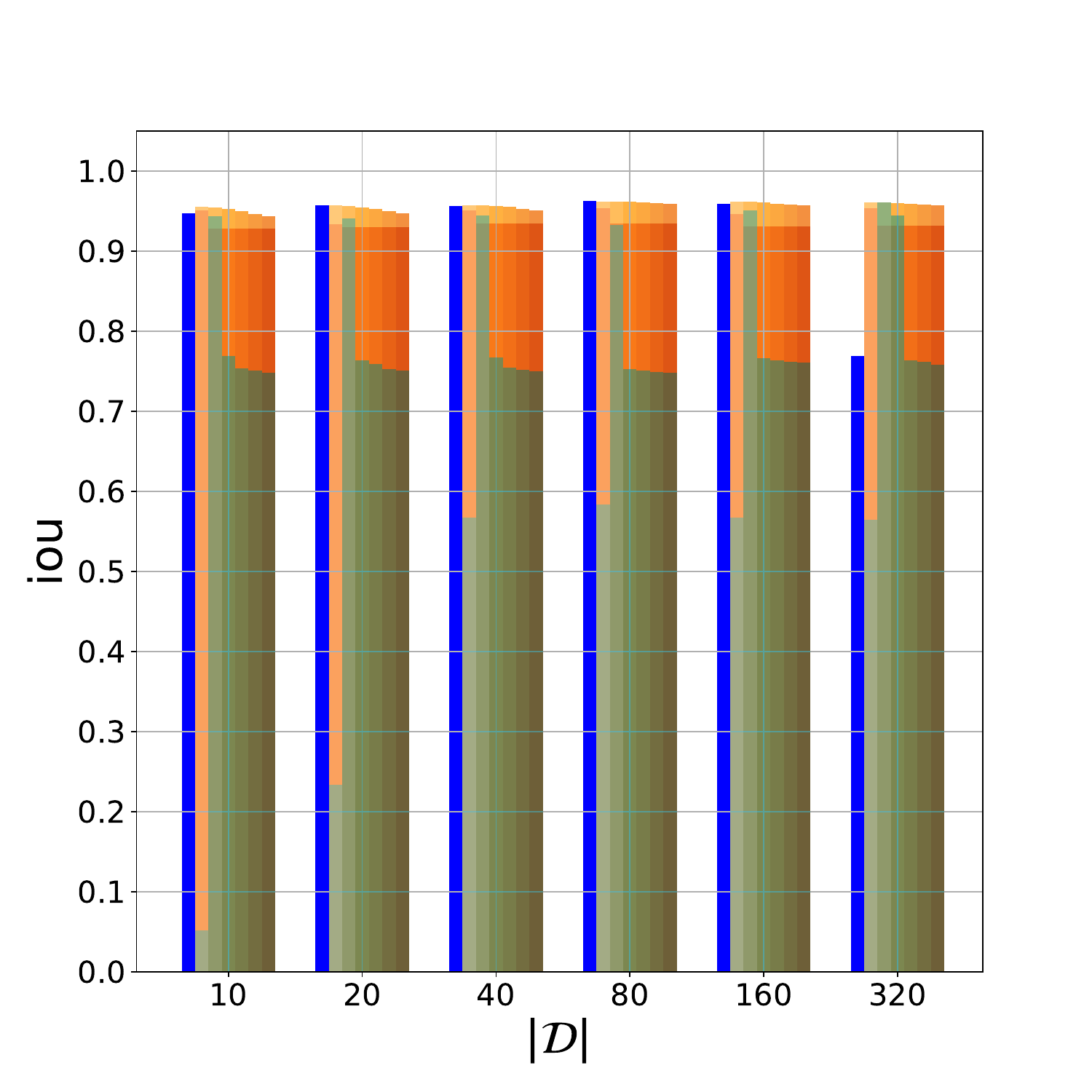}};
        \node (legend) [anchor=south] at ($(noise.north east)+(0,0.5)$) {\includegraphics[clip, trim={0.cm, 0.2cm, 0cm, 0.2cm}, width=0.85\textwidth]{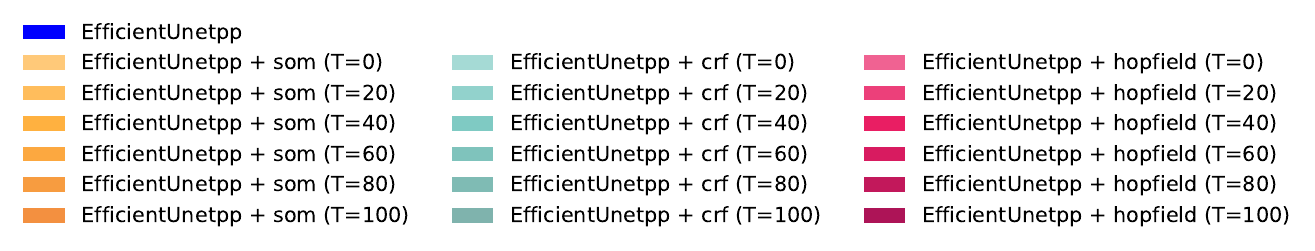}};

        \node (A) at ($(legend.north west)+(0.,0.2)$) {\sffamily{\textbf{A}}};
        \node (B) at ($(noise.north west)+(0.4,0.4)$) {\sffamily{\textbf{B}}};
        \node (C) at ($(examples.north west)+(0.4,0.4)$) {\sffamily{\textbf{C}}};
    \end{tikzpicture}
    \caption{IoU performance, in the \textbf{artificial} setting, of the versions studied: EfficientUnetpp, SOM, CRF, and Hopfield. The color scheme is shown in plot {\sffamily{\textbf{A}}}, which was chosen taking into account how the colors blend. For visual support of this figure, the Hopfield barplot is behind the SOM barplot, which is in turn behind the CRF bars. In {\sffamily{\textbf{B}}}, the impact of noise is shown. And {\sffamily{\textbf{C}}} shows the impact of the number of examples.}
    \label{fig:iou_artificial}
\end{figure}

%% file: results/figures/artificial/pvalues_artificial.tex
\begin{figure}[t]
    \centering
    \begin{tikzpicture}

        \node (center) at (0,0) {};
    
        \node (classes_fig) at ($(center)+(0,0)$) {\includegraphics[clip,trim={0.cm, 1cm, 6cm, 0.5cm},width=0.67\textwidth]{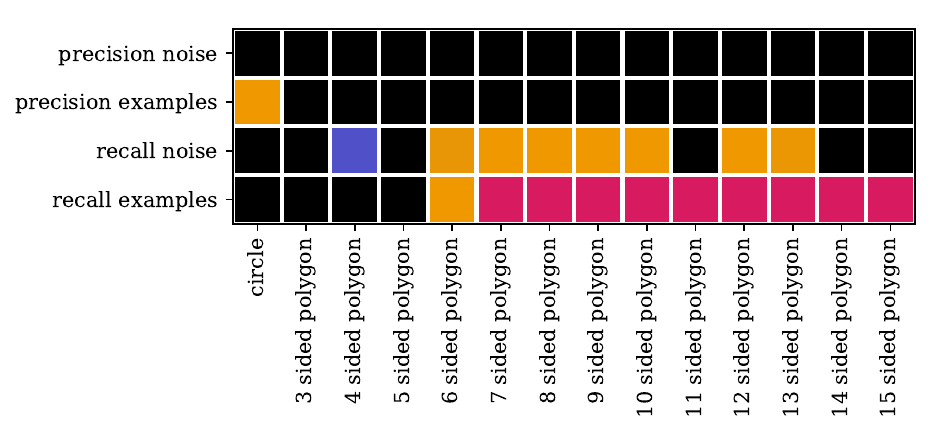}};
    
        \node (colorbar_fig) at ($(classes_fig.west)+(-2,-0.5)$) {\includegraphics[clip,trim={16.5cm, 3cm, 0cm, 0cm},width=0.3\textwidth]{results/figures/artificial/pvalues/pvalue_classes.pdf}};

        \node (noise_and_examples_fig) at ($(classes_fig.south)+(-1.8,-1.8)$) {\includegraphics[width=0.9\textwidth]{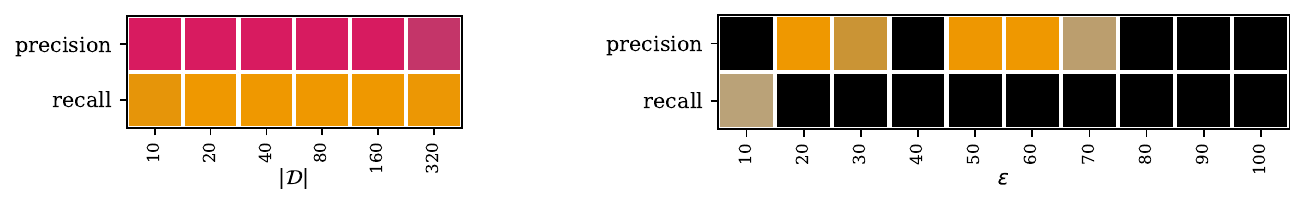}};
    
        \node (classes_xaxis) at ($(classes_fig.south)+(1.1,-0.0)$) {Classes};
    
        \node (A) at ($(colorbar_fig.north)+(-2, 0)$) {\sffamily{\textbf{A}}};
        \node (B) at ($(A.east)+(3.7, 0)$) {\sffamily{\textbf{B}}};
        \node (C) at ($(noise_and_examples_fig.north)+(-6.7, 0)$) {\sffamily{\textbf{C}}};
    
        \node (D) at ($(noise_and_examples_fig.north)+(-0.6, 0)$) {\sffamily{\textbf{D}}};
    
    

    \end{tikzpicture}
    \caption{Experiments on artificial shapes data show a clear superiority for both SOM and Hopfield types of recurrency, across the considered metrics. {\sffamily{\textbf{A}}} shows the color scheme used in this figure. In total, we compared four models and each is associated to a color. If one model has a better IoU/precision/recall with statistical significance ($p$-value$ < 0.05$), then the respective color is associated to that square in the other figures ({\sffamily{\textbf{B}}}, {\sffamily{\textbf{C}}}, and {\sffamily{\textbf{D}}}). {\sffamily{\textbf{B}}} shows the best model that outperformed the others with statistical significance in precision and recall for the \textit{noise} and \textit{number of examples} settings. We do not see any statistical significance in precision, but SOM and Hopfield recurrency show statistical significance in recall.}
    \label{fig:pvalues_artificial}
\end{figure}

%% file: results/figures/cad/iou.tex
\begin{figure}[t]
    \centering
    \begin{tikzpicture}
        \node (center) at (0,0) {};

        \node (noise) at ($(center)+(0,0)$) {\includegraphics[clip, trim={0.cm, 0cm, 0cm, 3cm}, width=0.45\textwidth]{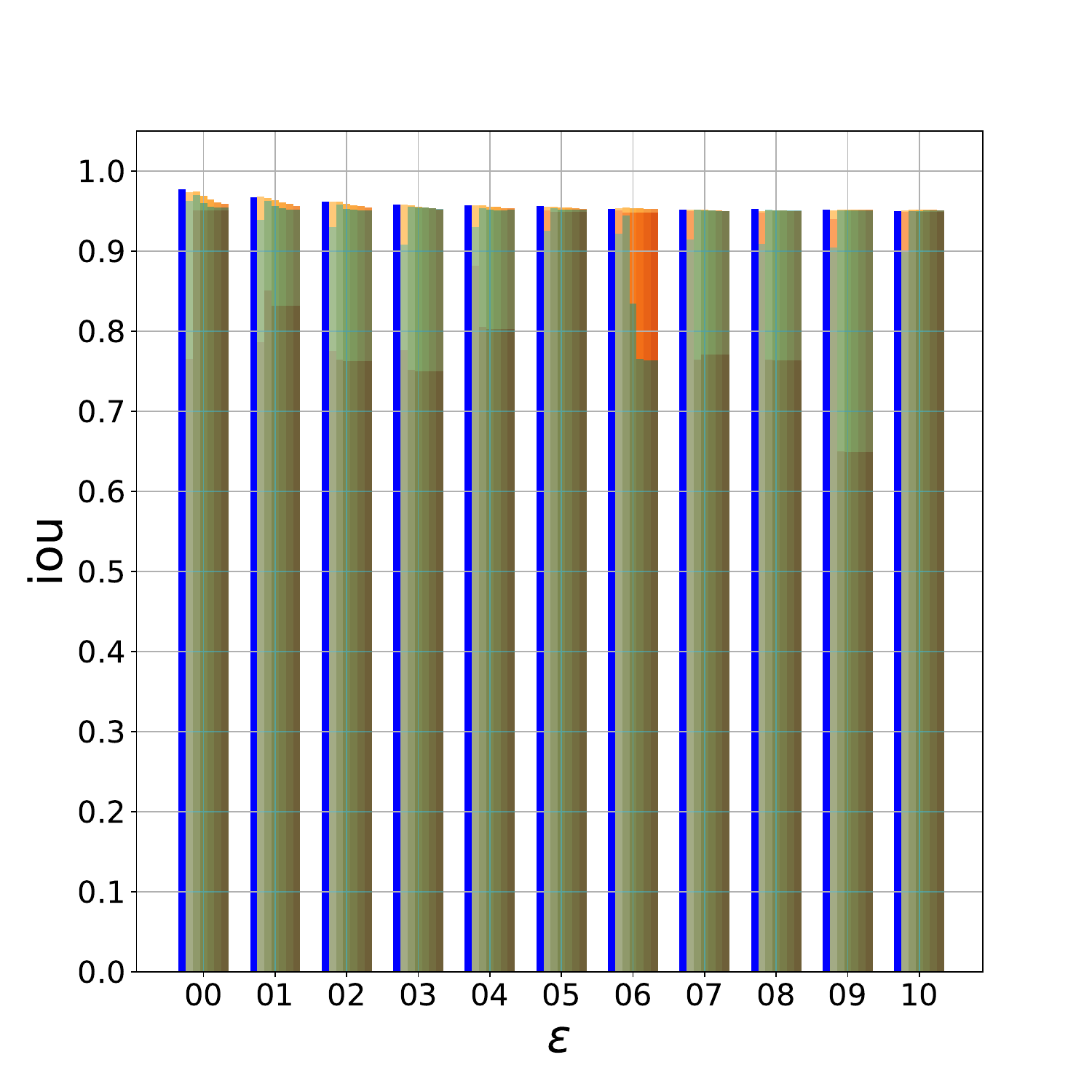}}; 
        \node (examples) [anchor=west] at ($(noise.east)+(0.,0)$) {\includegraphics[clip, trim={0.cm, 0cm, 0cm, 3cm}, width=0.45\textwidth]{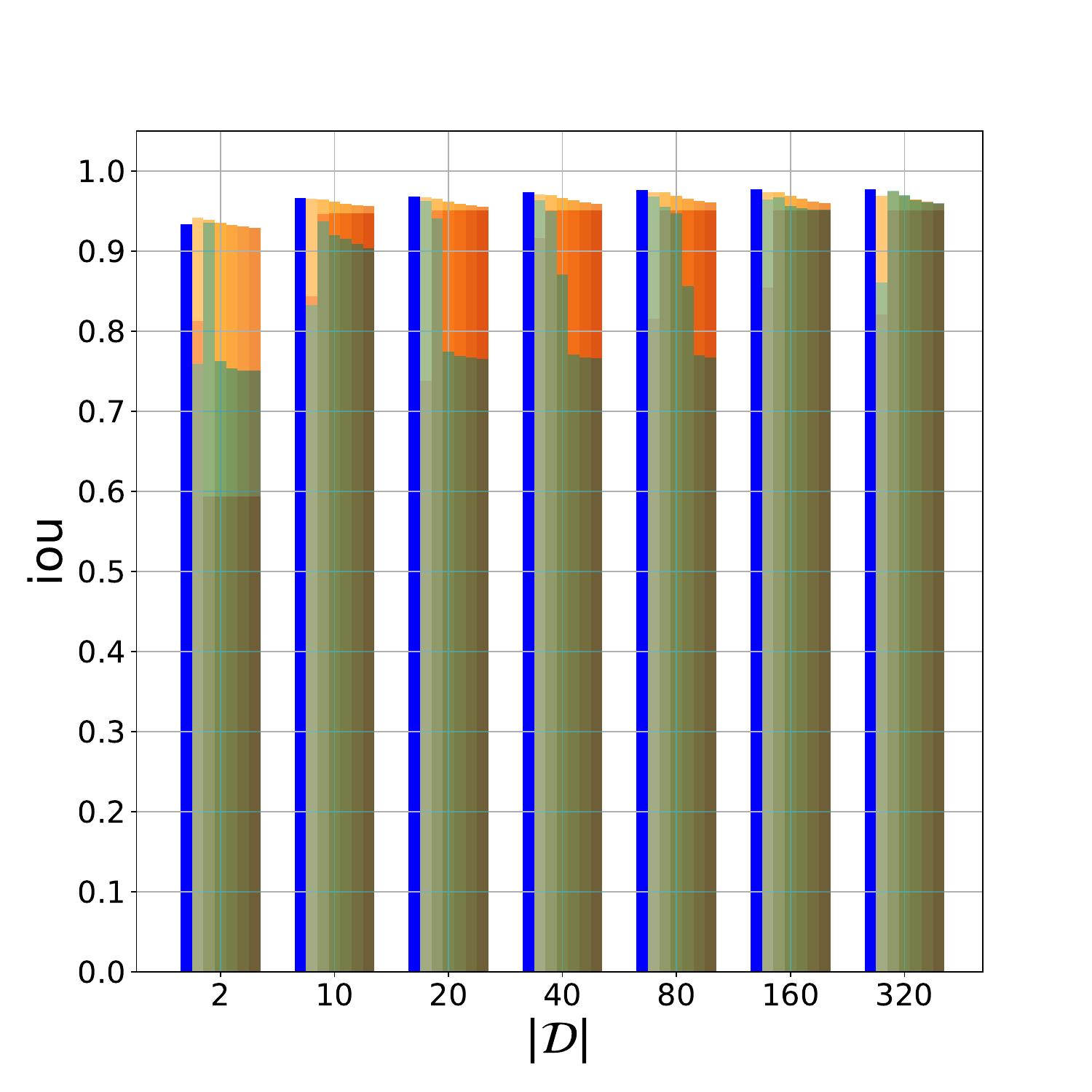}};

        \node (A) at ($(noise.north west)+(0.4,0.4)$) {\sffamily{\textbf{A}}};
        \node (B) at ($(examples.north west)+(0.4,0.4)$) {\sffamily{\textbf{B}}};
    \end{tikzpicture}
    \caption{IoU performance, in the CAD dataset. Figure {\sffamily{\textbf{A}}} shows the impact of different values in noise and figure {\sffamily{\textbf{B}}} shows the impact of the training set size.}
    \label{fig:iou_cad}
\end{figure}
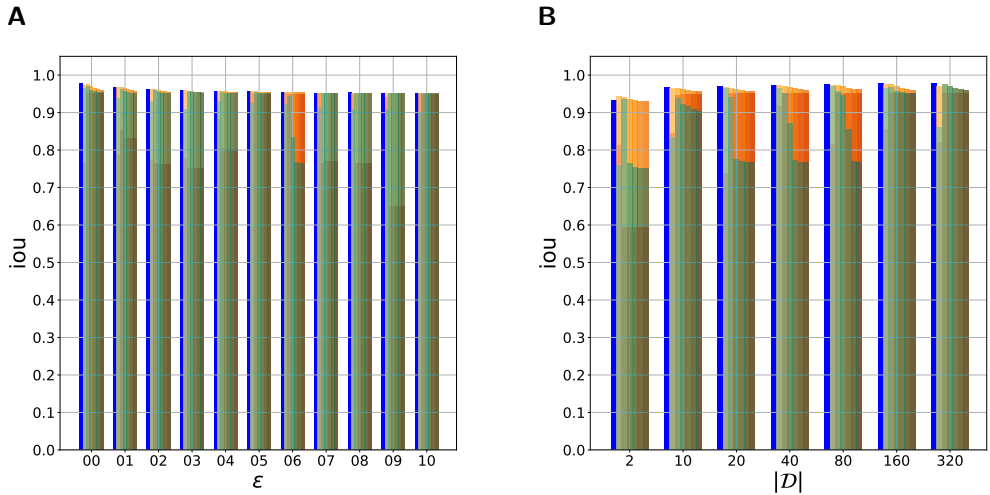

%% file: results/figures/cad/pvalues_cad.tex
\begin{figure}[t]
    \centering
    \begin{tikzpicture}
        \node (center) at (0,0) {};

        
        \node (classes) at ($(center)+(0,0)$) {\includegraphics[clip,trim={0cm, 1cm, 11cm, 0.5cm},width=0.25\textwidth]{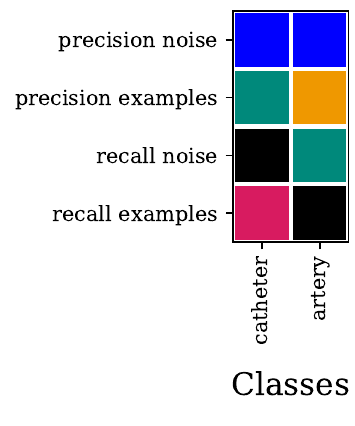}};

        \node (noise) [anchor=north west] at ($(classes.north east)+(0,0)$) {\includegraphics[clip,trim={10cm, 0cm, 0cm, 0cm},width=0.6\textwidth]{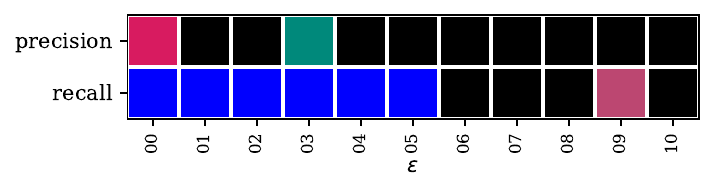}};

        \node (examples) [anchor=north] at ($(noise.south)+(0,0.4)$) {\includegraphics[clip,trim={0cm, 0cm, 14cm, 0.3cm},width=0.4\textwidth]{results/figures/cad/pvalues/pvalue_noise_and_examples.pdf}};

        \node (A) at ($(classes.north west)+(0,0)$) {{\sffamily{\textbf{A}}}};
        \node (B) at ($(noise.north west)+(0.5,0)$) {{\sffamily{\textbf{B}}}};
        \node (C) at ($(examples.north west)+(-1.,0)$) {{\sffamily{\textbf{C}}}};
    \end{tikzpicture}
    \caption{Experiments on CAD data show that all models did not handle well noise nor limited sample size. However, the statistical significance tests show a tendency for recurrent models to outperform the feed-forward version even though the difference was bigger in the artificial setting. Figure {\sffamily{\textbf{A}}} reports the statistical significance for the classes, \textit{catheter} and \textit{artery}. Figure {\sffamily{\textbf{B}}} reports the same for the values of noise tested. Figure {\sffamily{\textbf{C}}} reports the same for different sample sizes.}
    \label{fig:pvalues_cad}
\end{figure}

%% file: discussion/discussion.tex
\section{Discussion}\label{section:discussion}


\input{discussion/artificial}

\input{discussion/cad}

%% file: discussion/artificial.tex
In this study, we compare the feed forward version of the EfficientUnetpp \cite{lourencco2021encoder} and the same network appended with energy based recurrency, explained in section \ref{section:recurrent_models}. In general, models that outperformed the others with statistical significance have, in their majority, a type of recurrency (see figure \ref{fig:pvalues_artificial}).

\textbf{Self-organizing maps benefit segmentation in noisy settings.} Our results show a clear advantage for the self-organizing recurrency in noisy settings (see figures \ref{fig:iou_artificial} {\sffamily{B}} and \ref{fig:pvalues_artificial} {\sffamily{D}}). We hypothesize that this advantage is due to two special characteristics of our SOM implementation: \textit{propagation of certainty} and \textit{component separation using an image filter}. Propagation uncertainty seems to ease the work for the EfficientUnepp, because theoretically the segmentation belief map needs only a small set of pixels with high certainty on the correct class and the self-organization will propagate that certainty to neighbour pixels. The latter promotes sparse belief maps (few pixels with a high norm, according to equation \ref{equation:high_response_message_passing}). The other trait, separating image components using a filter, ensures that we do high-certainty beliefs do not contaminate sections of the image that belong to a different class. At the cost of under higher noise settings, we may stop beliefs from propagating at all, making the gain of self-organization smaller. Nonetheless, in preliminary experiments, we found this feature necessary for the whole algorithm to be stable and reach energy local minima that are close to the ground truth.


\textbf{Noise and limited sample settings impact recall.} Our results suggest that both noisy and limited sample settings impact the recall of all models. Some more than others. Whilst this is more noticeable in the IoU of the noisy analysis, it is also present when we assess the recall of each class for the limited sample settings. \textbf{Recurrency handles limited sample settings better than feed-forward.} Though figure \ref{fig:iou_artificial} {\sffamily{\textbf{C}}} does not show much difference of IoU between the models compared, when assessing the recall of each class (see figures \ref{fig:pvalues_artificial} {\sffamily{\textbf{B}}} and {\sffamily{\textbf{C}}}), we are able to observe an impact on recall. This tells us that models have trouble classifying the correct class in this setting.

\textbf{Self-organizing maps are good for noise, while Hopfield networks are best suited for limited sample settings.} Our results suggest that the feed-forward version of the EfficientUnetpp is not the best for the settings considered. However, there is not a consensus of what is the best type of recurrency. In noisy settings, self-organizing maps are superior. In limited sample settings, Hopfield networks had the best performance. For the noisy settings, we believe that the ability of the SOM recurrency to propagate beliefs that have high certainty benefits the model. While the filter cuts off the propagation of beliefs to parts of the image where those beliefs should not be propagated, the propagation inside each graph component improves the segmentation quality. And since images were corrupted using Gaussian noise, the memory retrieval mechanism deteriorates when the noise is too much. Which is validated by the very poor IoU metric of Hopfield recurrency. But one of the reasons self-organization benefits the EfficientUnetpp in noisy settings is because the feed-forward model was able to learn internal representation so that it outputs a good belief map. When the number of examples is low, self-organization is no longer advantageous. Instead, memory retrieval is more useful in this setting. When the model has stored memories that correspond to specific classes, the memory size compensates for the lack of size of the training set (although the first is built with the second). This explains how suitable the memory retrieval mechanism is for limited sample settings.

%% file: discussion/cad.tex
\textbf{Medical imaging data has a high amount of noise.} The medical imaging data, CAD data, contains high amounts of noise. It not only has noise in the input image, but it also has some inconsistencies on the masks that were manually labeled. This is natural in these types of datasets, where the amount of samples is limited and the association between input and output is hard to uncover. The performance of all models deteriorates when we add even more noise and for the SOM and CRF cases, as the number of iterations increases, the performance worsens. The random nature of blood arteries makes the memory retrieval mechanism of the Hopfield network not suitable because in true random phenomena, nothing is the same. In other words, there are no common patterns. This shows that the task is difficult and the recurrency models employed are not suitable for this task. \textbf{Models struggle with small frequency classes, such as catheter and artery.} The focal loss, explained in section \ref{subsection:metrics}, emphasizes errors for classes that appear in a small area of the image. However, both forward and recurrent models struggle to identify these correctly when the noise increases. Specifically, SOM and CRF recurrencies appear to worsen their precision with the number of iterations, meaning that previously correct pixels (correctly labeled by the forward version of the network) change to a different class because of the recurrent process. This suggests that both competitive and relation types of recurrency are not fit for problems that have consistently small frequency classes.

%% file: related_work/related_work.tex
\section{Related work}\label{section:related_work}

\noindent\textbf{Belief propagation recurrency.} \input{related_work/relation/relation}

\textbf{Associative memories.} \input{related_work/associative/associative}

\textbf{Non energy based recurrency.} \input{related_work/non_energy_based/non_energy_based}

There is a general consensus that we need recurrency in models in order to emulate neuronal processes, such as memory retrieval and visual processing. There have been extraordinary findings that suggest recurrency is fundamental in uncertainty quantification and reaction time processes. However, these models do not follow an energy function. In this study, we applied multiple energy-based types of recurrency to assess how they perform in noise and limited sample settings. We plan to release the source code soon in a github repository.

%% file: related_work/relation/relation.tex
\input{related_work/relation/kuck2020belief} 
\input{related_work/relation/kohonen1982self} \input{related_work/relation/zheng2015conditional}

%% file: related_work/relation/kuck2020belief.tex
One very well-known type of recurrency is belief propagation. This type of model consists on a graph of factors and nodes. Jointly these define a joint probability distribution. This is exactly the formulation of a conditional random field \cite{zheng2015conditional}. One can define the model as a graph neural network that performs message passing and whose messages are themselves gradients of an energy function. Some call this technique belief propagation networks \cite{kuck2020belief, satorras2021neural} .

%% file: related_work/relation/kohonen1982self.tex
Self-organizing recurrency can be a type of belief propagation recurrency, with the addition of a competitive component. The name of self-organization was introduced by \citet{kohonen1982self}, but the update rule is also used in particle optimization \cite{kennedy1995particle}.

%% file: related_work/relation/zheng2015conditional.tex
\citet{zheng2015conditional} made a breakthrough in image segmentation at the time, by introducing conditional random fields on the output of a neural network. A conditional random field can learn the relations between labels, i.e. how likely is to see this class next to the other one a.k.a. compatibility. This methodology also fits into belief propagation, since we are propagating beliefs based on the general compatibility of those beliefs.

%% file: related_work/associative/associative.tex
\input{related_work/associative/hopfield} \input{related_work/associative/grushin2023training}

%% file: related_work/associative/hopfield.tex
Hopfield networks were introduced by \citet{hopfield1982neural}, at the time referred to as associative memory networks. These have recently attracted attention of researchers, with significative advances made by \citet{krotov2016dense}, \citet{ramsauer2020hopfield} and \citet{hoover2024energy}.

%% file: related_work/associative/grushin2023training.tex
\citet{grushin2023training} discusses the use of recurrency with internal states. The input is mapped to the hidden state as $W \cdot x$ and the hidden state is also mapped to the input with the same weights $W^\top \cdot x$. One can think of $W$ as the memories of the model. The novelty of this study is that it evolves the internal states in time. \citet{krotov2016dense} proposed an energy function that is capable of a much higher memory capacity than the original version \cite{hopfield1982neural}, then \citet{demircigil2017model} showed that representing the energy function with a \textit{LogSumExp} has an even larger capacity. These findings later propelled the community to study the connection between the memory retrieval mechanism with attention \cite{vaswani2017attention}. Recently,  \citet{krotov2021hierarchical} proposed a theory that uses multiple layers with internal states and each layer send information to the next and the previous one. However, it remains an open question on how we can train this type of model.

%% file: related_work/non_energy_based/non_energy_based.tex
 We refer to this type of recurrency as the techniques used in \cite{elman1990finding}. These networks have a hidden state, $\mathbf{h}$ and an input, $\mathbf{x}$, and typically have the form $W\cdot \mathbf{x}+U\cdot \mathbf{h}$. However, this type of recurrency may not be stable nor reach an equilibrium. Nonetheless, studies have shown that using this recurrency allows us to better understand some human cognitive processes. For instance, \input{related_work/non_energy_based/spoerer2020recurrent} \input{related_work/non_energy_based/goetschalckx2024computing}

%% file: related_work/non_energy_based/spoerer2020recurrent.tex
\citet{spoerer2020recurrent} developed a recurrent neural network, inspired on the fact that the visual cortex contains feedback connections that are activated during the process of object recognition. The results suggest that as you let the system evolve in time (more recurrent/feedback iterations) the higher the accuracy of the neural network. This observation made the authors hypothesize a link between additional iterations and human reaction times. With an additional experiment involving recorded human reaction times to images of the same dataset that the models were trained on, the authors were able to support this hypothesis.

%% file: related_work/non_energy_based/goetschalckx2024computing.tex
\citet{goetschalckx2024computing} delves further into analyzing the human reaction time using a recurrent neural network. Their model is able to quantify uncertainty on a task where the goal is to detect if two points are in the same object. The model propagates flows from those points to the rest of image. The weights of the recurrent neural network detect when we are at the boundary of an object. Further, it appears that the model is stable due to being optimized with the contractor recurrent back propagation algorithm \cite{linsley2005stable}. Unfortunately, the use of this algorithm remains limited to the community\footnote{Source code is not available on \href{https://github.com/c-rbp/c-rbp}{github}}.

%% file: conclusions/conclusions.tex
\section{Conclusion}\label{section:conclusion}


Our results did not validate our original hypothesis that recurrent neural networks should outperform feed-forward networks in the settings we studied. Although we found that recurrency improves segmentation in our artificial dataset experiments, this improvement was not observed for the medical imaging data, and adding recurrency actually worsens the initial segmentation 
computed by the base model. In terms of noise, it seems that self-organizing recurrency is slightly better suited for this setting, while there was no consensus in limited sample settings.

Energy-based recurrency has the advantage of being stable and it is perfect if the energy function is well suited for the task. However, it is not always the case for the latter as designing a good energy function is very difficult. Nonetheless, we hypothesize that incorporating the focal loss in the energy function (such that it is unsupervised) might be advantageous. Another type of recurrency worth exploring is the one used in \citet{spoerer2020recurrent} and \citet{goetschalckx2024computing}.

\textbf{Future work.} While self-organizing recurrency performed well in noisy settings, Hopfield networks were more effective with limited samples. However, neither model achieved consistent performance across both conditions. Future work could look at hybrid models that leverage both SOM and Hopfield properties or explore ensemble methods that dynamically apply the best-performing architecture based on data characteristics.

%% file: results/noise/artificial.tex
\begin{figure}[ht]
    \centering

    \begin{subfigure}[b]{0.24\textwidth}
        \begin{tikzpicture}
            \node (figure) at (0,0) {\includegraphics[width=\textwidth]{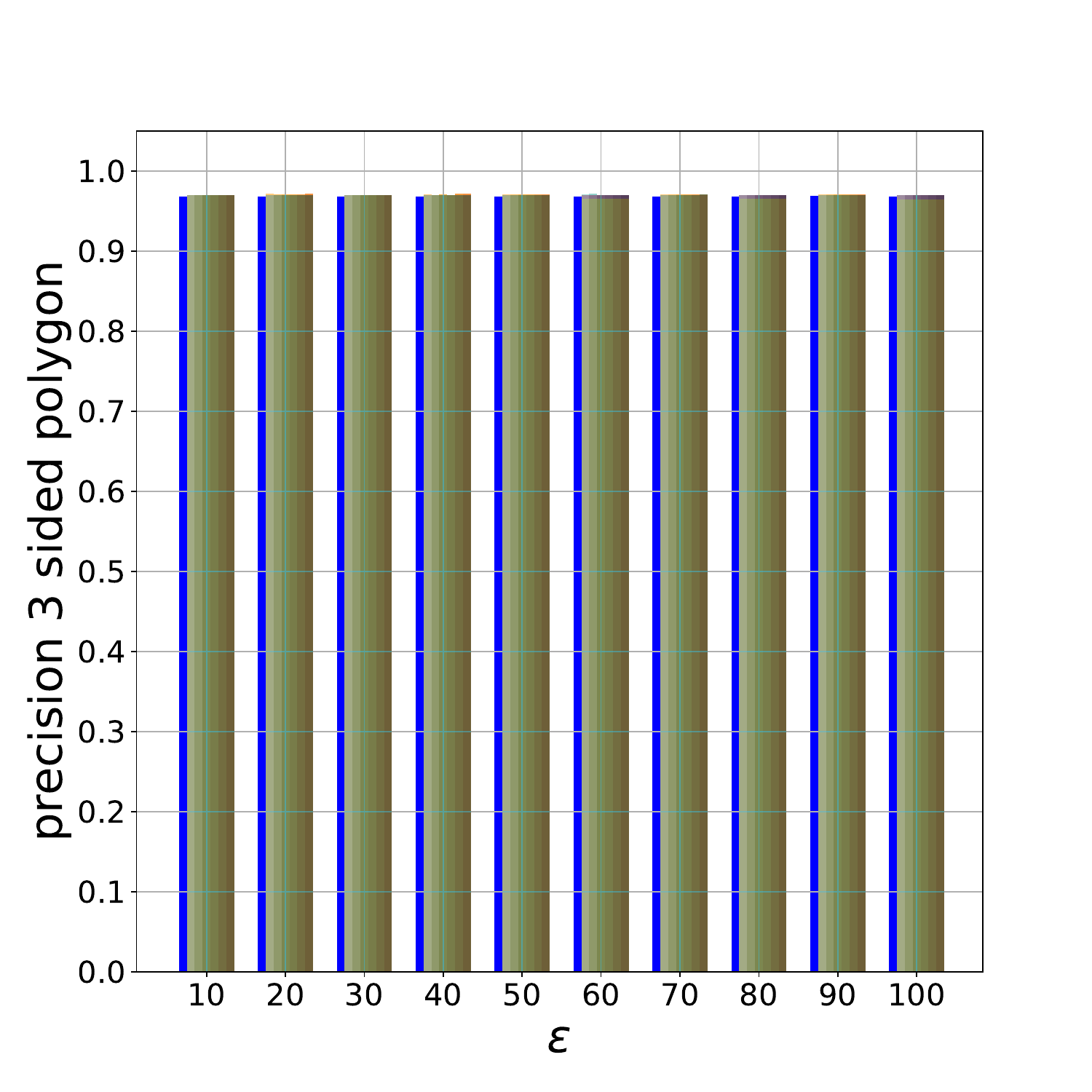}};
        \end{tikzpicture}
    \end{subfigure}
    \begin{subfigure}[b]{0.24\textwidth}
        \begin{tikzpicture}
            \node (figure) at (0,0) {\includegraphics[width=\textwidth]{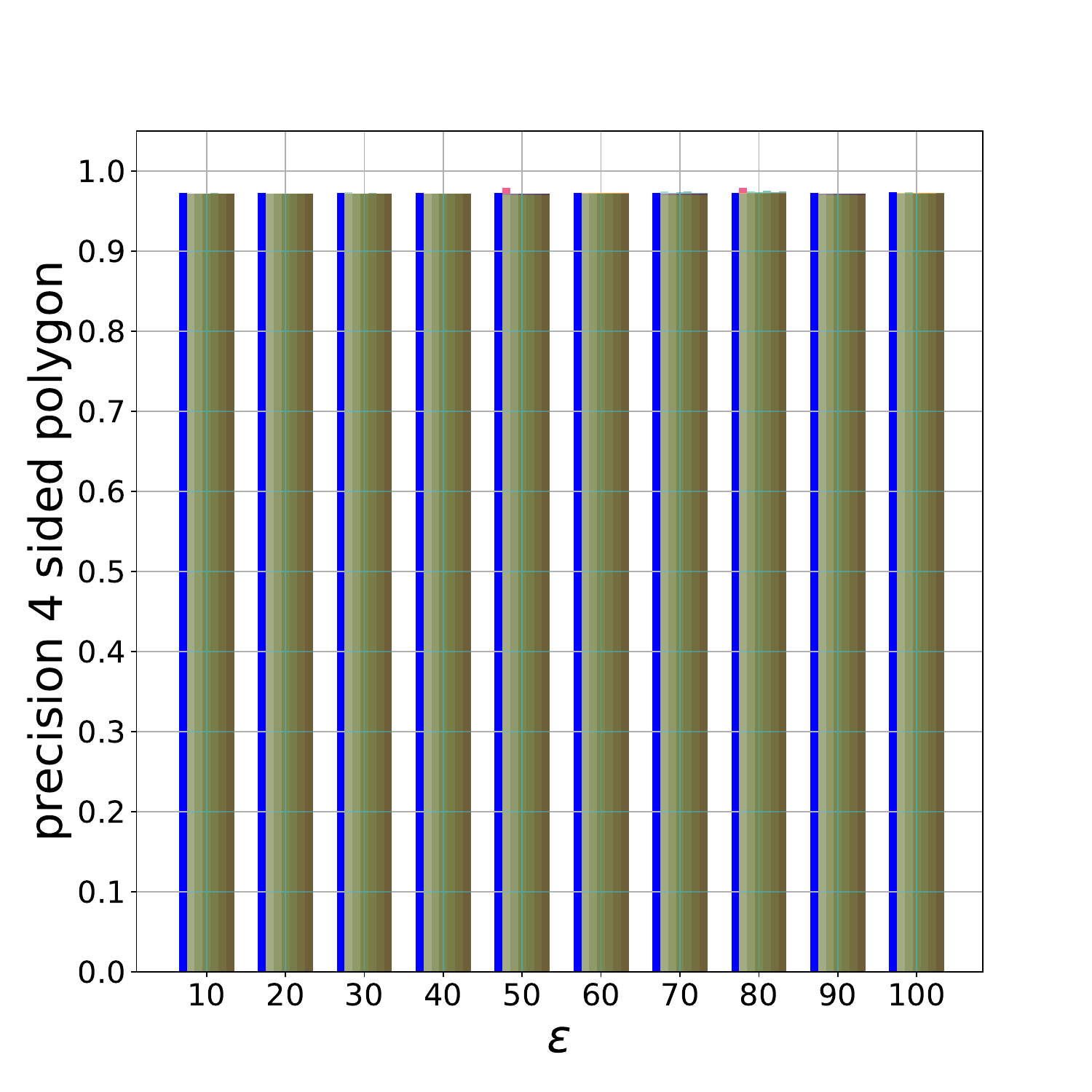}};
        \end{tikzpicture}
    \end{subfigure}
    \begin{subfigure}[b]{0.24\textwidth}
        \begin{tikzpicture}
            \node (figure) at (0,0) {\includegraphics[width=\textwidth]{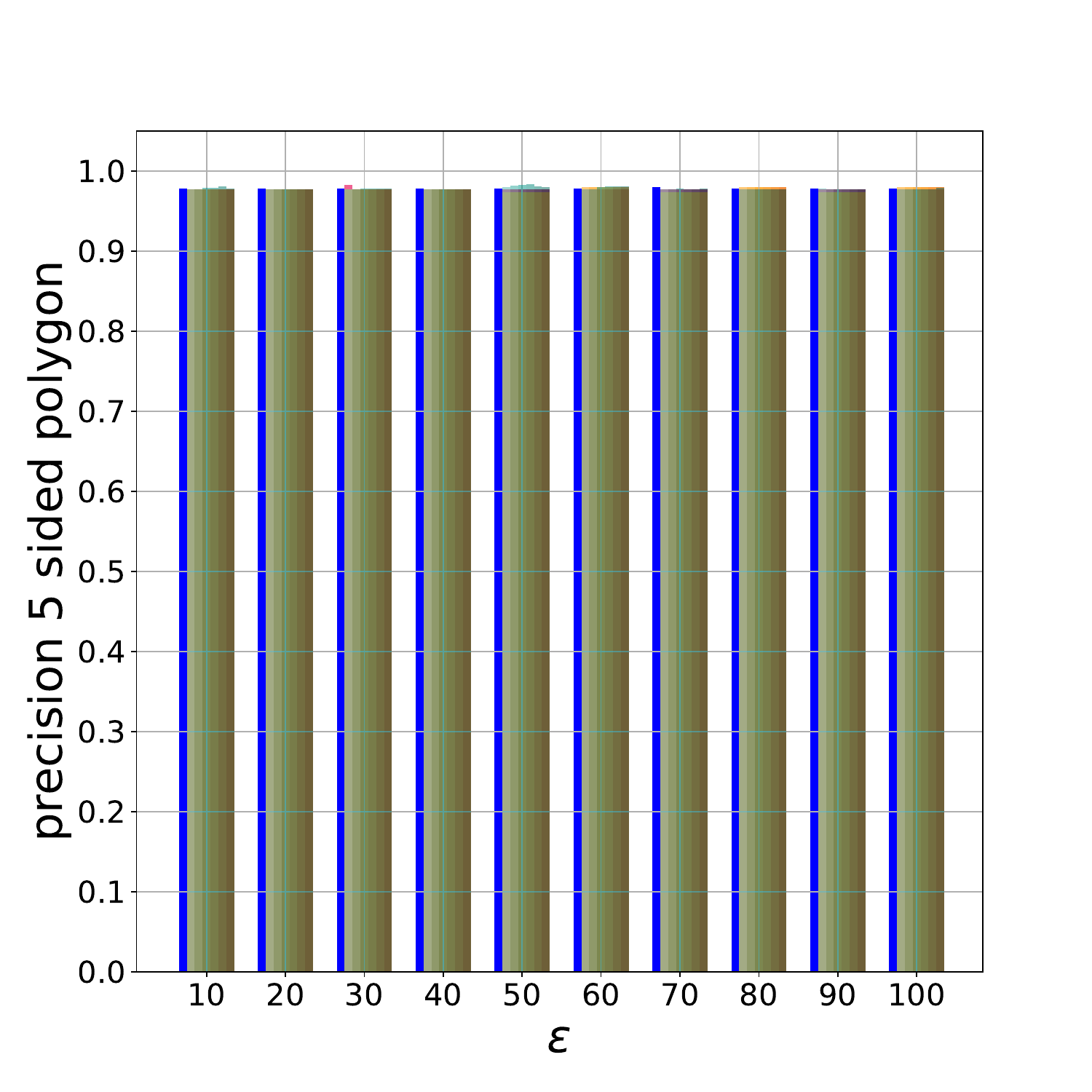}};
        \end{tikzpicture}
    \end{subfigure}
    \begin{subfigure}[b]{0.24\textwidth}
        \begin{tikzpicture}
            \node (figure) at (0,0) {\includegraphics[width=\textwidth]{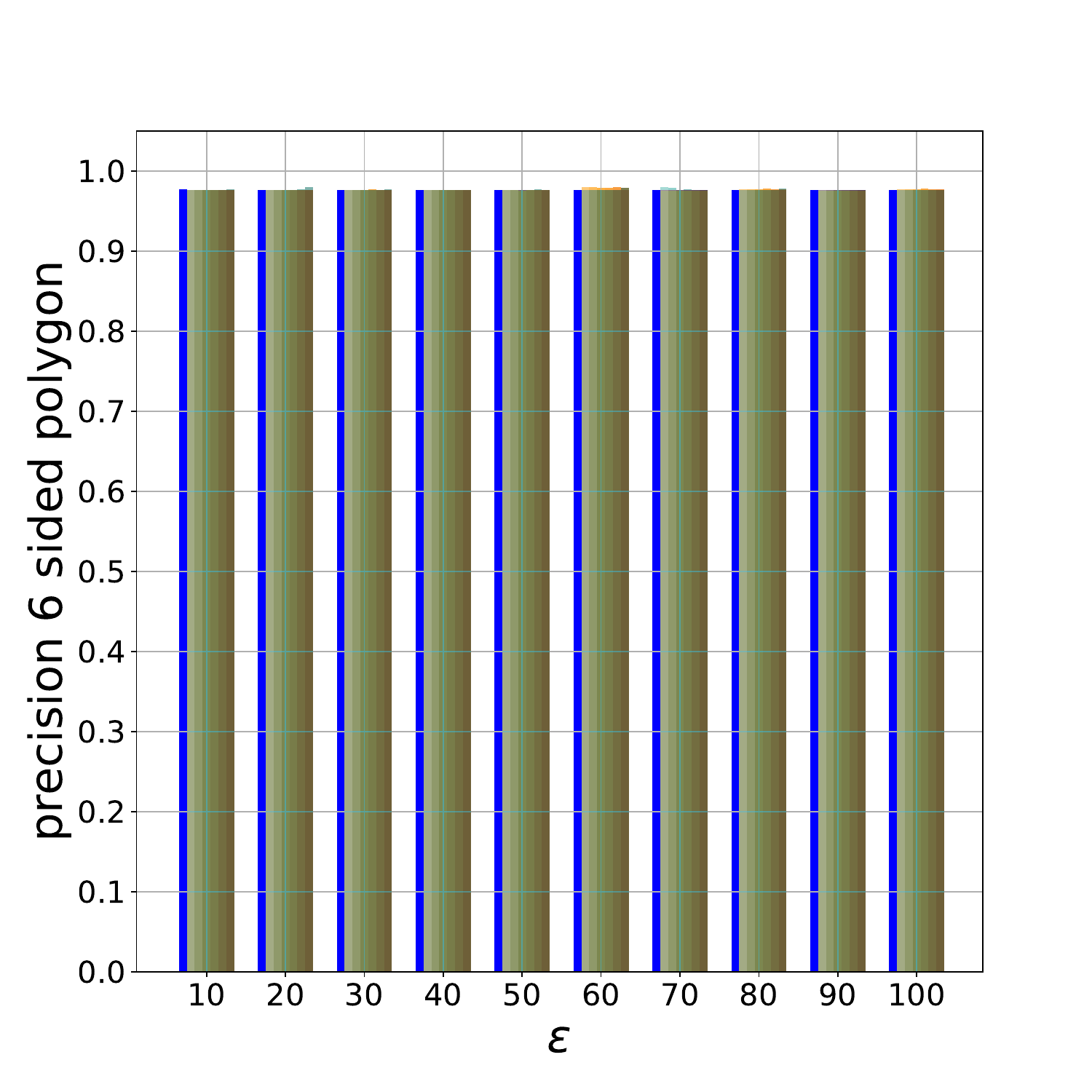}};
        \end{tikzpicture}
    \end{subfigure}

    \begin{subfigure}[b]{0.24\textwidth}
        \begin{tikzpicture}
            \node (figure) at (0,0) {\includegraphics[width=\textwidth]{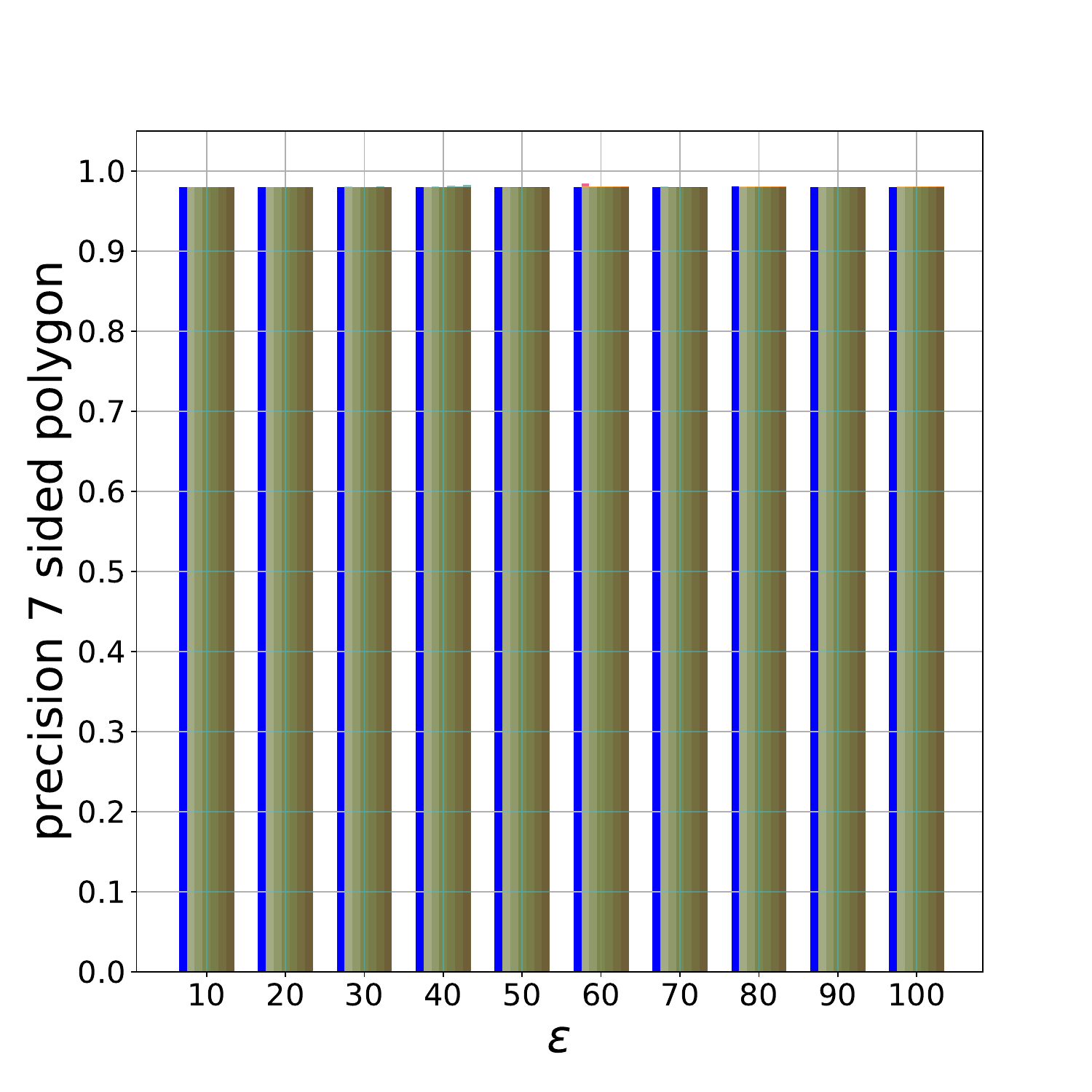}};
        \end{tikzpicture}
    \end{subfigure}
    \begin{subfigure}[b]{0.24\textwidth}
        \begin{tikzpicture}
            \node (figure) at (0,0) {\includegraphics[width=\textwidth]{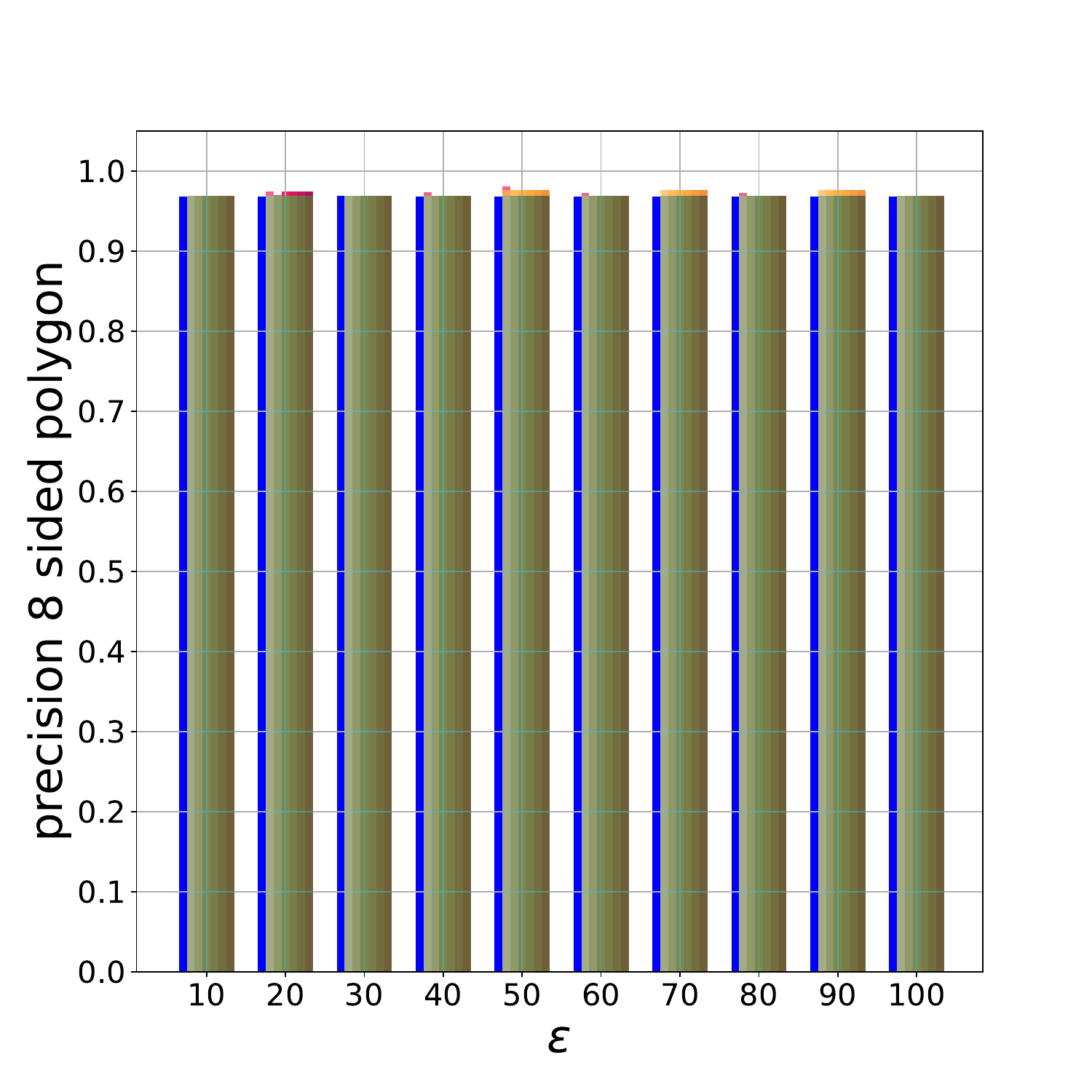}};
        \end{tikzpicture}
    \end{subfigure}
    \begin{subfigure}[b]{0.24\textwidth}
        \begin{tikzpicture}
            \node (figure) at (0,0) {\includegraphics[width=\textwidth]{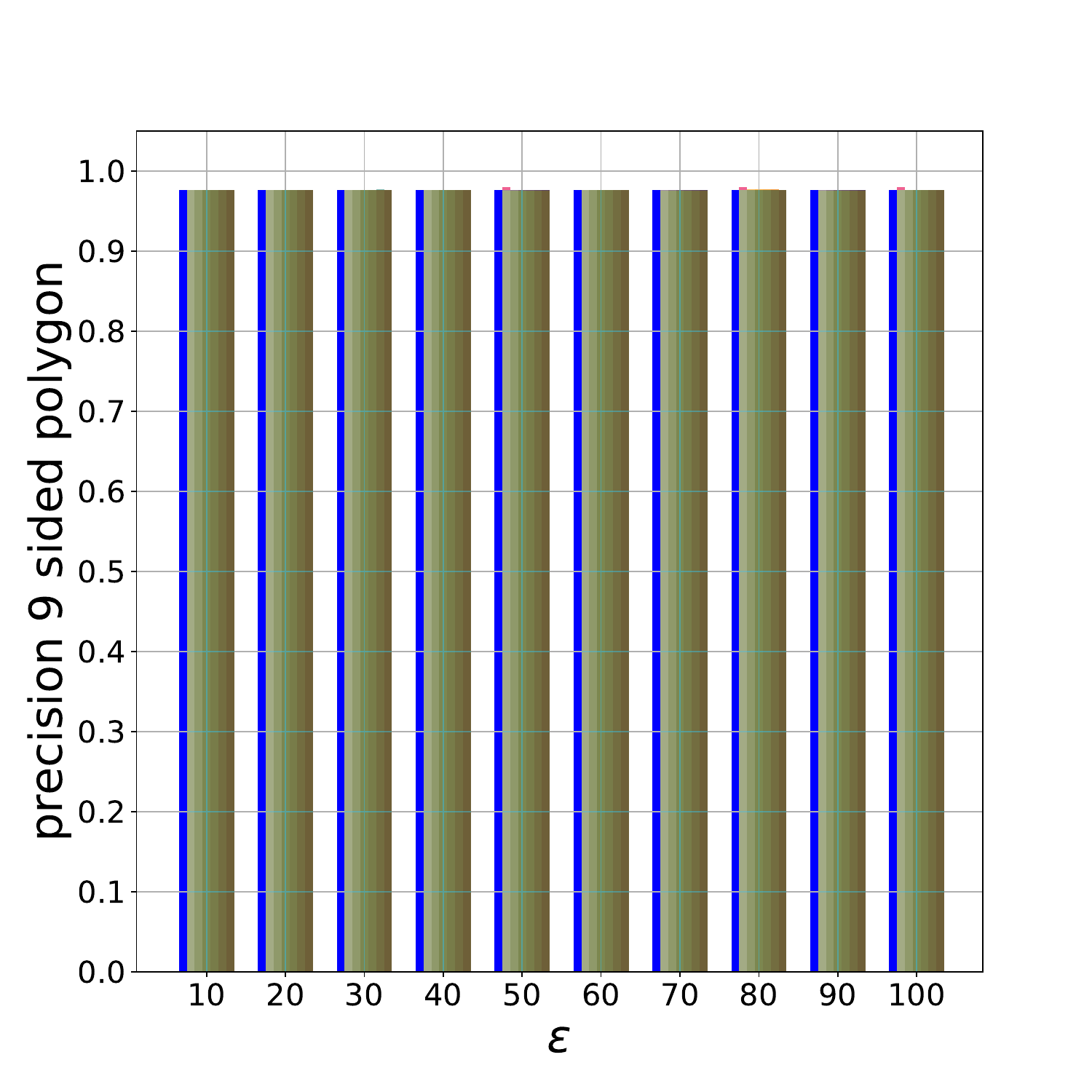}};
        \end{tikzpicture}
    \end{subfigure}
    \begin{subfigure}[b]{0.24\textwidth}
        \begin{tikzpicture}
            \node (figure) at (0,0) {\includegraphics[width=\textwidth]{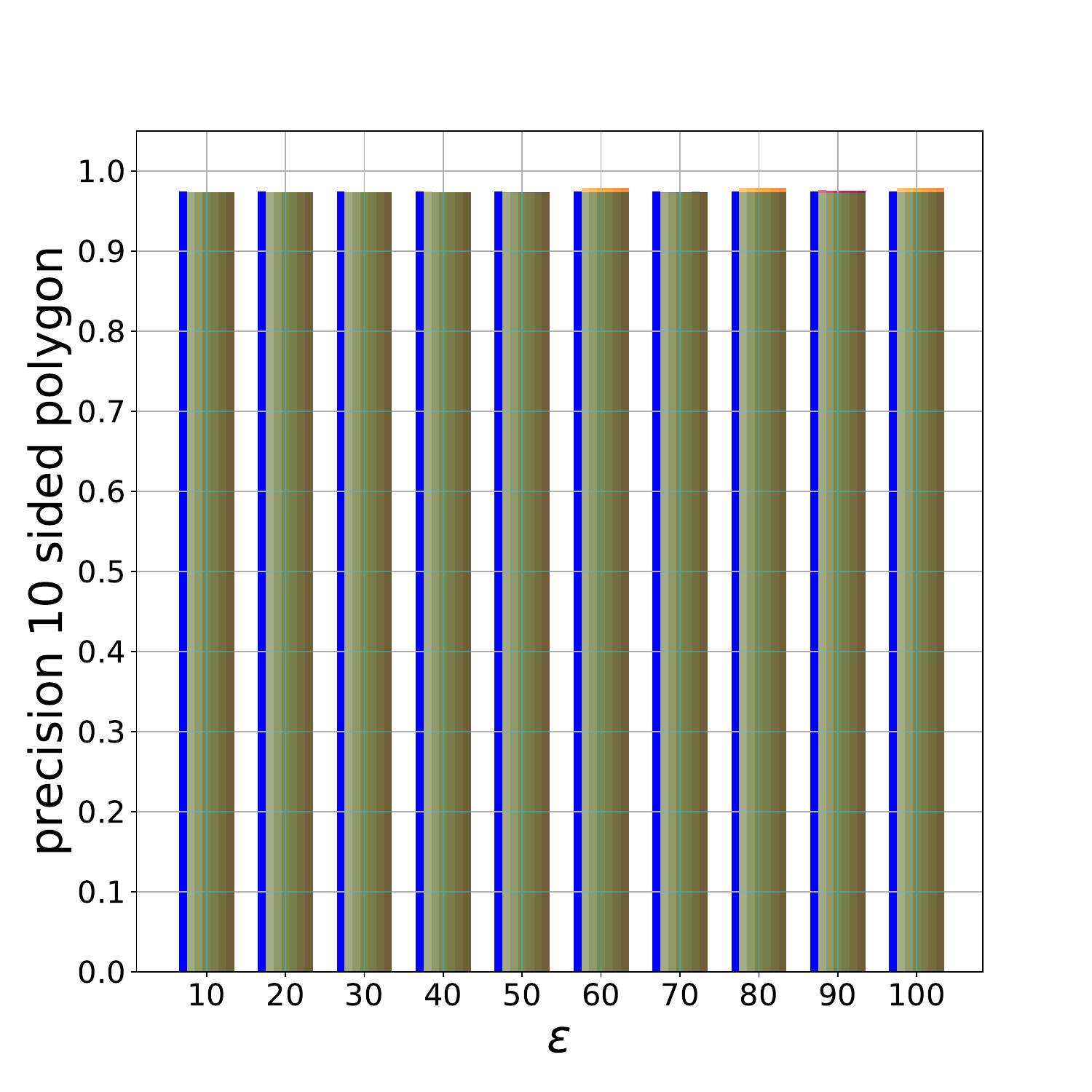}};
        \end{tikzpicture}
    \end{subfigure}

    \begin{subfigure}[b]{0.24\textwidth}
        \begin{tikzpicture}
            \node (figure) at (0,0) {\includegraphics[width=\textwidth]{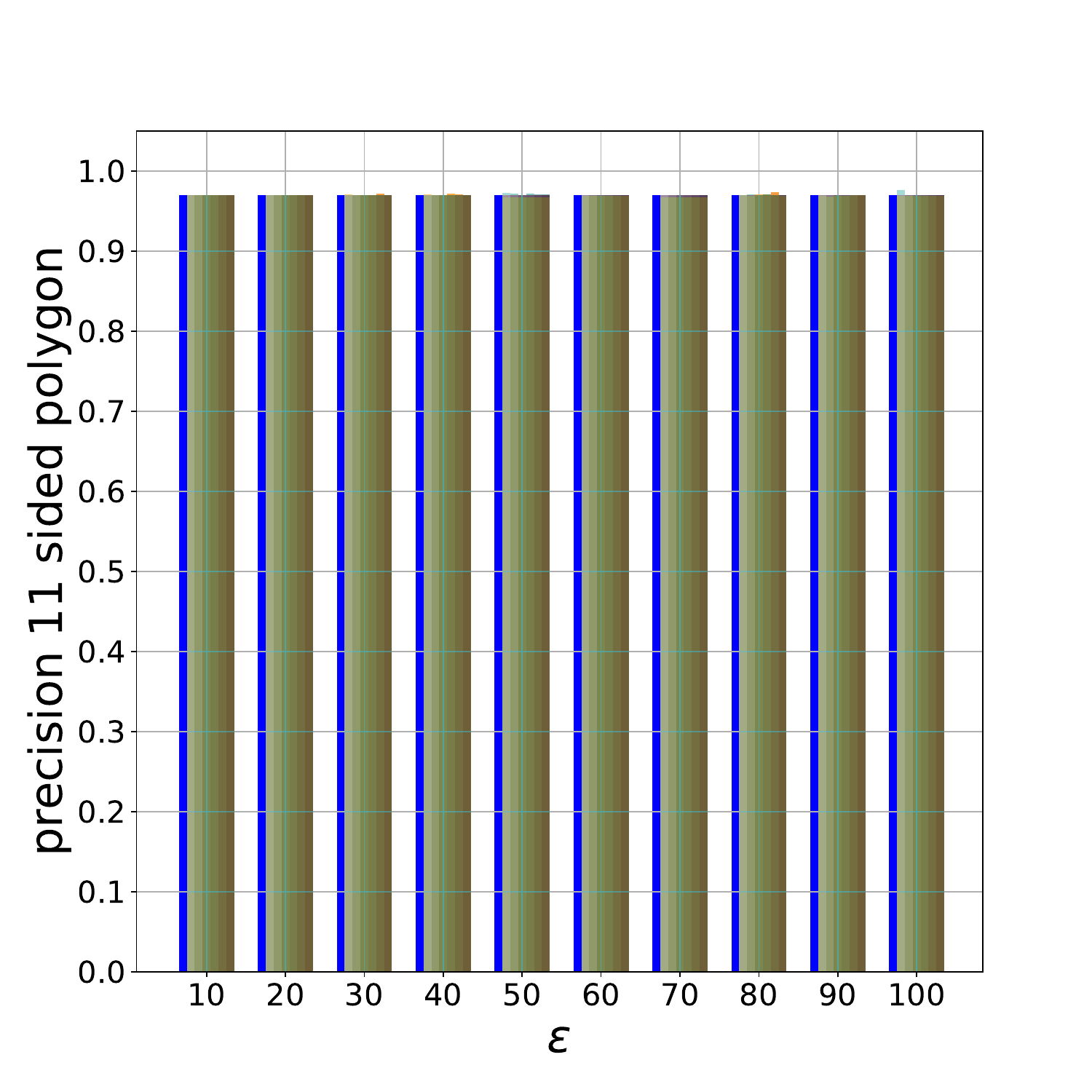}};
        \end{tikzpicture}
    \end{subfigure}
    \begin{subfigure}[b]{0.24\textwidth}
        \begin{tikzpicture}
            \node (figure) at (0,0) {\includegraphics[width=\textwidth]{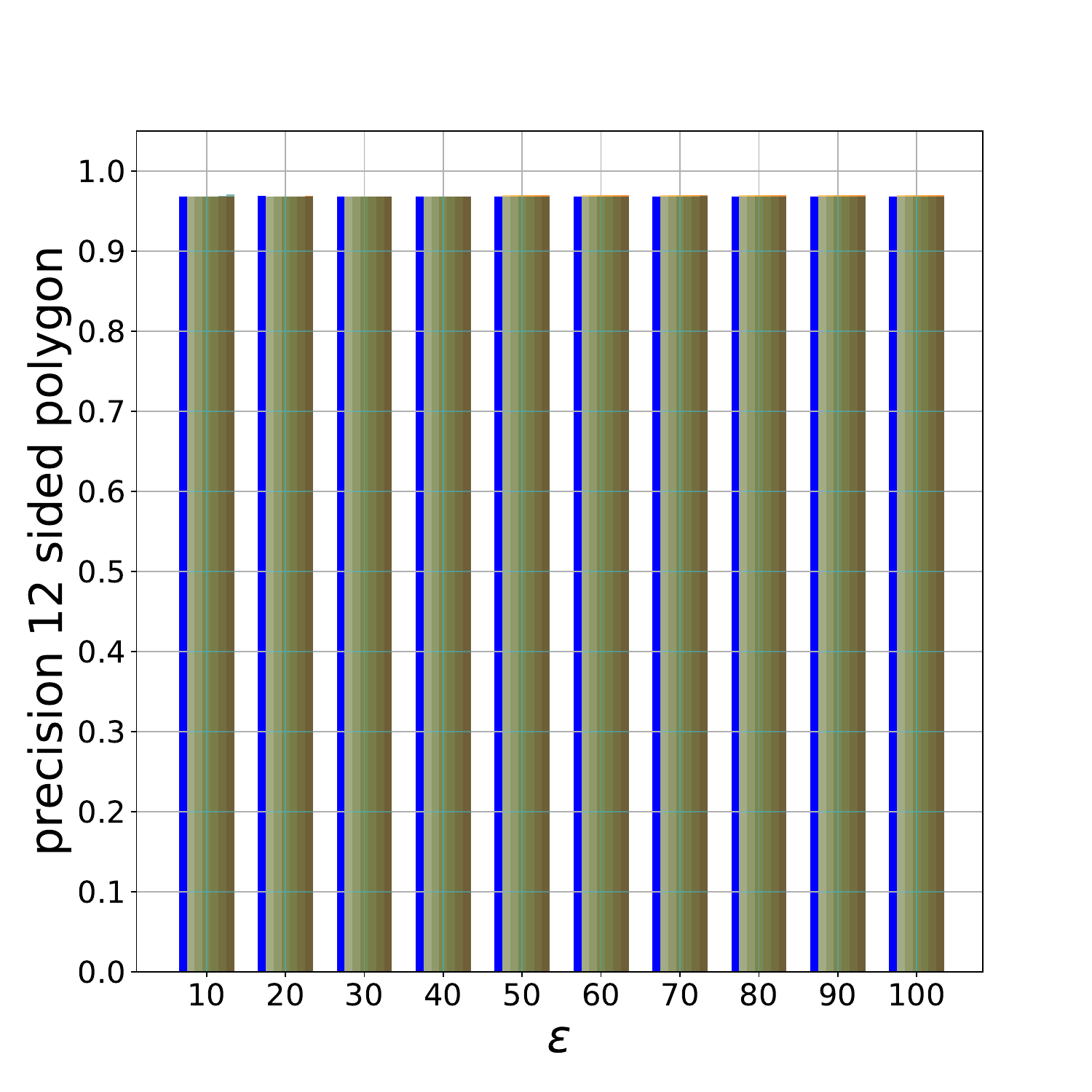}};
        \end{tikzpicture}
    \end{subfigure}
    \begin{subfigure}[b]{0.24\textwidth}
        \begin{tikzpicture}
            \node (figure) at (0,0) {\includegraphics[width=\textwidth]{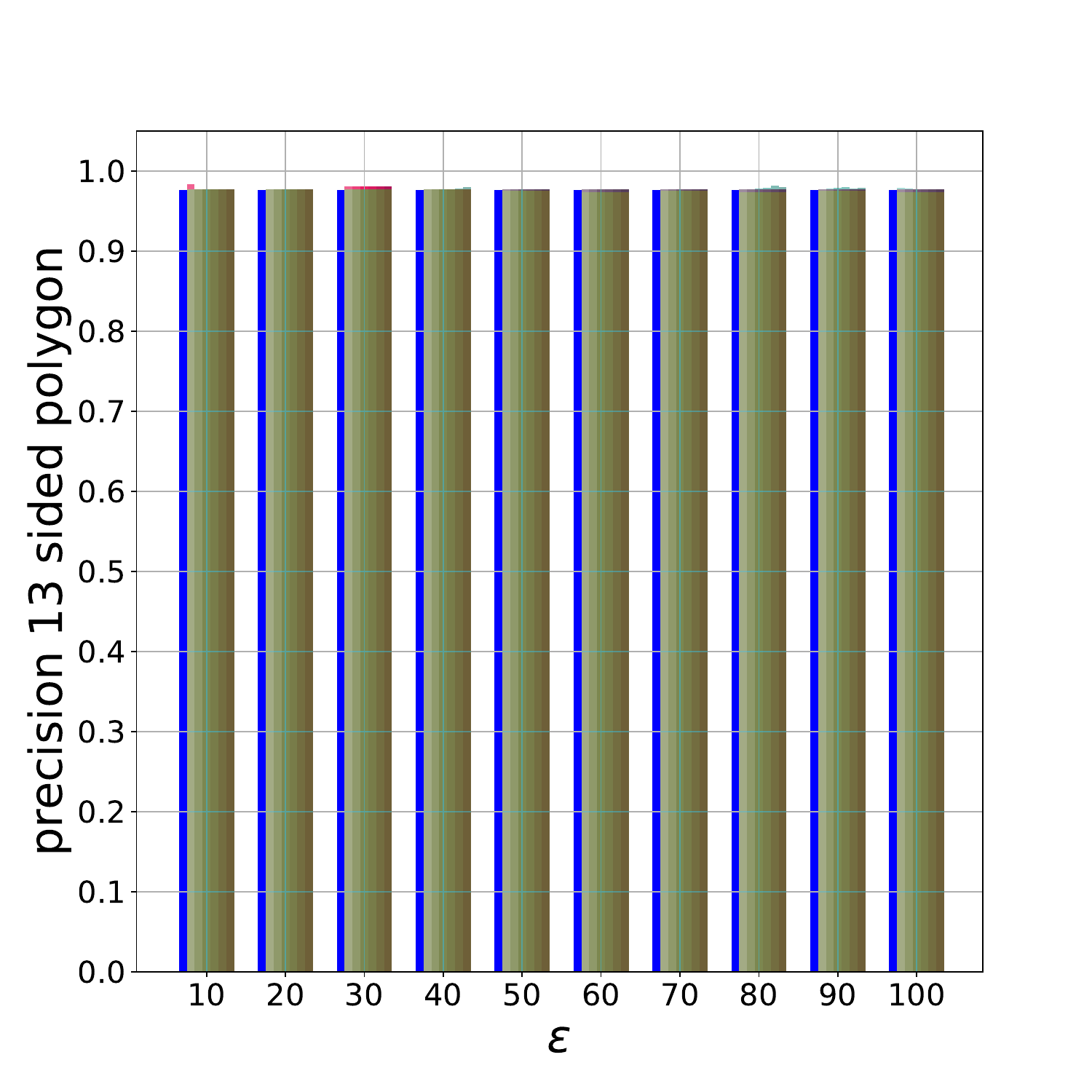}};
        \end{tikzpicture}
    \end{subfigure}
    \begin{subfigure}[b]{0.24\textwidth}
        \begin{tikzpicture}
            \node (figure) at (0,0) {\includegraphics[width=\textwidth]{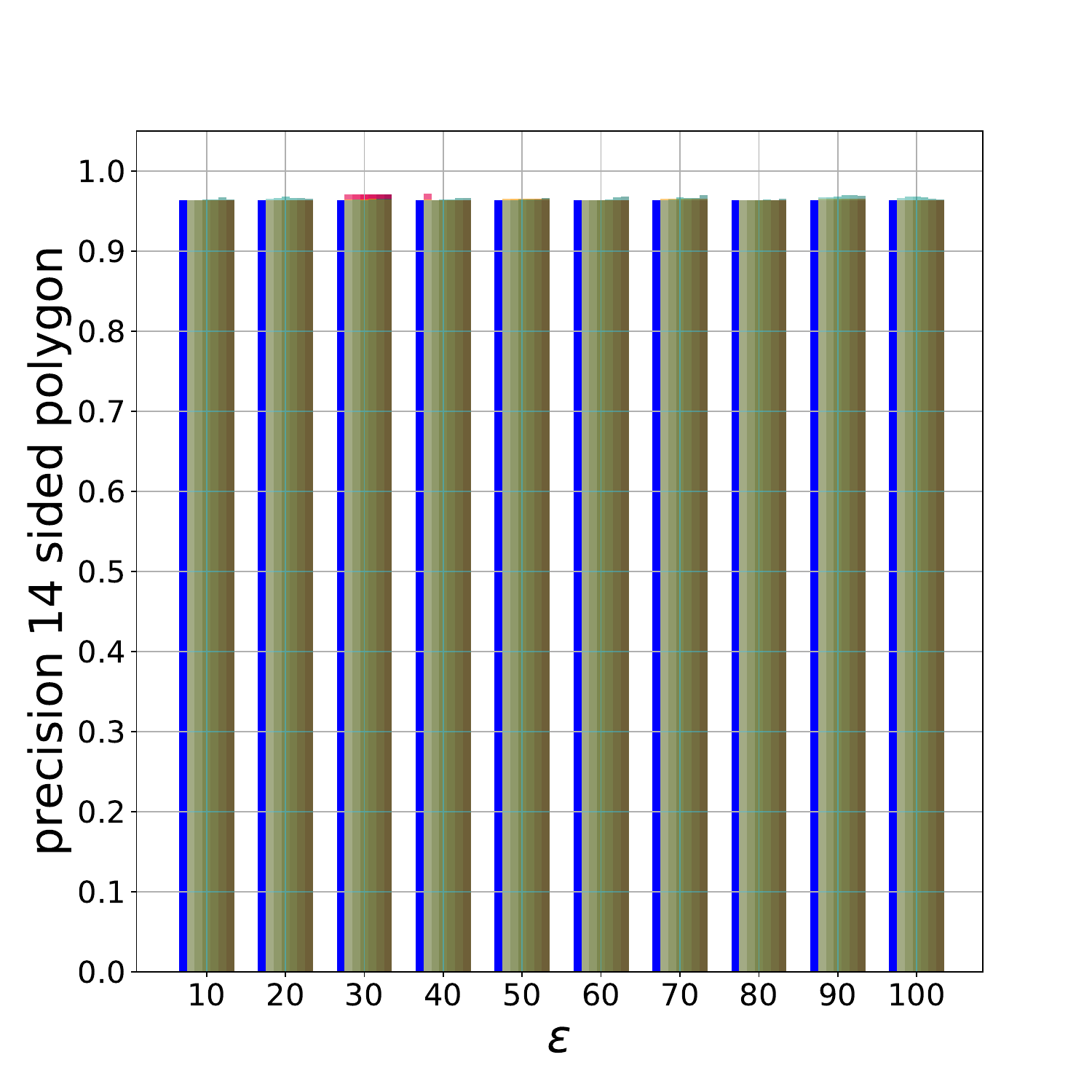}};
        \end{tikzpicture}
    \end{subfigure}

    \begin{subfigure}[b]{0.24\textwidth}
        \begin{tikzpicture}
            \node (figure) at (0,0) {\includegraphics[width=\textwidth]{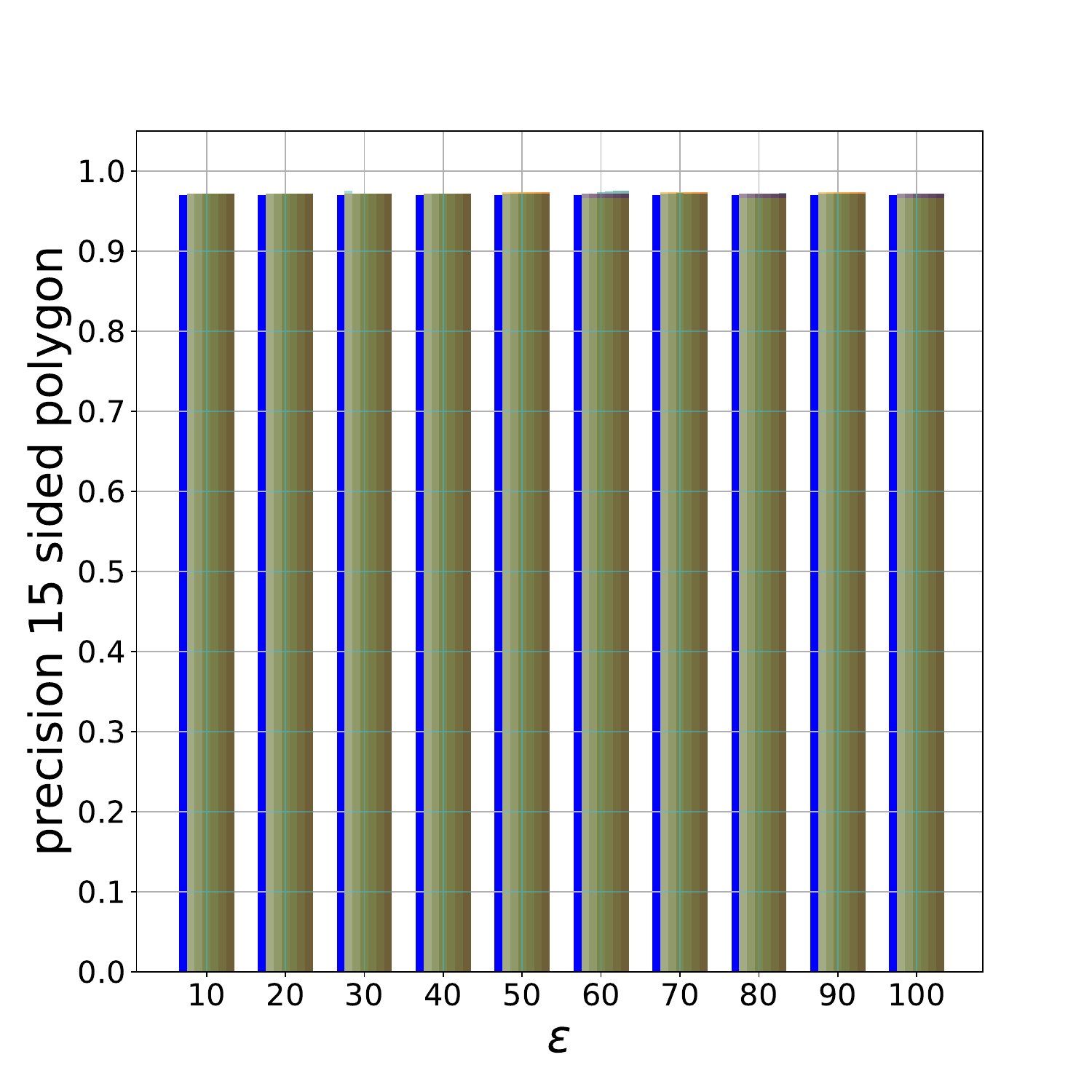}};
        \end{tikzpicture}
    \end{subfigure}
    \begin{subfigure}[b]{0.24\textwidth}
        \begin{tikzpicture}
            \node (figure) at (0,0) {\includegraphics[width=\textwidth]{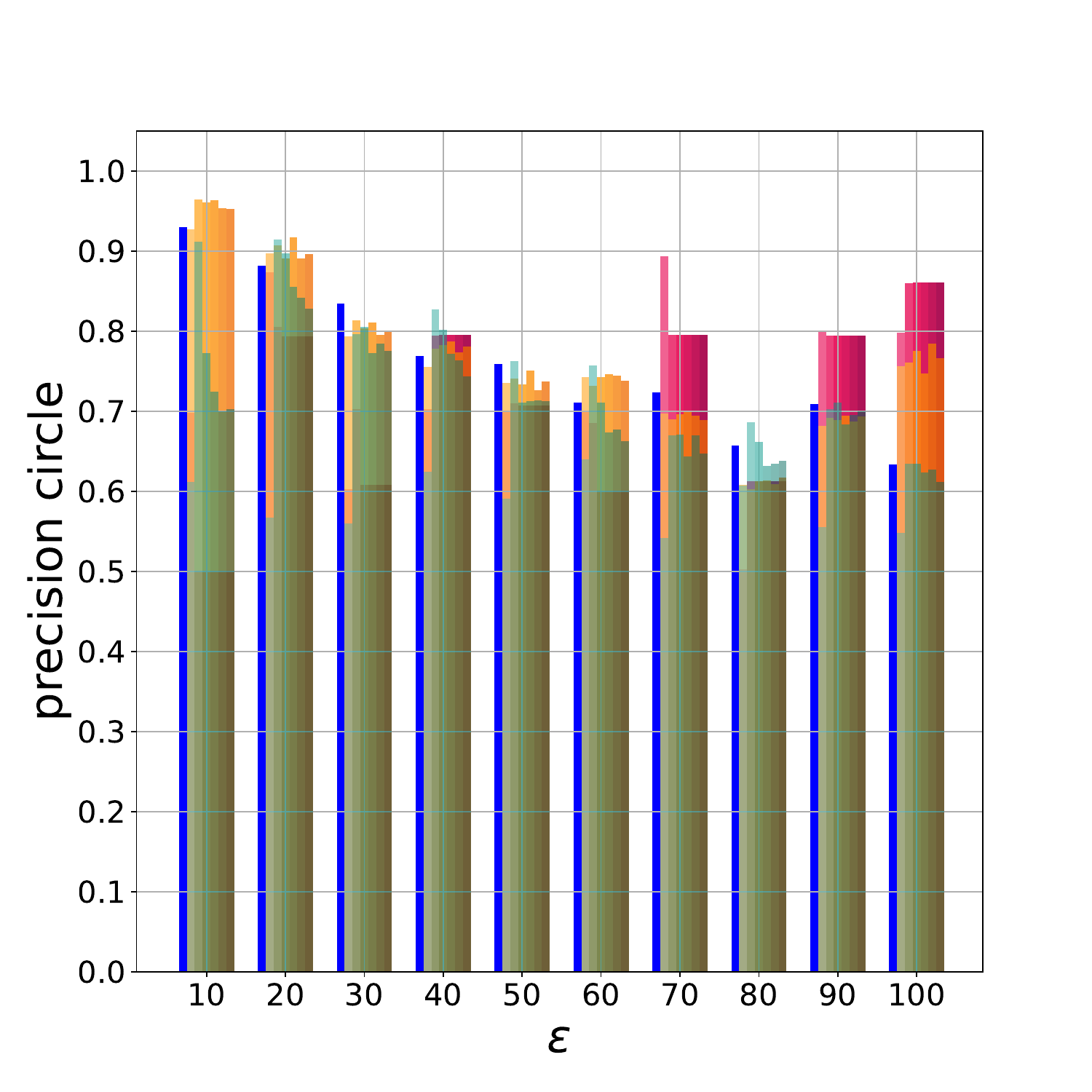}};
        \end{tikzpicture}
    \end{subfigure}
\end{figure}

\newpage

\begin{figure}[ht]
    \centering

    \begin{subfigure}[b]{0.24\textwidth}
        \begin{tikzpicture}
            \node (figure) at (0,0) {\includegraphics[width=\textwidth]{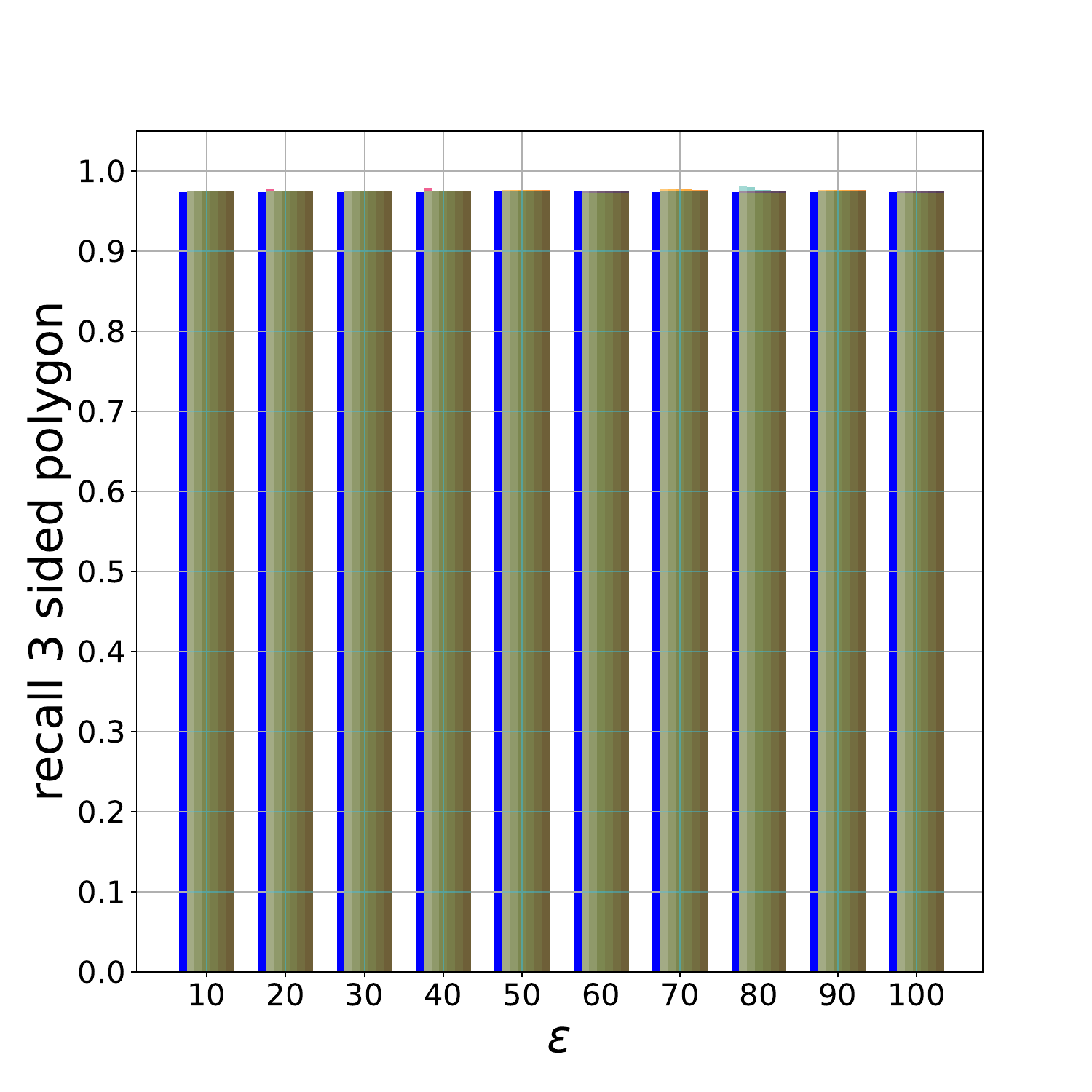}};
        \end{tikzpicture}
    \end{subfigure}
    \begin{subfigure}[b]{0.24\textwidth}
        \begin{tikzpicture}
            \node (figure) at (0,0) {\includegraphics[width=\textwidth]{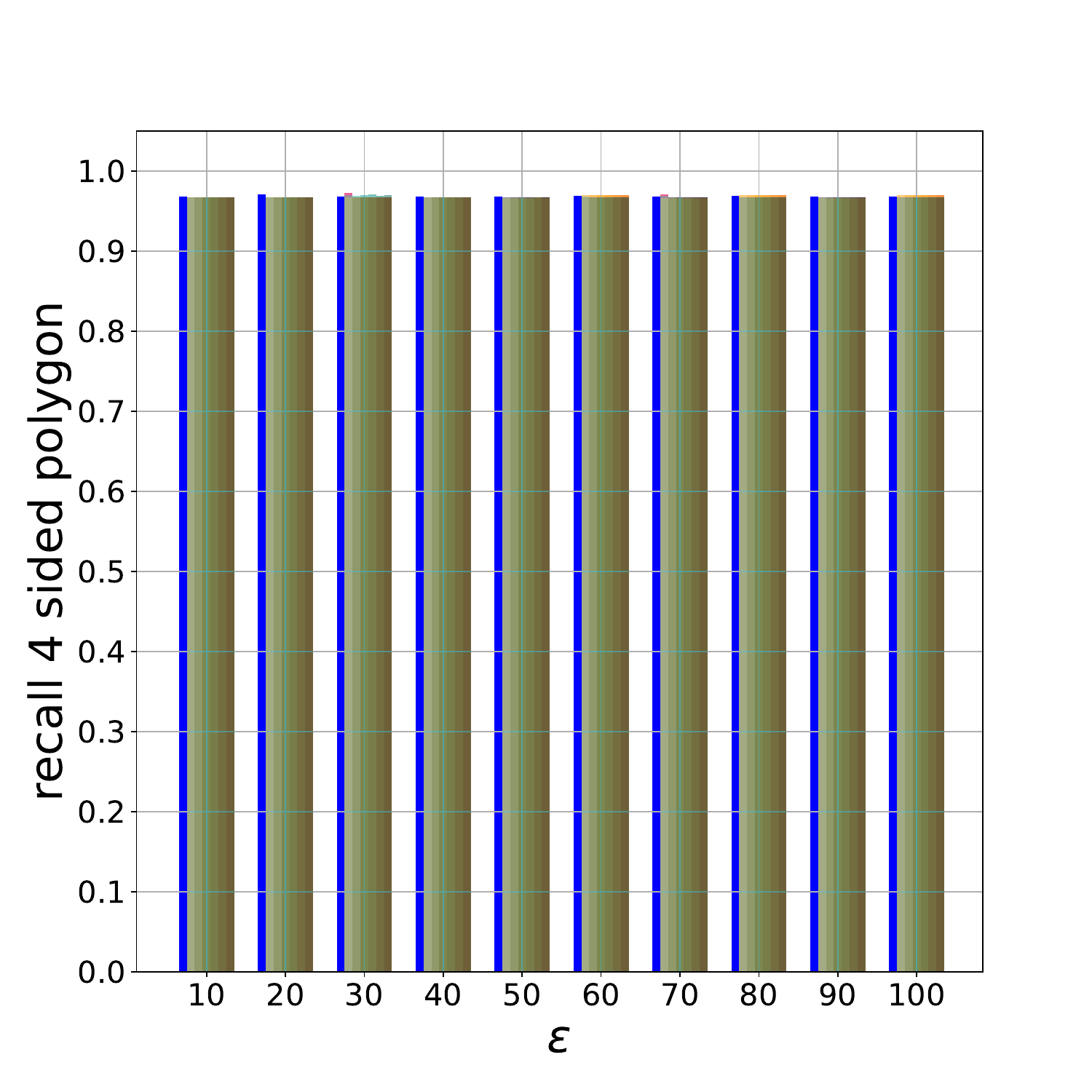}};
        \end{tikzpicture}
    \end{subfigure}
    \begin{subfigure}[b]{0.24\textwidth}
        \begin{tikzpicture}
            \node (figure) at (0,0) {\includegraphics[width=\textwidth]{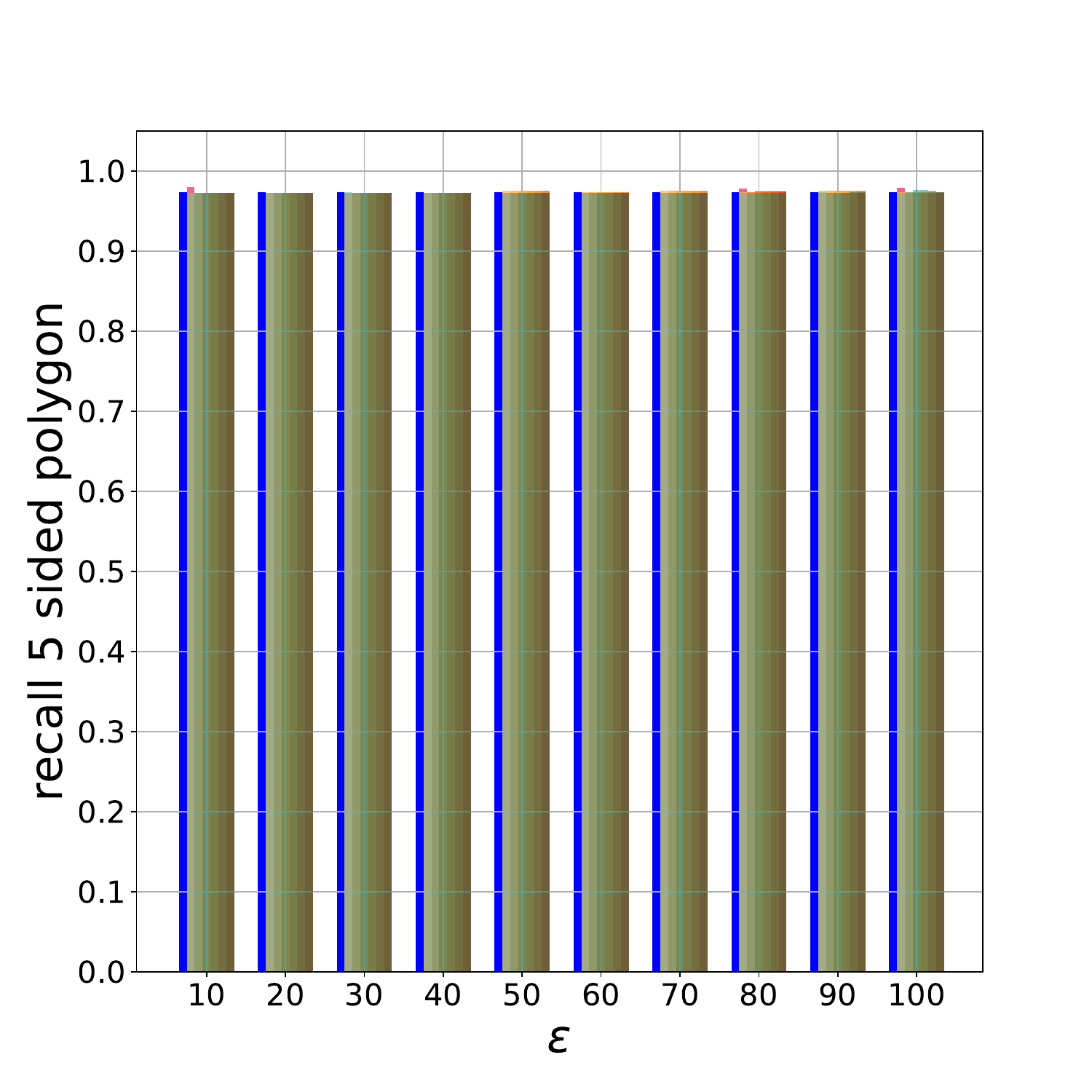}};
        \end{tikzpicture}
    \end{subfigure}
    \begin{subfigure}[b]{0.24\textwidth}
        \begin{tikzpicture}
            \node (figure) at (0,0) {\includegraphics[width=\textwidth]{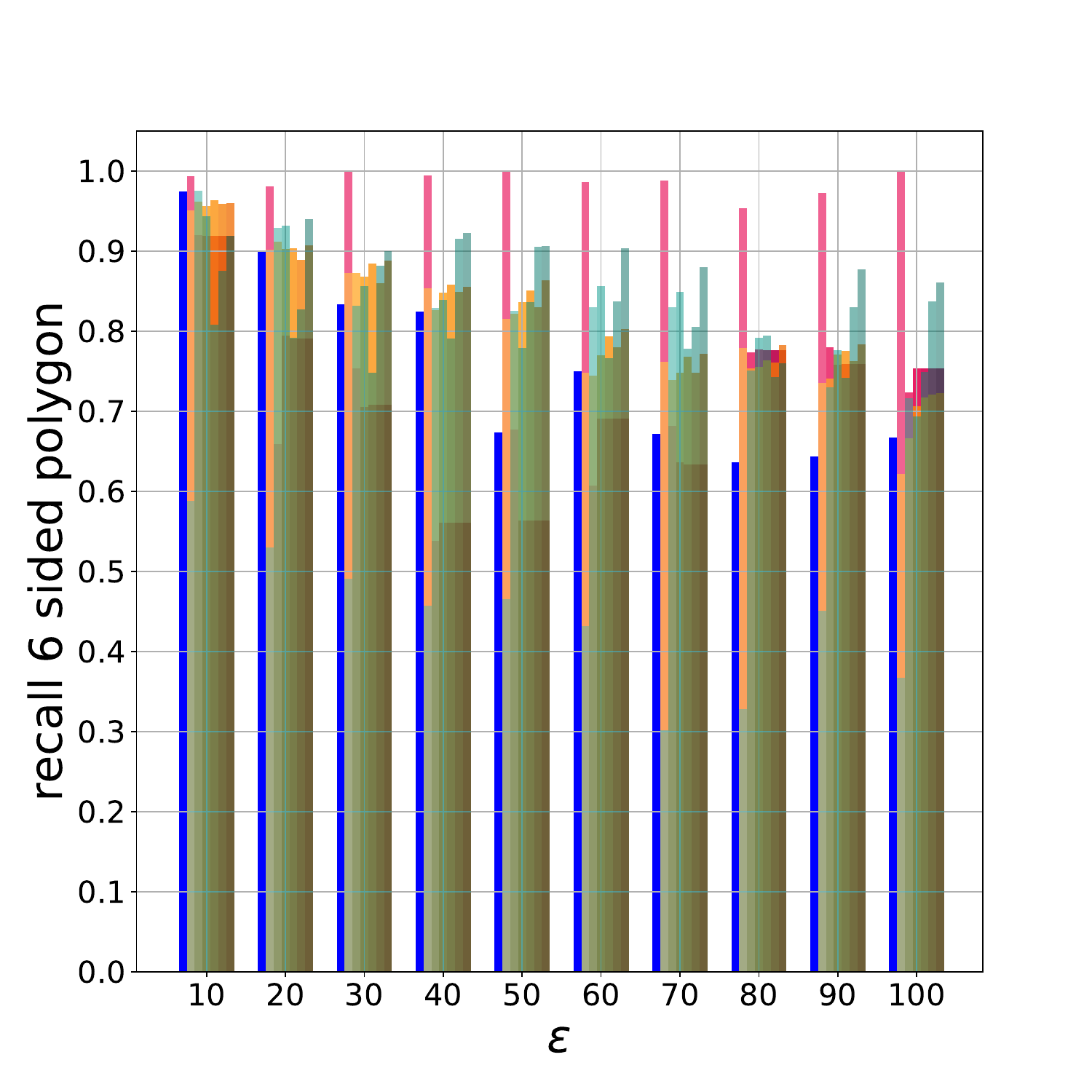}};
        \end{tikzpicture}
    \end{subfigure}

    \begin{subfigure}[b]{0.24\textwidth}
        \begin{tikzpicture}
            \node (figure) at (0,0) {\includegraphics[width=\textwidth]{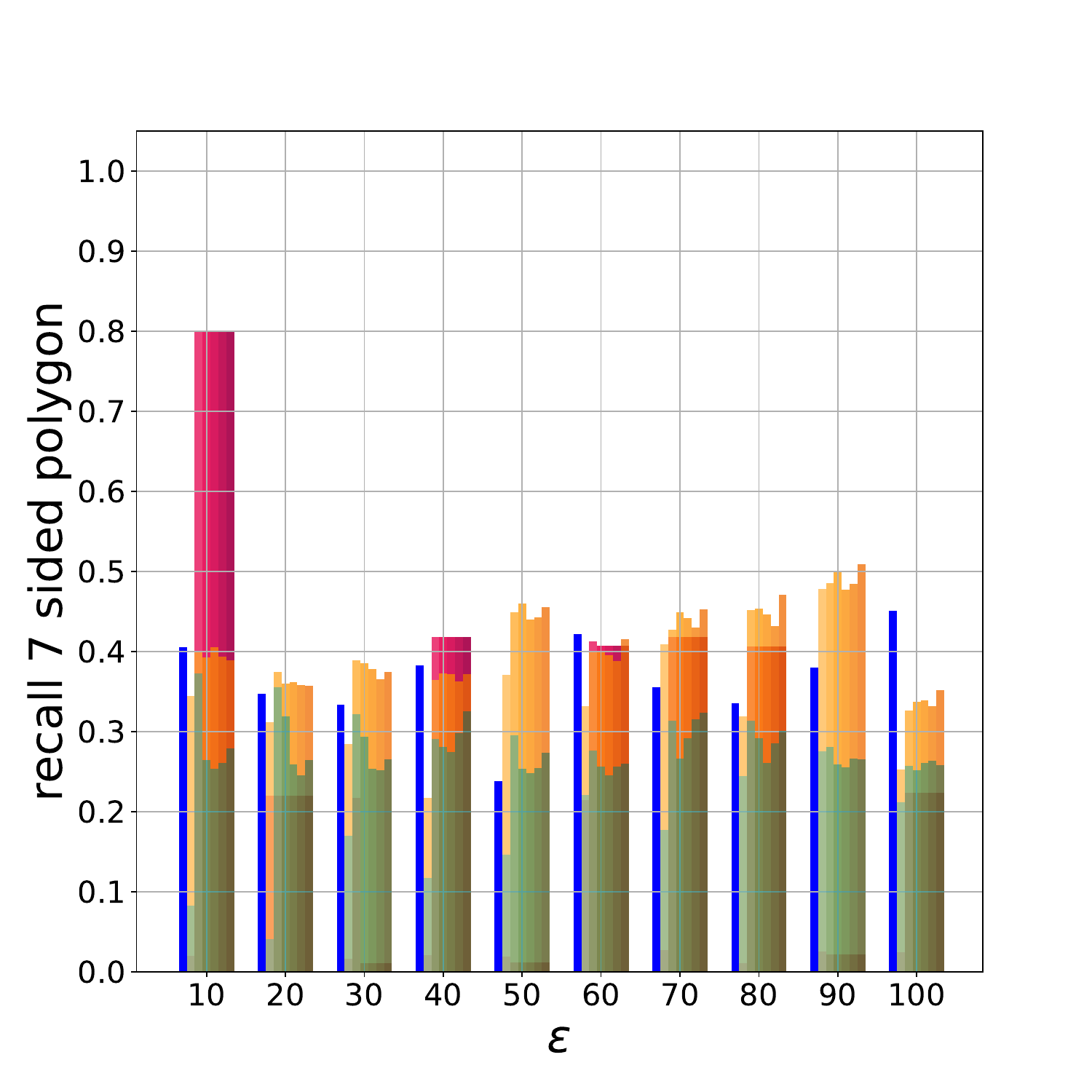}};
        \end{tikzpicture}
    \end{subfigure}
    \begin{subfigure}[b]{0.24\textwidth}
        \begin{tikzpicture}
            \node (figure) at (0,0) {\includegraphics[width=\textwidth]{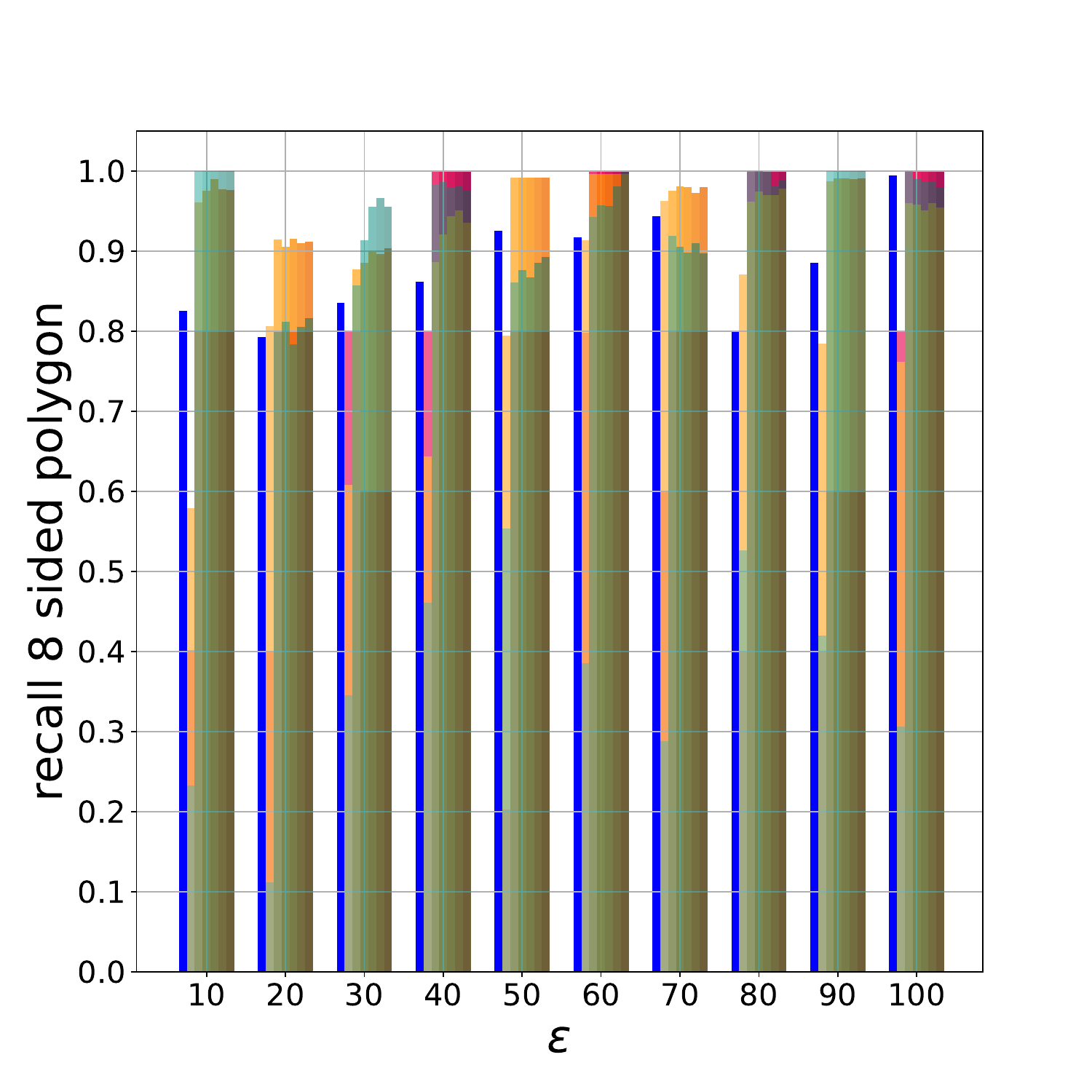}};
        \end{tikzpicture}
    \end{subfigure}
    \begin{subfigure}[b]{0.24\textwidth}
        \begin{tikzpicture}
            \node (figure) at (0,0) {\includegraphics[width=\textwidth]{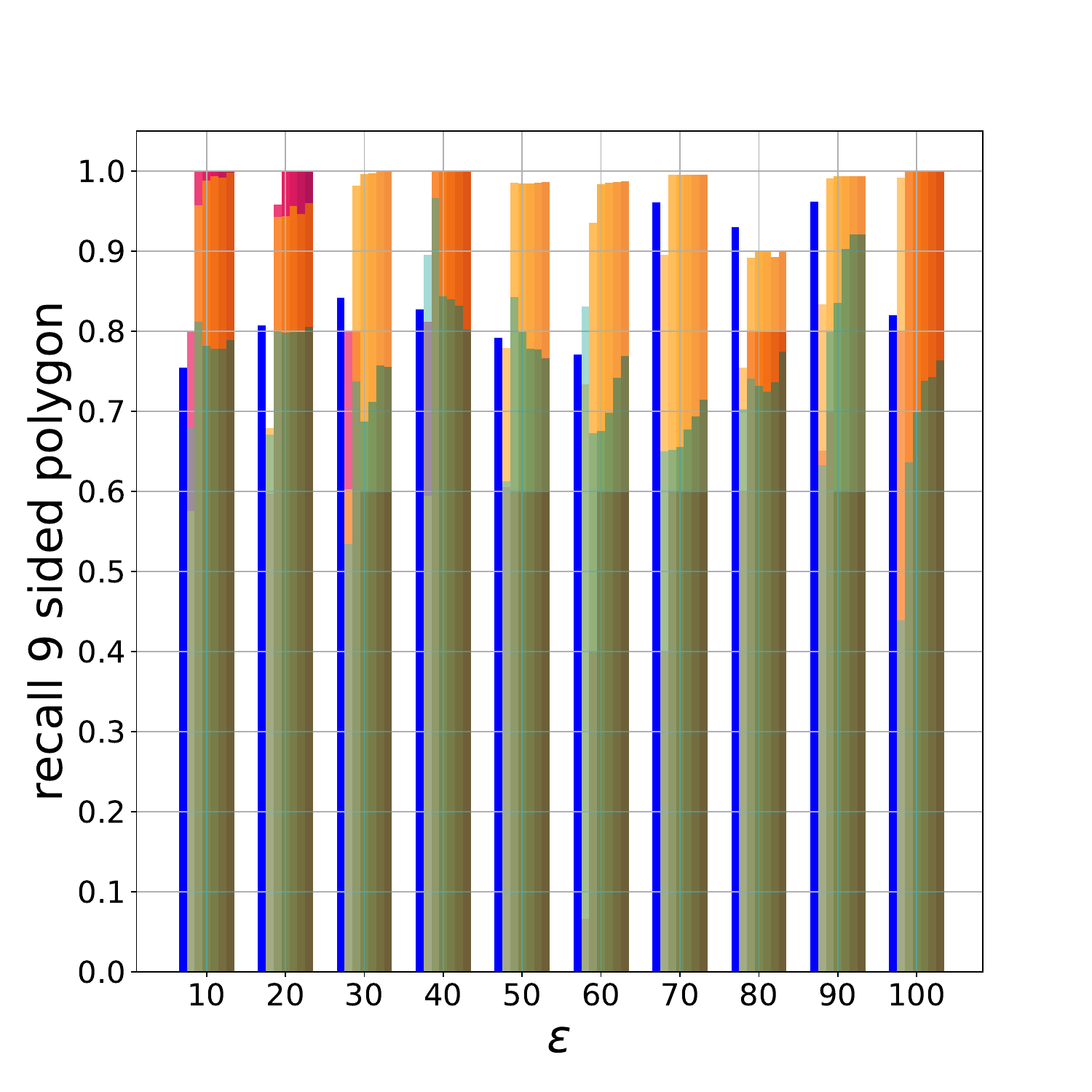}};
        \end{tikzpicture}
    \end{subfigure}
    \begin{subfigure}[b]{0.24\textwidth}
        \begin{tikzpicture}
            \node (figure) at (0,0) {\includegraphics[width=\textwidth]{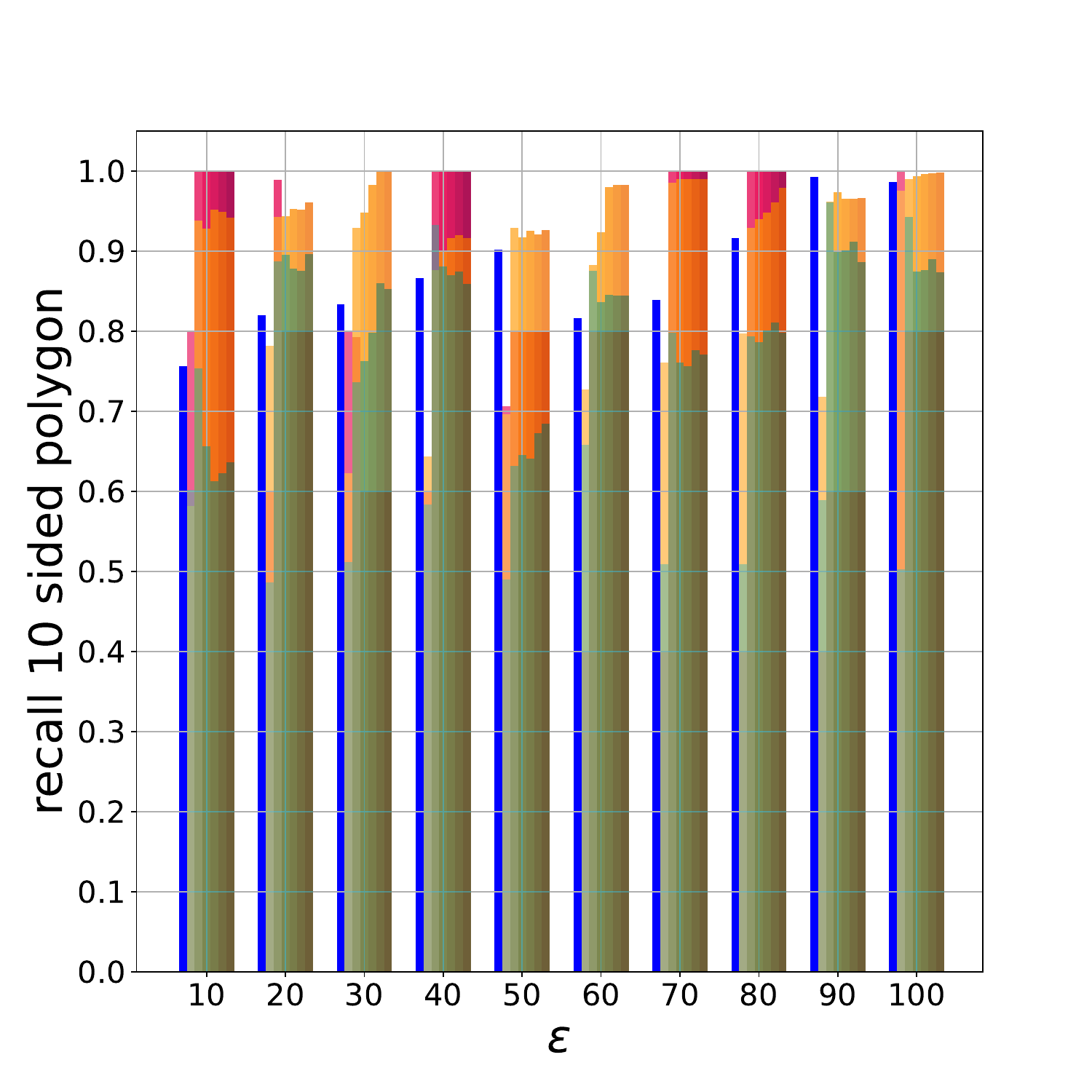}};
        \end{tikzpicture}
    \end{subfigure}

    \begin{subfigure}[b]{0.24\textwidth}
        \begin{tikzpicture}
            \node (figure) at (0,0) {\includegraphics[width=\textwidth]{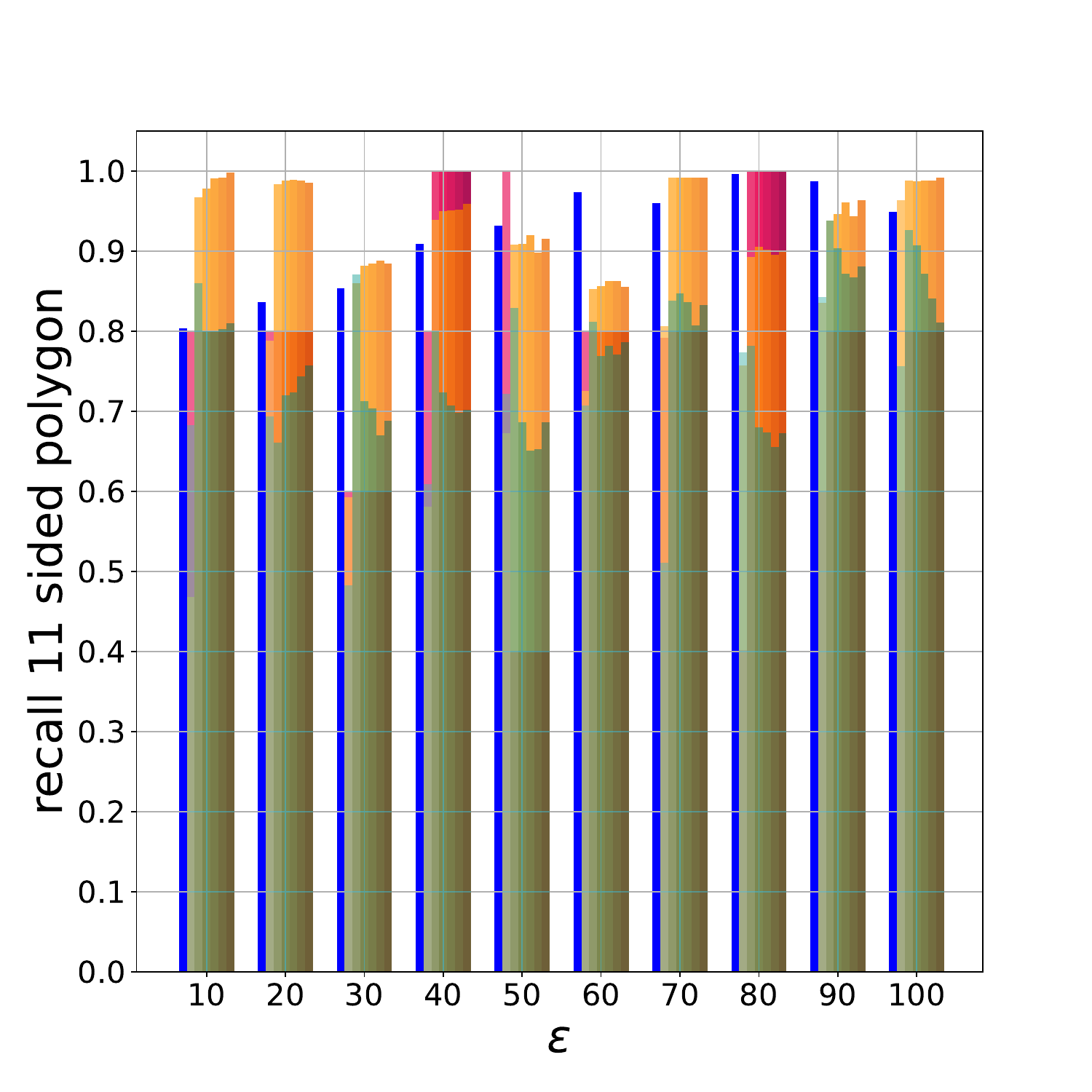}};
        \end{tikzpicture}
    \end{subfigure}
    \begin{subfigure}[b]{0.24\textwidth}
        \begin{tikzpicture}
            \node (figure) at (0,0) {\includegraphics[width=\textwidth]{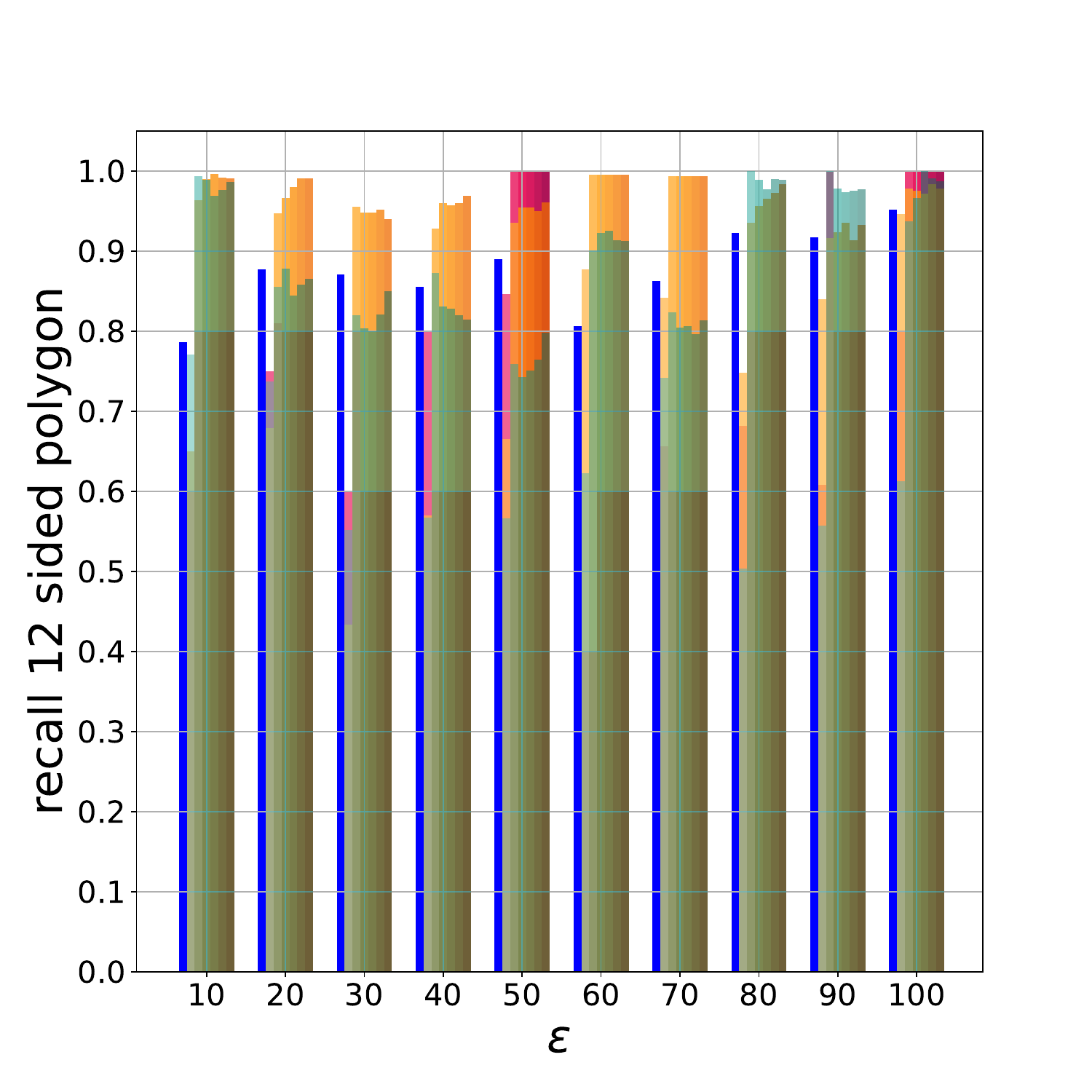}};
        \end{tikzpicture}
    \end{subfigure}
    \begin{subfigure}[b]{0.24\textwidth}
        \begin{tikzpicture}
            \node (figure) at (0,0) {\includegraphics[width=\textwidth]{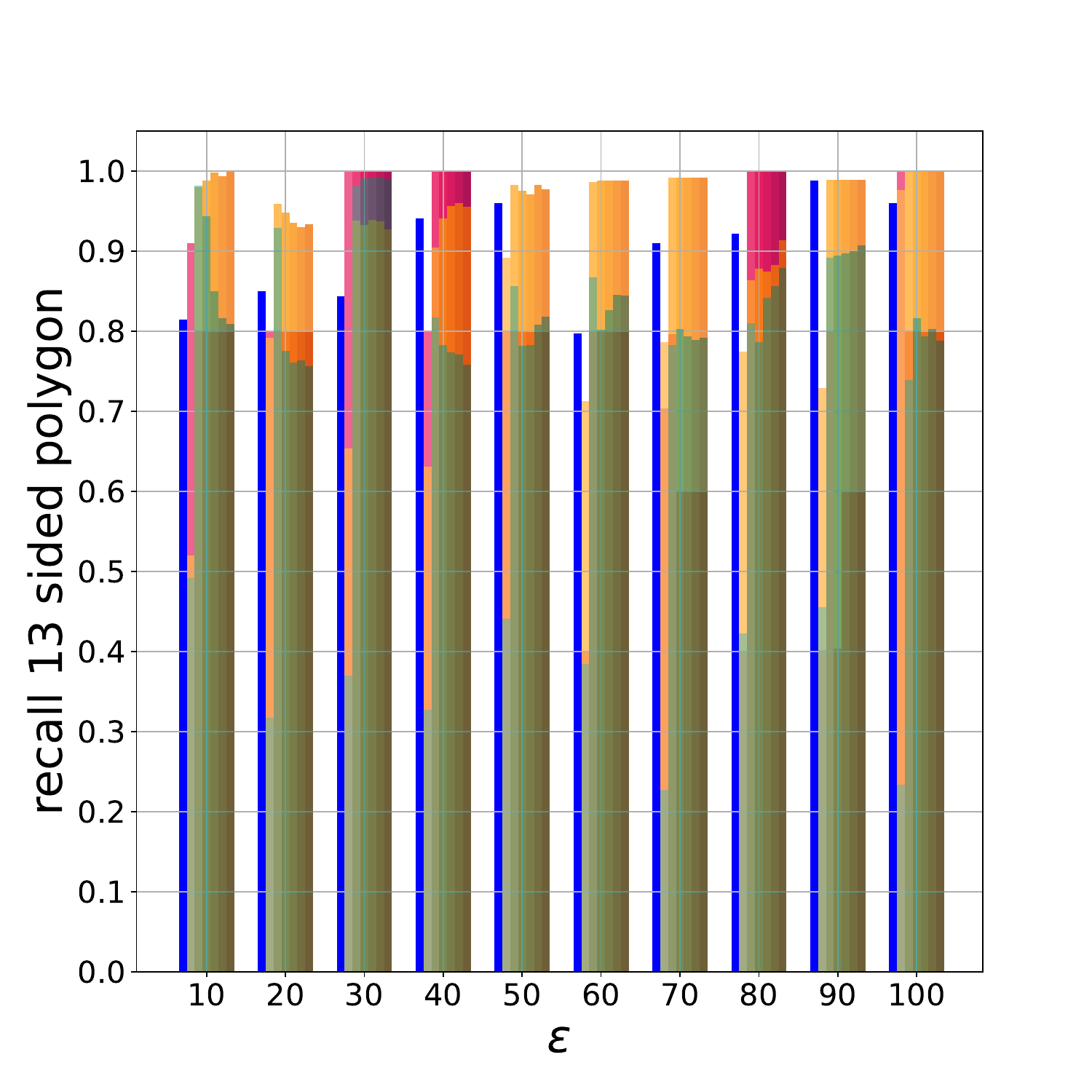}};
        \end{tikzpicture}
    \end{subfigure}
    \begin{subfigure}[b]{0.24\textwidth}
        \begin{tikzpicture}
            \node (figure) at (0,0) {\includegraphics[width=\textwidth]{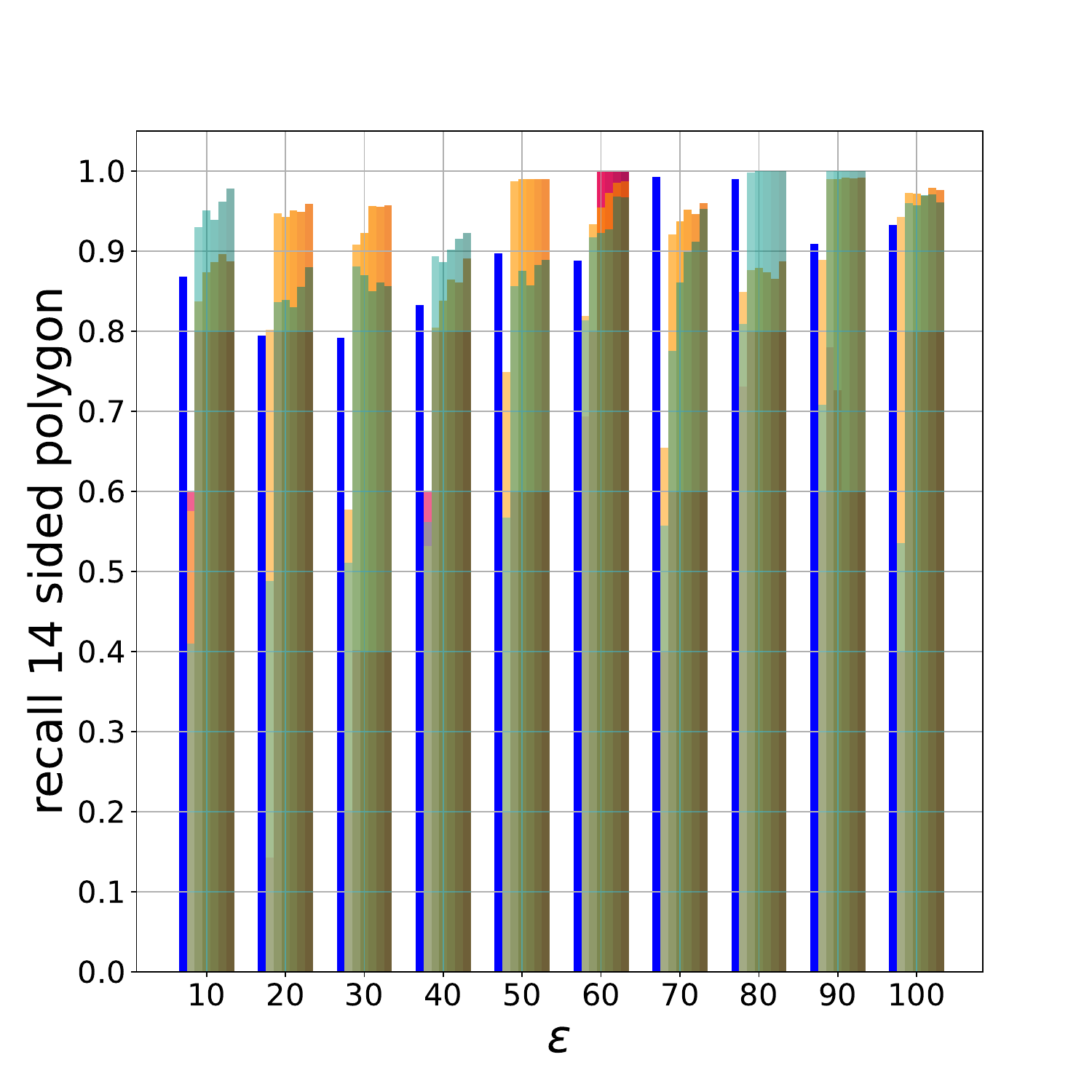}};
        \end{tikzpicture}
    \end{subfigure}

    \begin{subfigure}[b]{0.24\textwidth}
        \begin{tikzpicture}
            \node (figure) at (0,0) {\includegraphics[width=\textwidth]{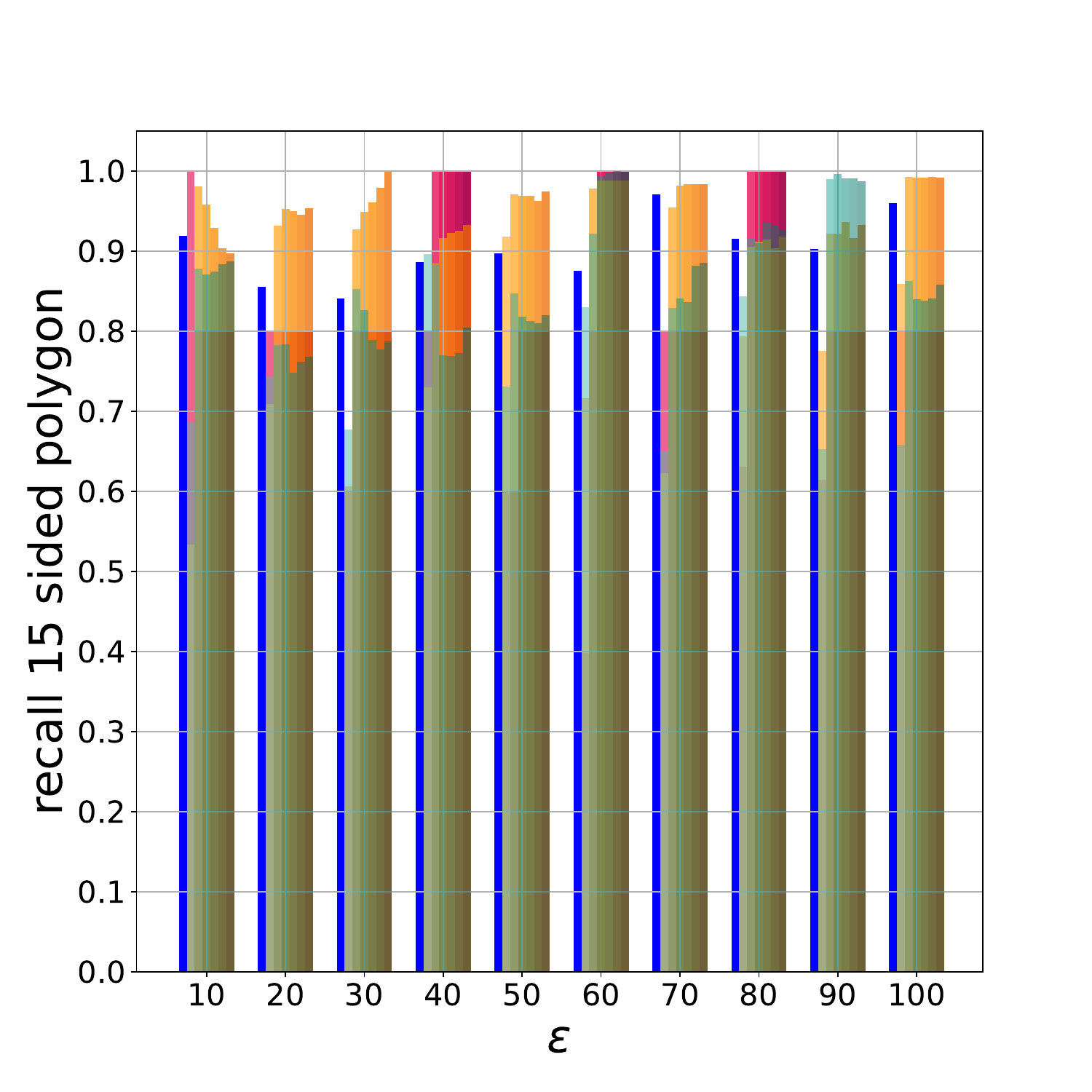}};
        \end{tikzpicture}
    \end{subfigure}
    \begin{subfigure}[b]{0.24\textwidth}
        \begin{tikzpicture}
            \node (figure) at (0,0) {\includegraphics[width=\textwidth]{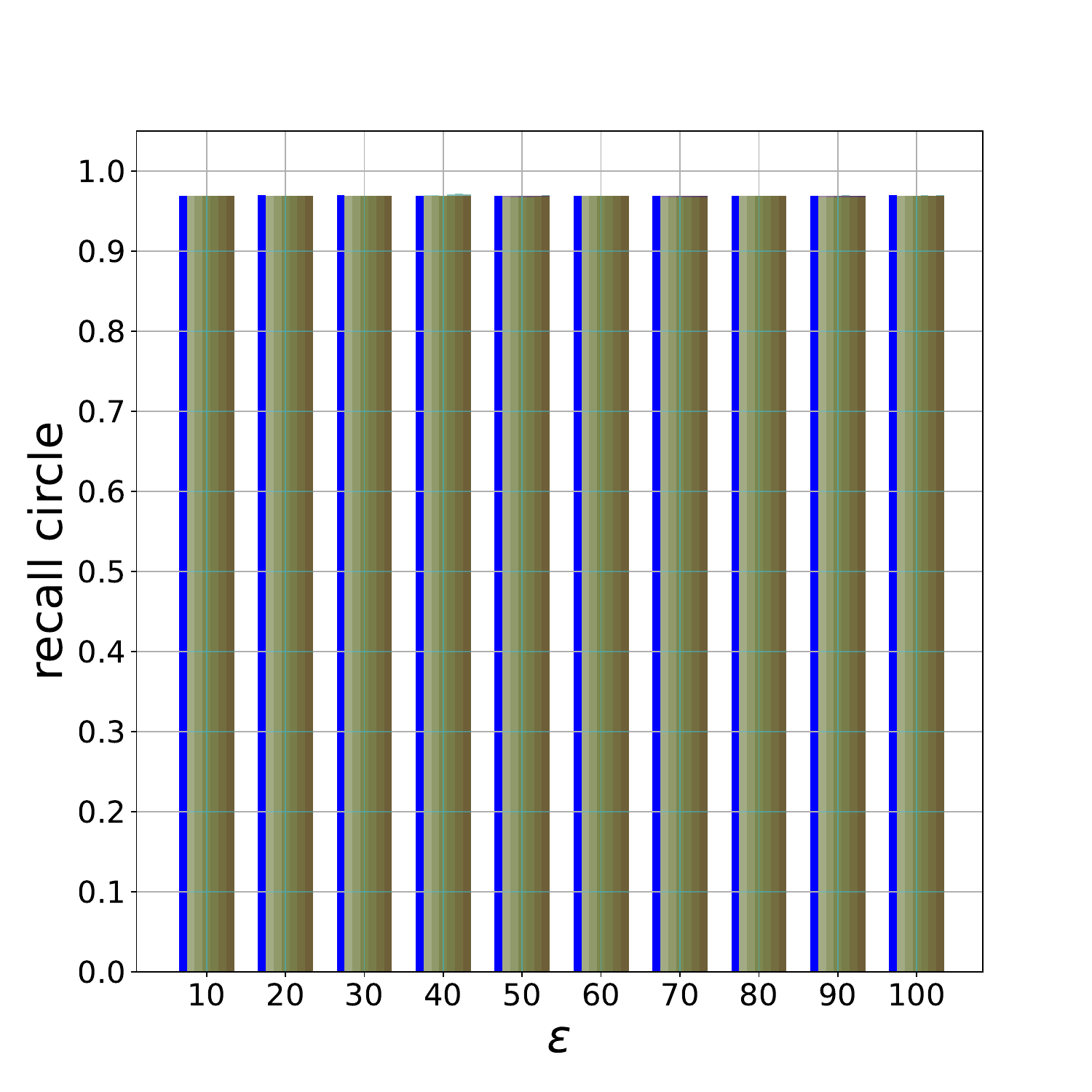}};
        \end{tikzpicture}
    \end{subfigure}
\end{figure}

%% file: results/examples/artificial.tex
\begin{figure}[ht]
    \centering

    \begin{subfigure}[b]{0.24\textwidth}
        \begin{tikzpicture}
            \node (figure) at (0,0) {\includegraphics[width=\textwidth]{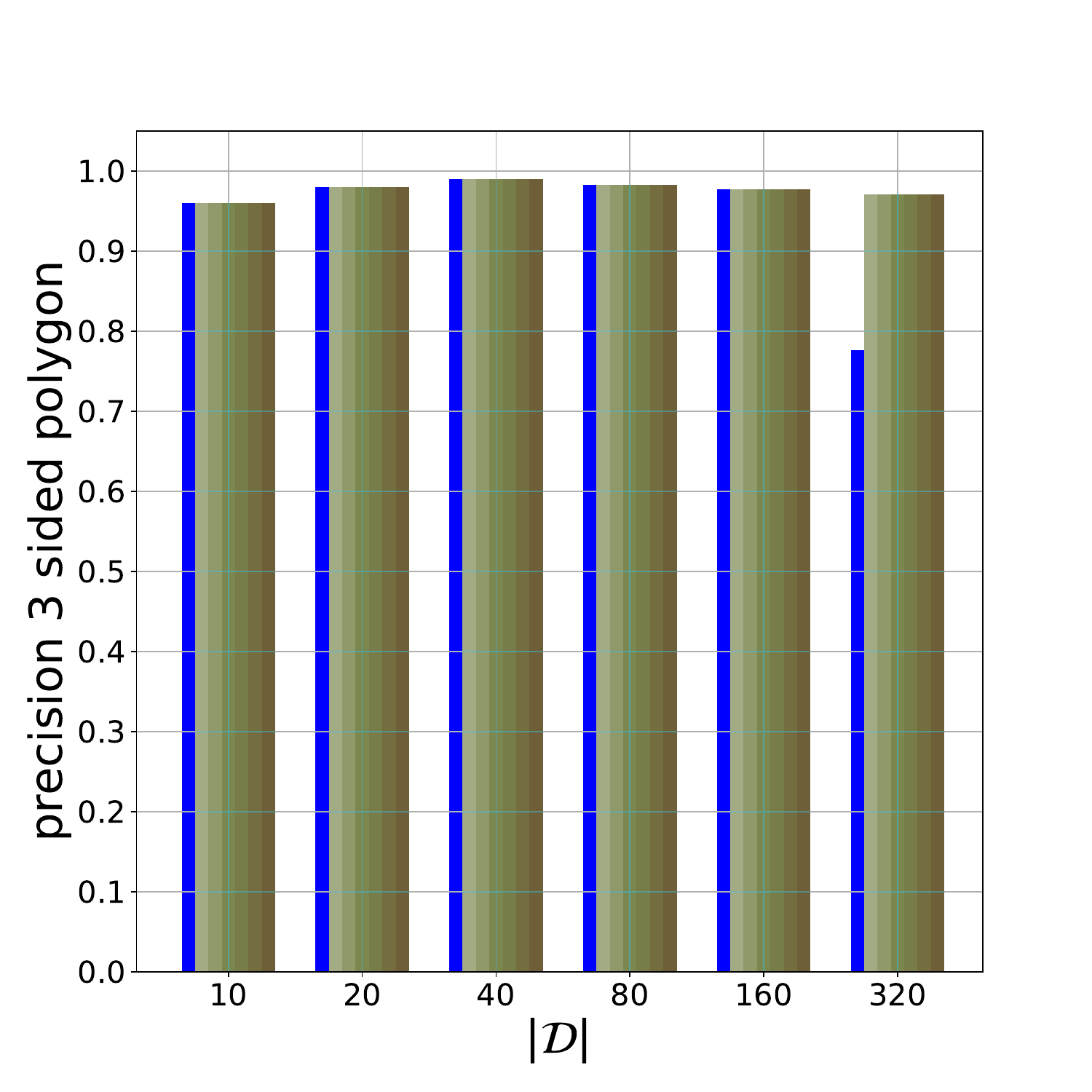}};
        \end{tikzpicture}
    \end{subfigure}
    \begin{subfigure}[b]{0.24\textwidth}
        \begin{tikzpicture}
            \node (figure) at (0,0) {\includegraphics[width=\textwidth]{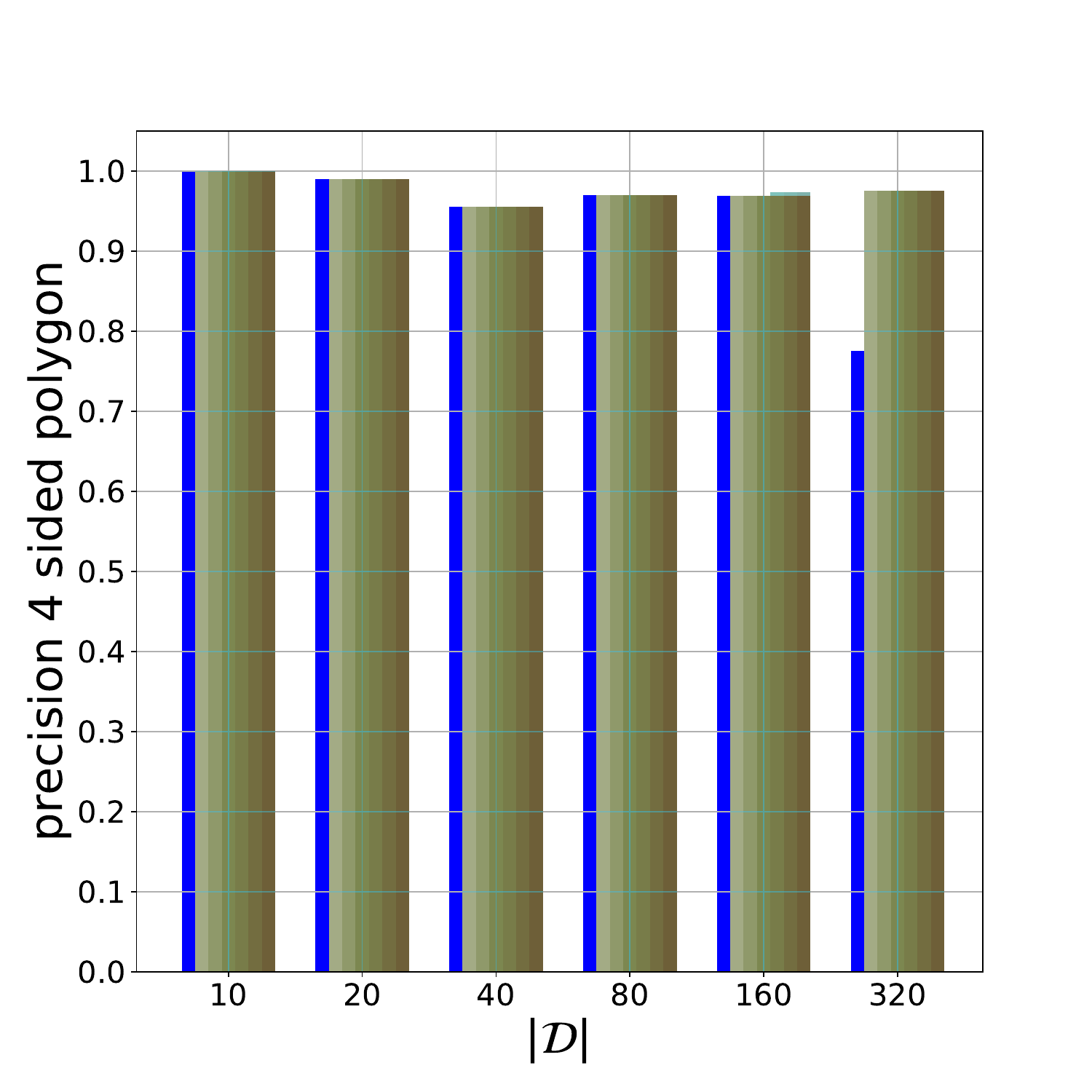}};
        \end{tikzpicture}
    \end{subfigure}
    \begin{subfigure}[b]{0.24\textwidth}
        \begin{tikzpicture}
            \node (figure) at (0,0) {\includegraphics[width=\textwidth]{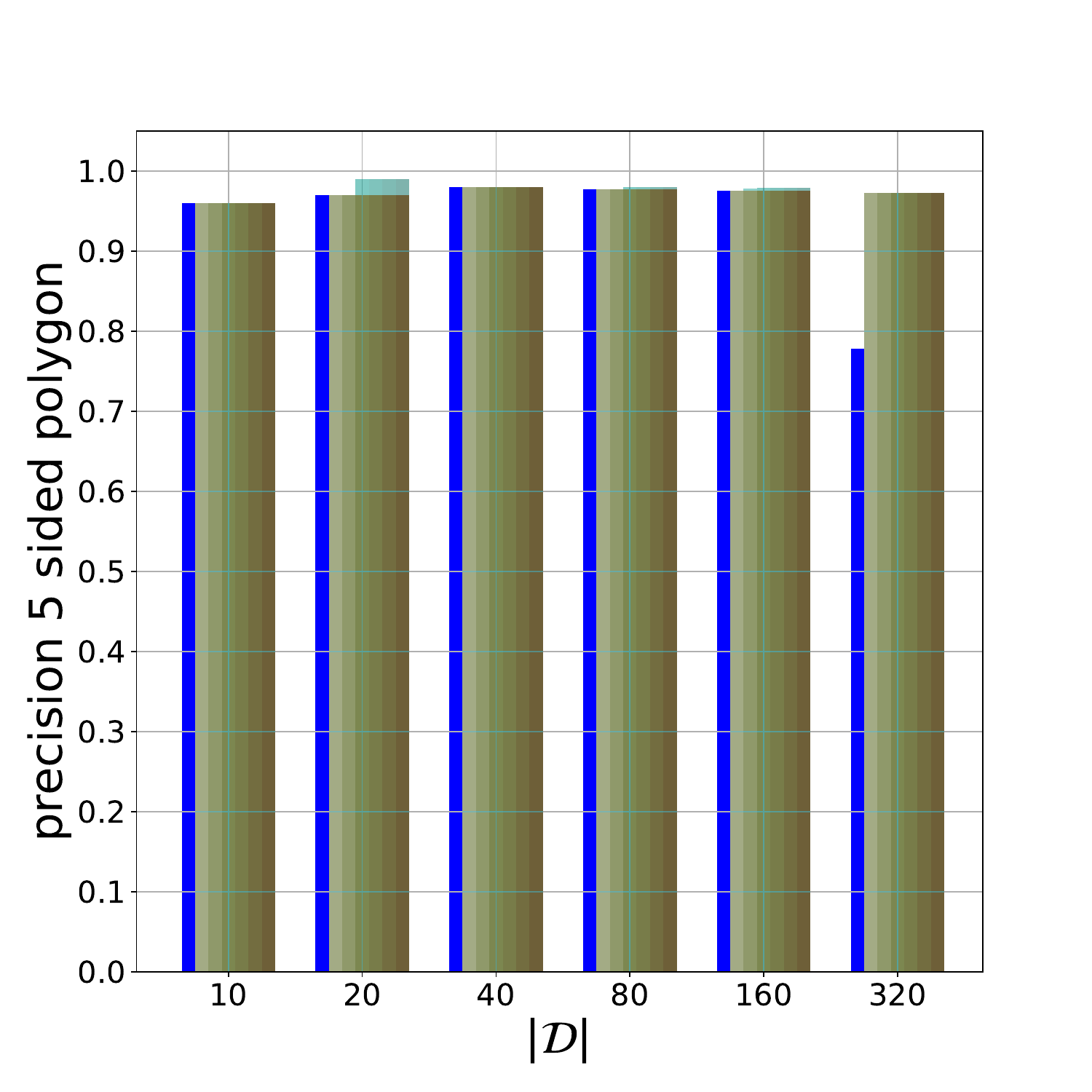}};
        \end{tikzpicture}
    \end{subfigure}
    \begin{subfigure}[b]{0.24\textwidth}
        \begin{tikzpicture}
            \node (figure) at (0,0) {\includegraphics[width=\textwidth]{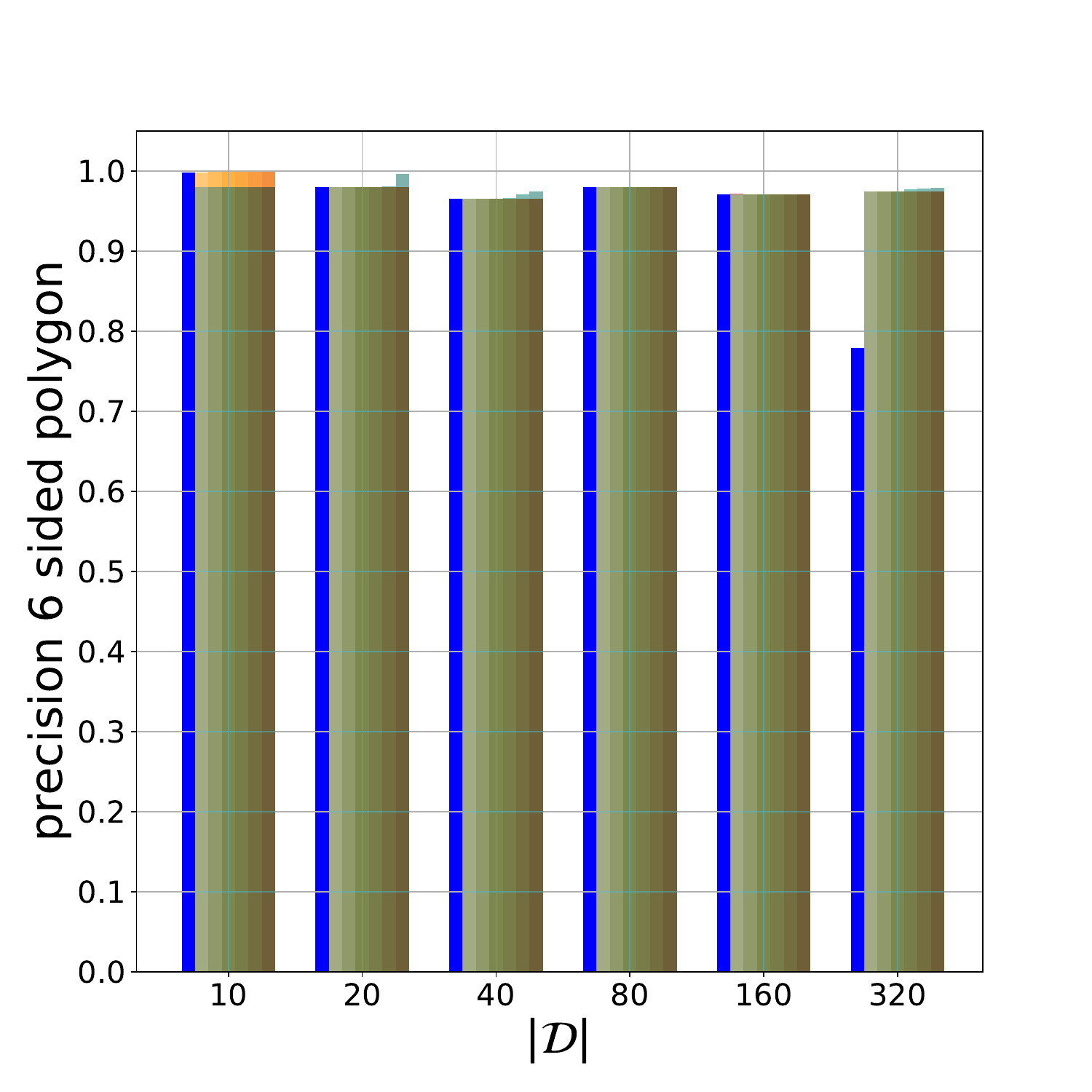}};
        \end{tikzpicture}
    \end{subfigure}

    \begin{subfigure}[b]{0.24\textwidth}
        \begin{tikzpicture}
            \node (figure) at (0,0) {\includegraphics[width=\textwidth]{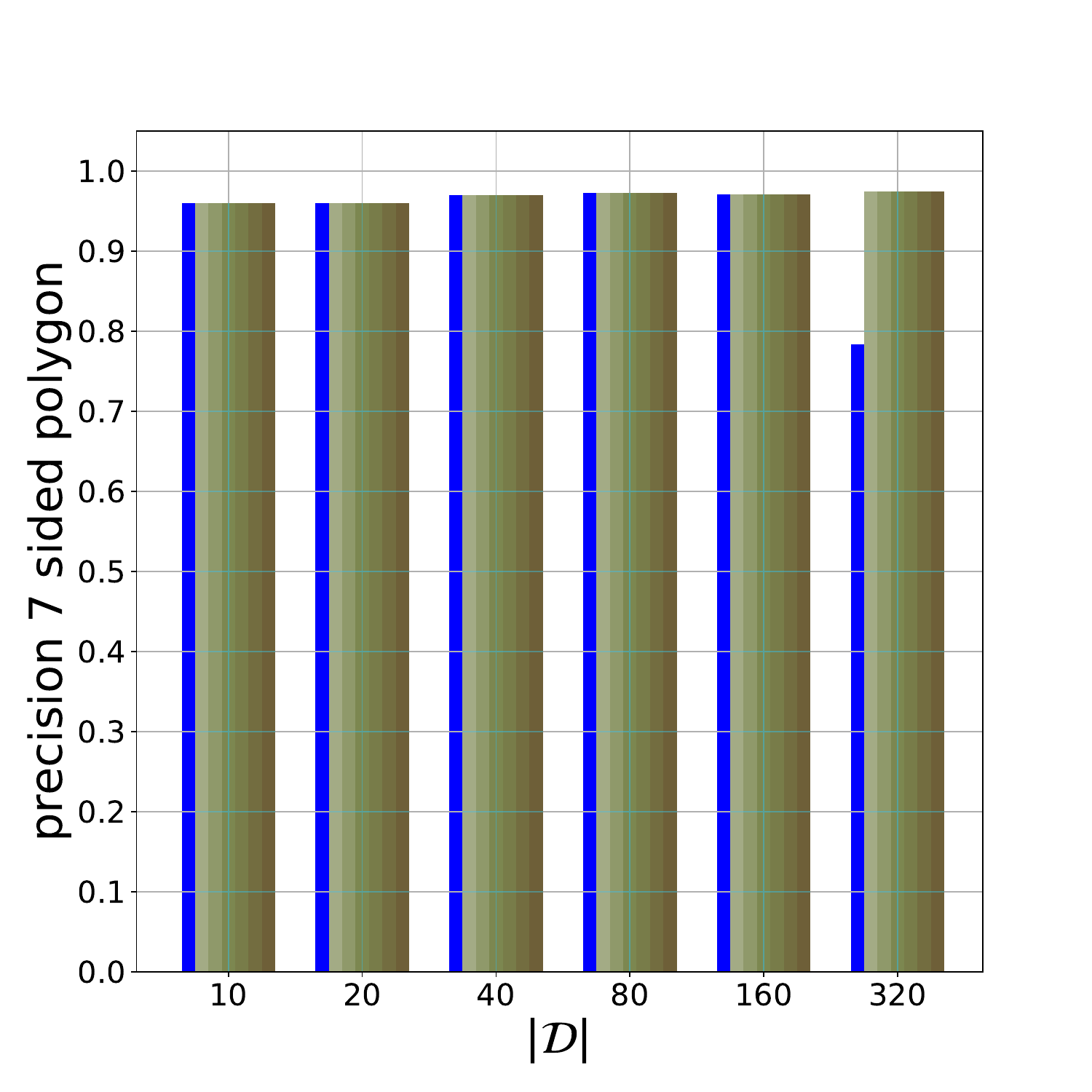}};
        \end{tikzpicture}
    \end{subfigure}
    \begin{subfigure}[b]{0.24\textwidth}
        \begin{tikzpicture}
            \node (figure) at (0,0) {\includegraphics[width=\textwidth]{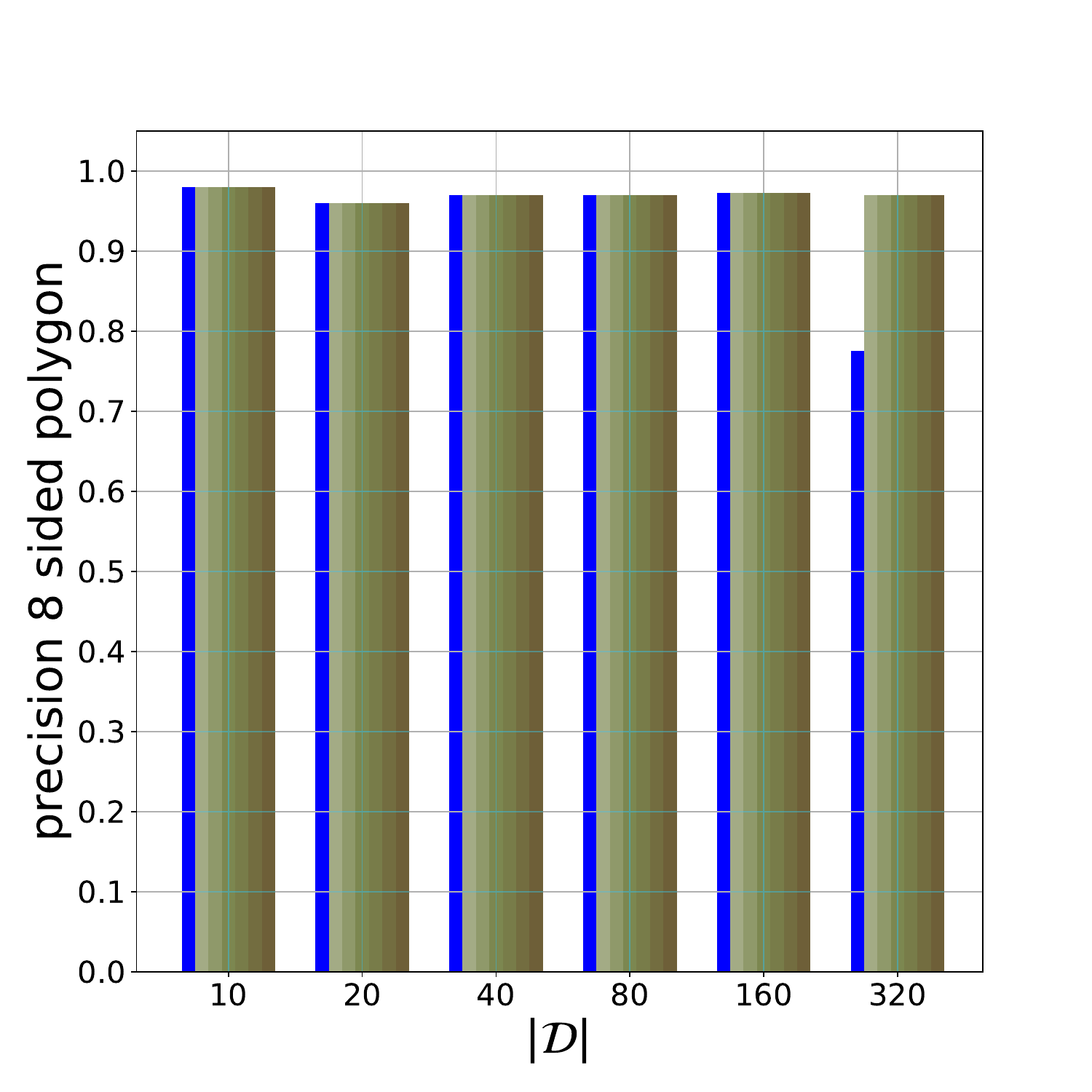}};
        \end{tikzpicture}
    \end{subfigure}
    \begin{subfigure}[b]{0.24\textwidth}
        \begin{tikzpicture}
            \node (figure) at (0,0) {\includegraphics[width=\textwidth]{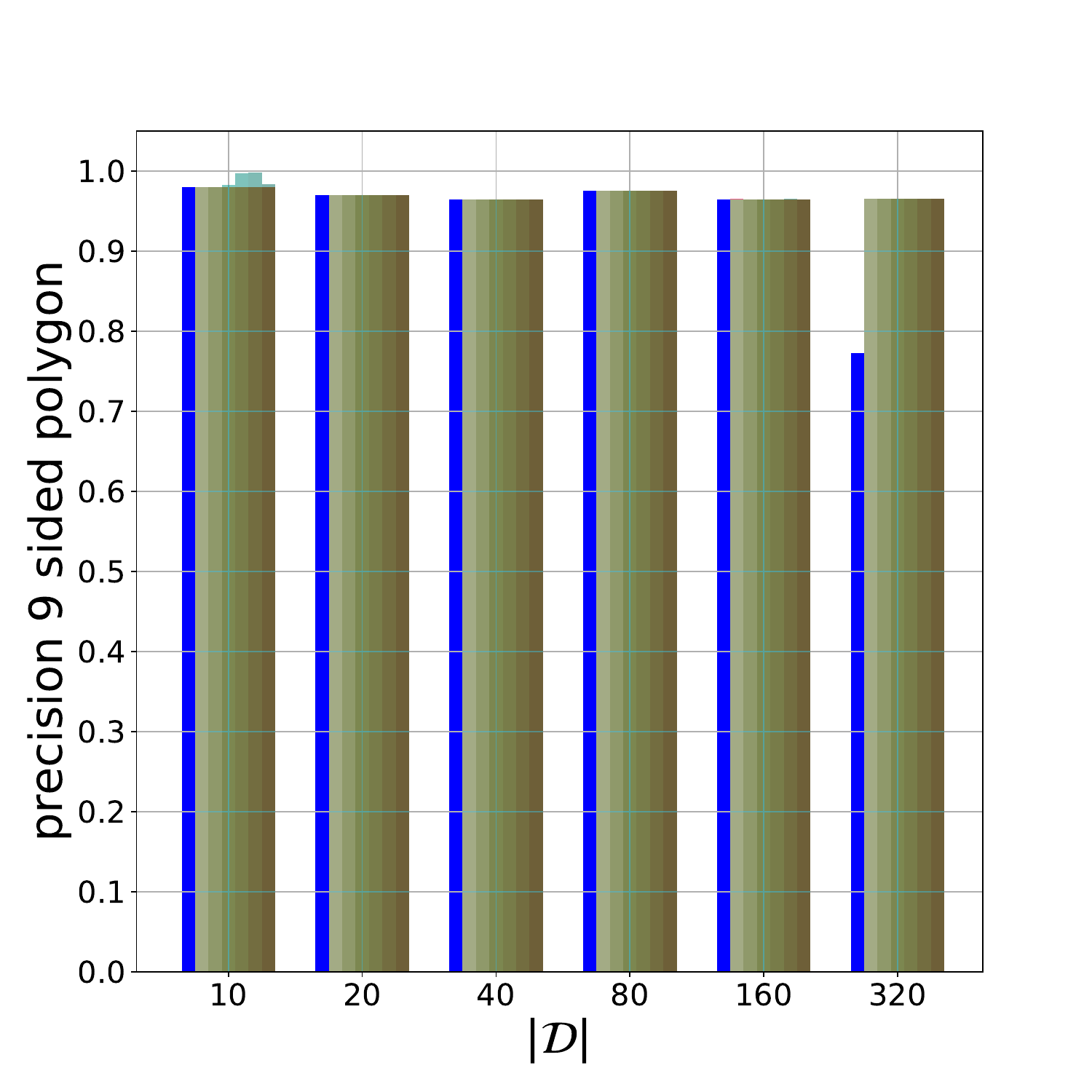}};
        \end{tikzpicture}
    \end{subfigure}
    \begin{subfigure}[b]{0.24\textwidth}
        \begin{tikzpicture}
            \node (figure) at (0,0) {\includegraphics[width=\textwidth]{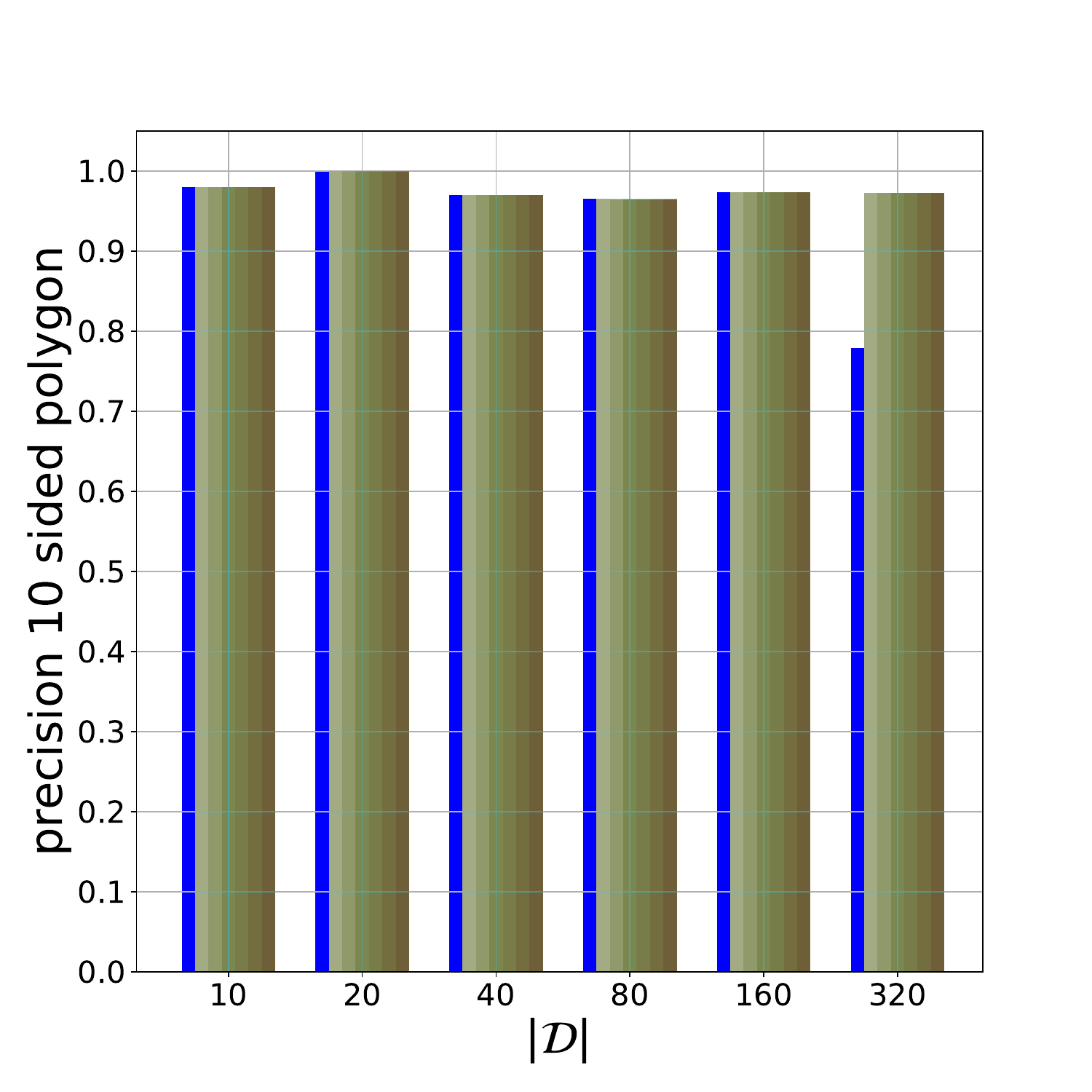}};
        \end{tikzpicture}
    \end{subfigure}

    \begin{subfigure}[b]{0.24\textwidth}
        \begin{tikzpicture}
            \node (figure) at (0,0) {\includegraphics[width=\textwidth]{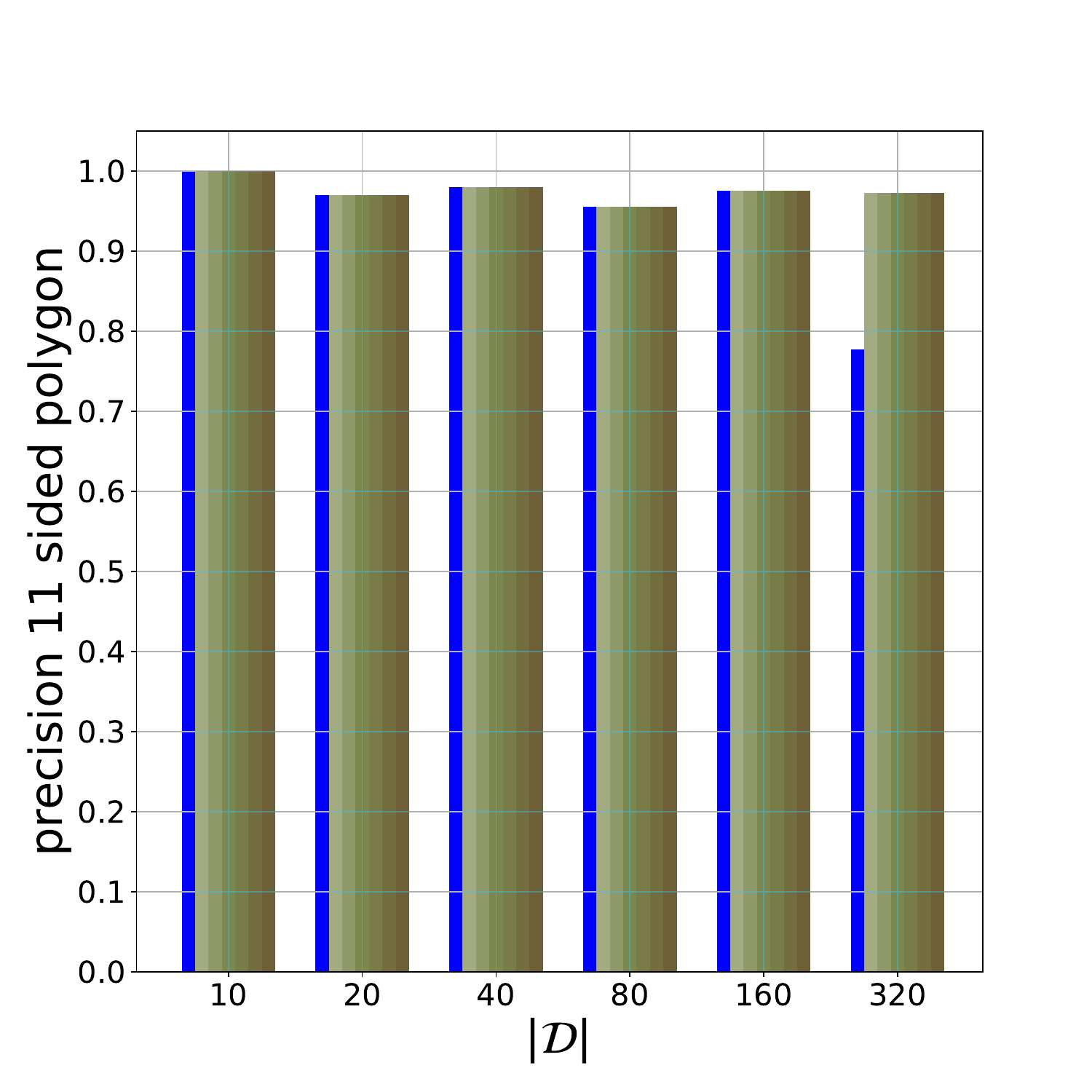}};
        \end{tikzpicture}
    \end{subfigure}
    \begin{subfigure}[b]{0.24\textwidth}
        \begin{tikzpicture}
            \node (figure) at (0,0) {\includegraphics[width=\textwidth]{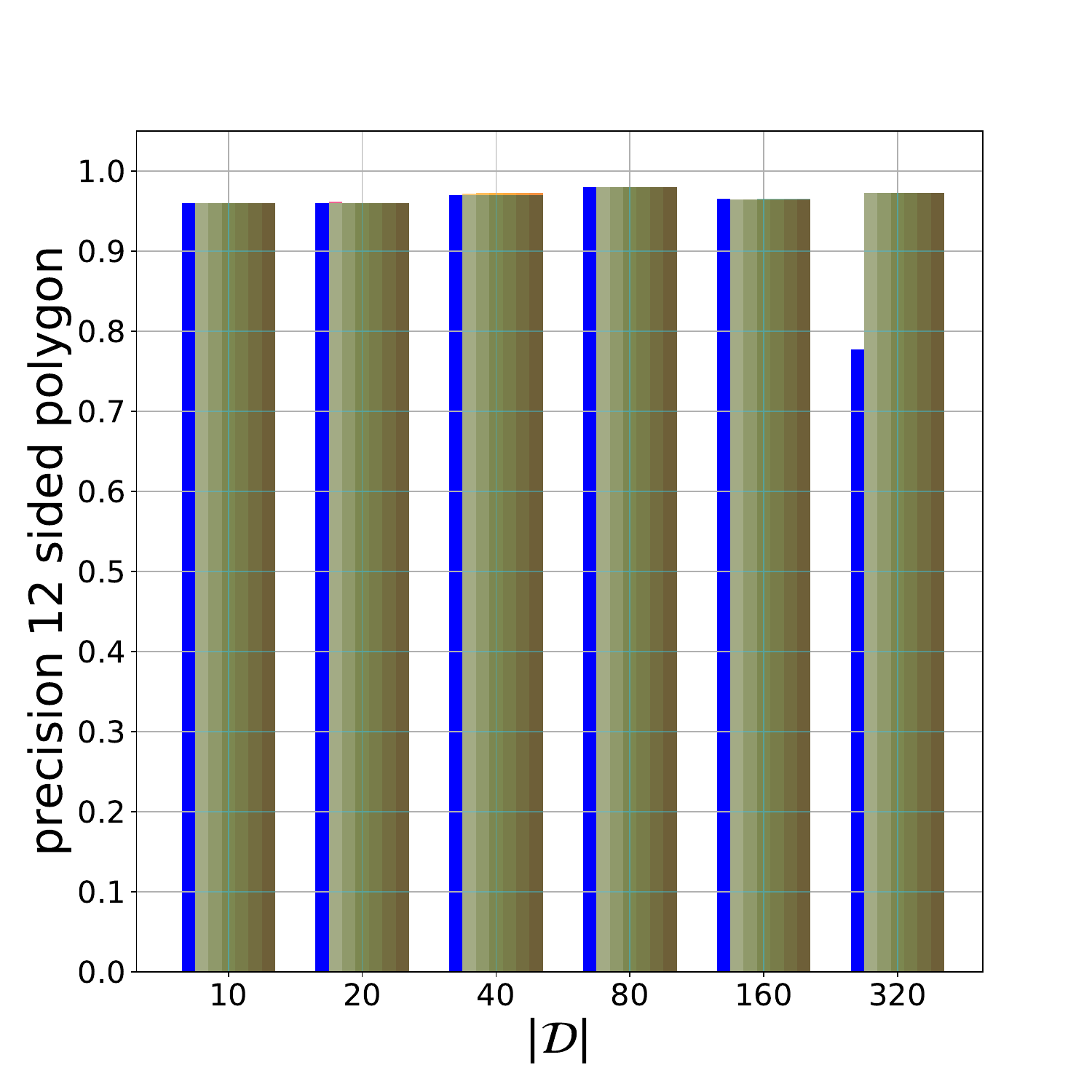}};
        \end{tikzpicture}
    \end{subfigure}
    \begin{subfigure}[b]{0.24\textwidth}
        \begin{tikzpicture}
            \node (figure) at (0,0) {\includegraphics[width=\textwidth]{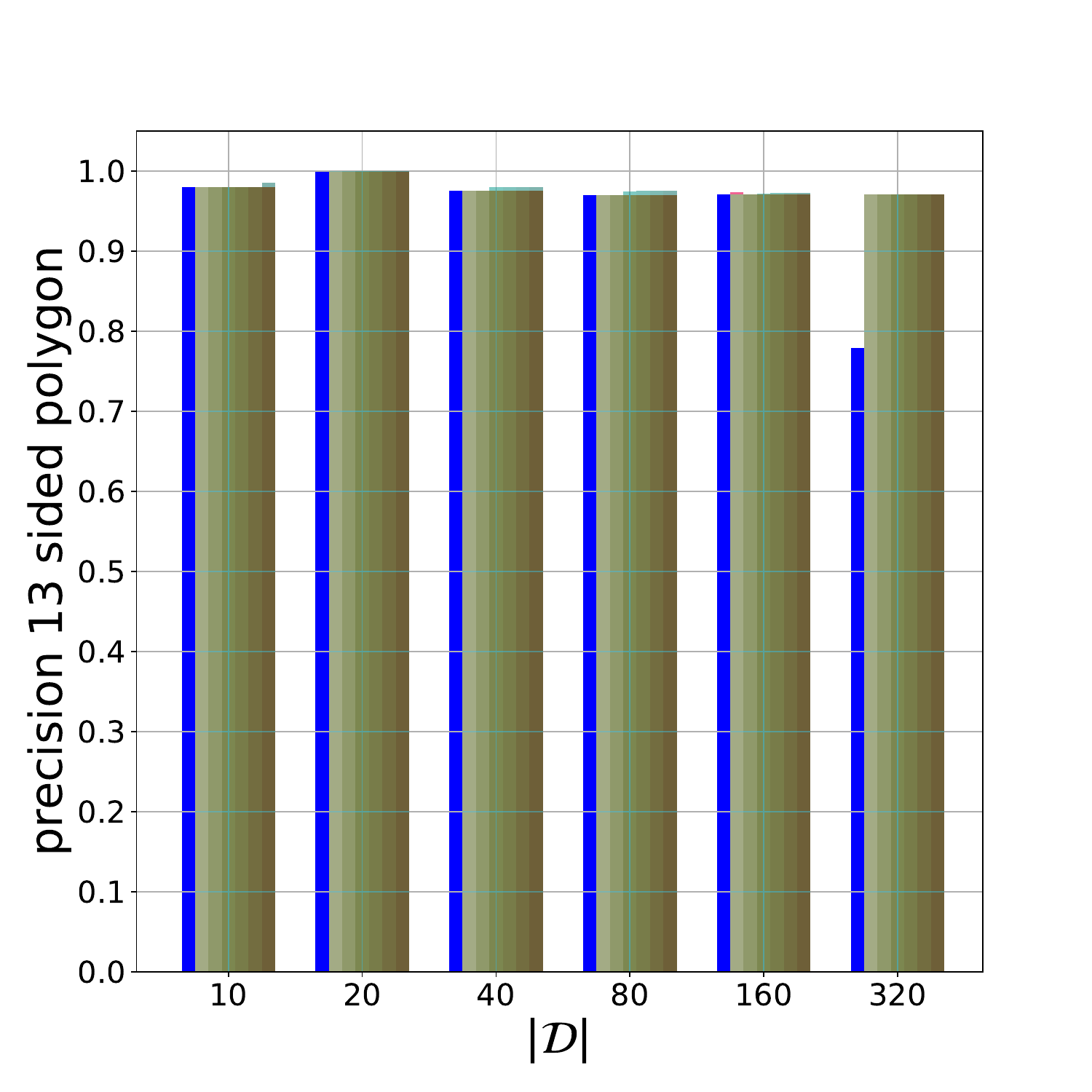}};
        \end{tikzpicture}
    \end{subfigure}
    \begin{subfigure}[b]{0.24\textwidth}
        \begin{tikzpicture}
            \node (figure) at (0,0) {\includegraphics[width=\textwidth]{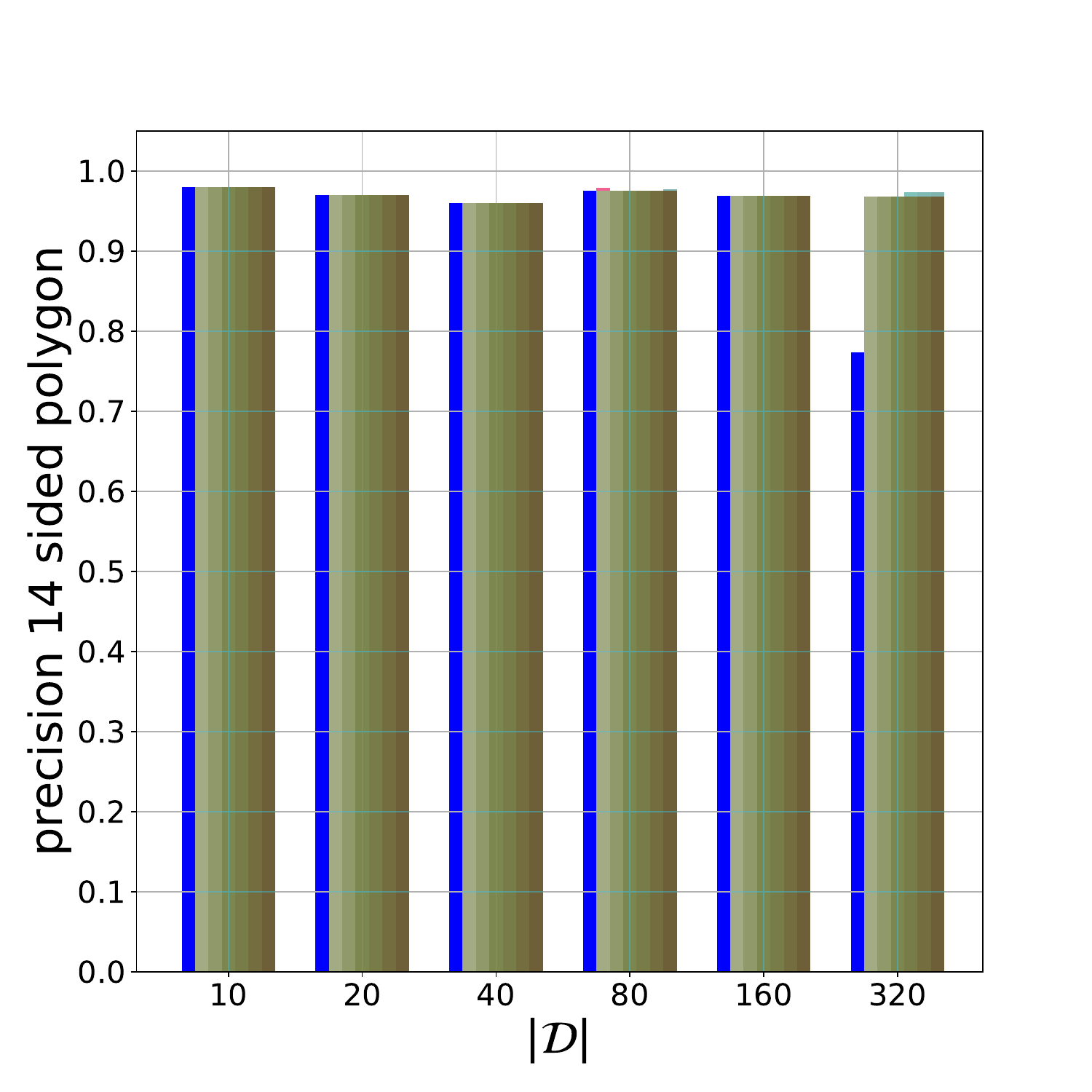}};
        \end{tikzpicture}
    \end{subfigure}

    \begin{subfigure}[b]{0.24\textwidth}
        \begin{tikzpicture}
            \node (figure) at (0,0) {\includegraphics[width=\textwidth]{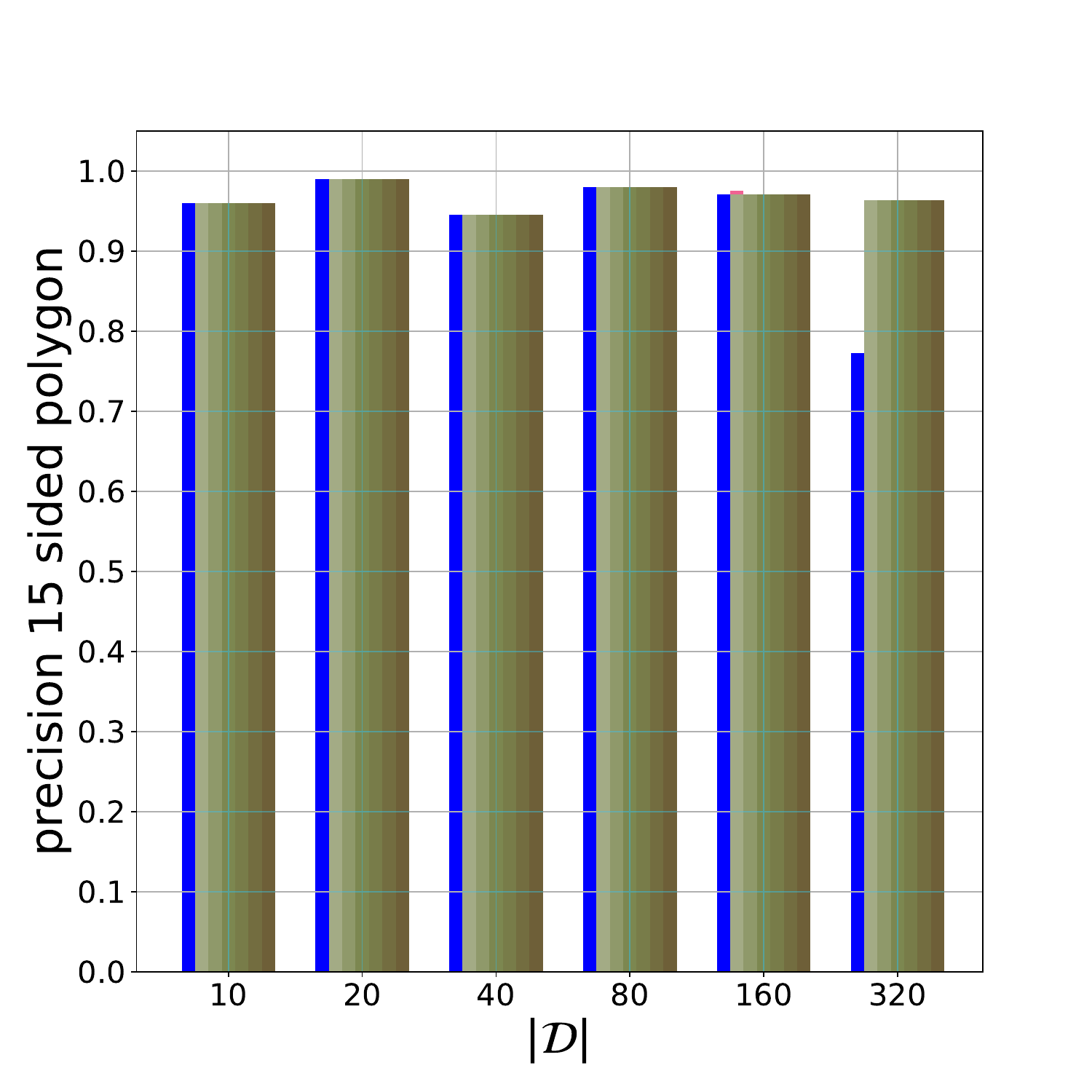}};
        \end{tikzpicture}
    \end{subfigure}
    \begin{subfigure}[b]{0.24\textwidth}
        \begin{tikzpicture}
            \node (figure) at (0,0) {\includegraphics[width=\textwidth]{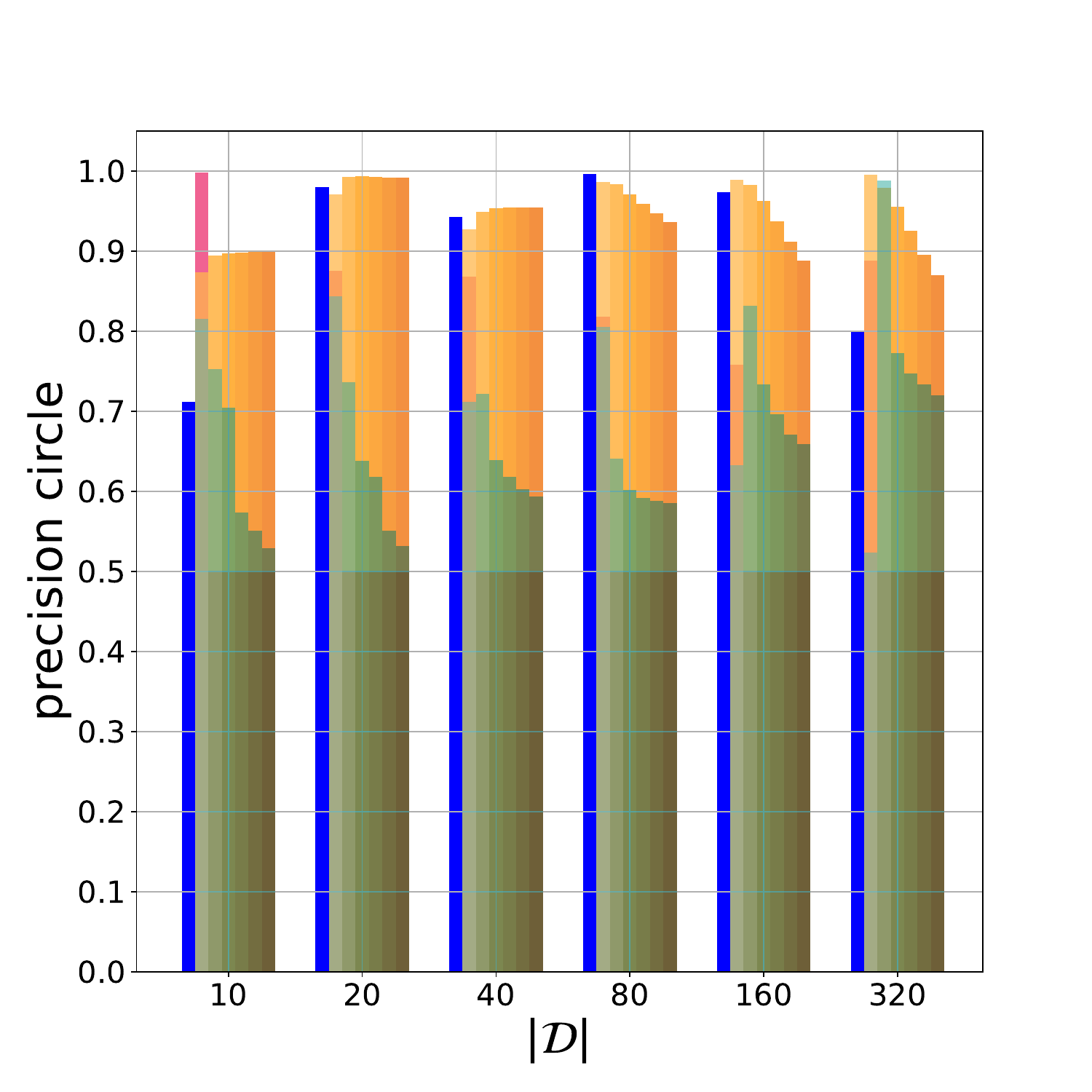}};
        \end{tikzpicture}
    \end{subfigure}
\end{figure}

\newpage

\begin{figure}[ht]
    \centering

    \begin{subfigure}[b]{0.24\textwidth}
        \begin{tikzpicture}
            \node (figure) at (0,0) {\includegraphics[width=\textwidth]{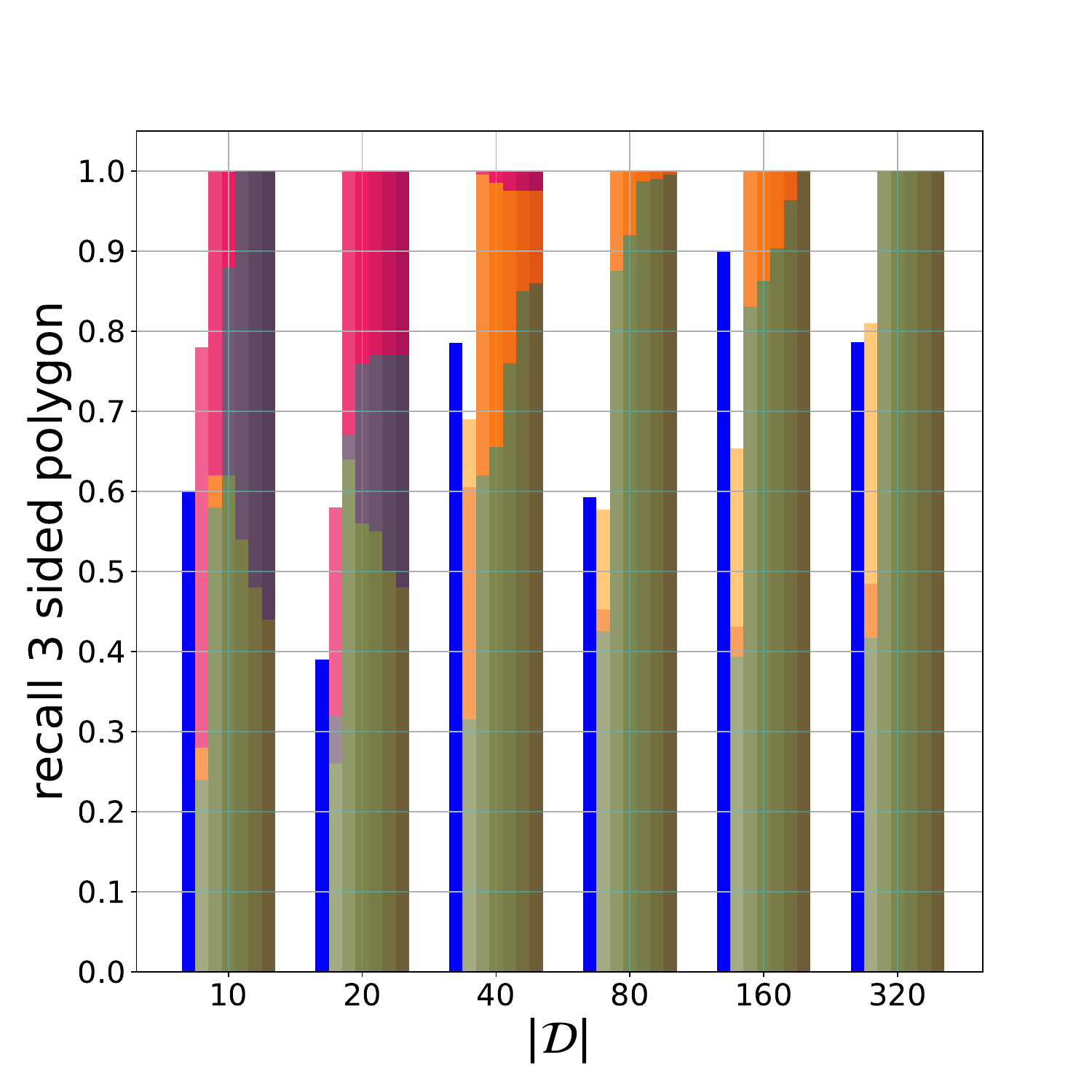}};
        \end{tikzpicture}
    \end{subfigure}
    \begin{subfigure}[b]{0.24\textwidth}
        \begin{tikzpicture}
            \node (figure) at (0,0) {\includegraphics[width=\textwidth]{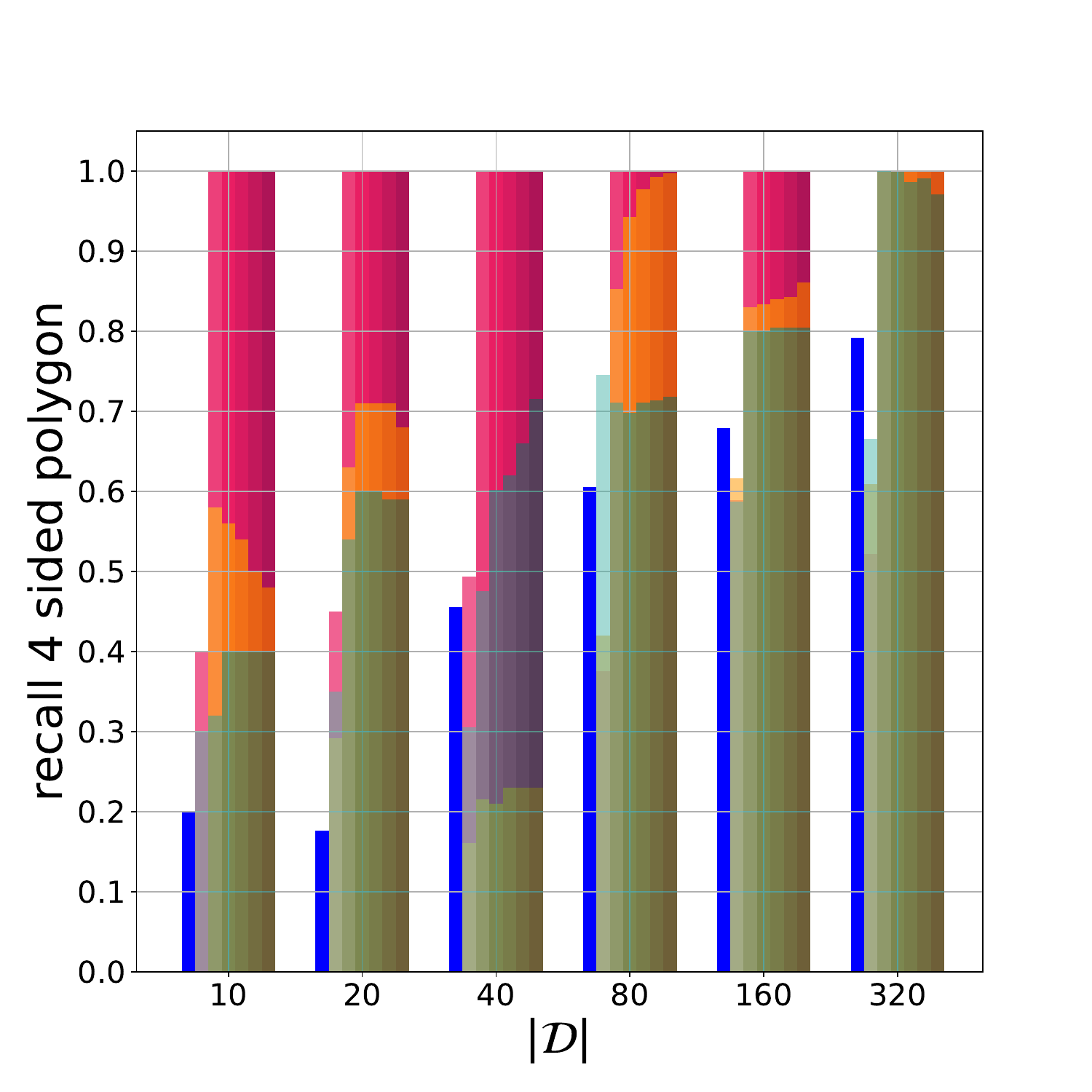}};
        \end{tikzpicture}
    \end{subfigure}
    \begin{subfigure}[b]{0.24\textwidth}
        \begin{tikzpicture}
            \node (figure) at (0,0) {\includegraphics[width=\textwidth]{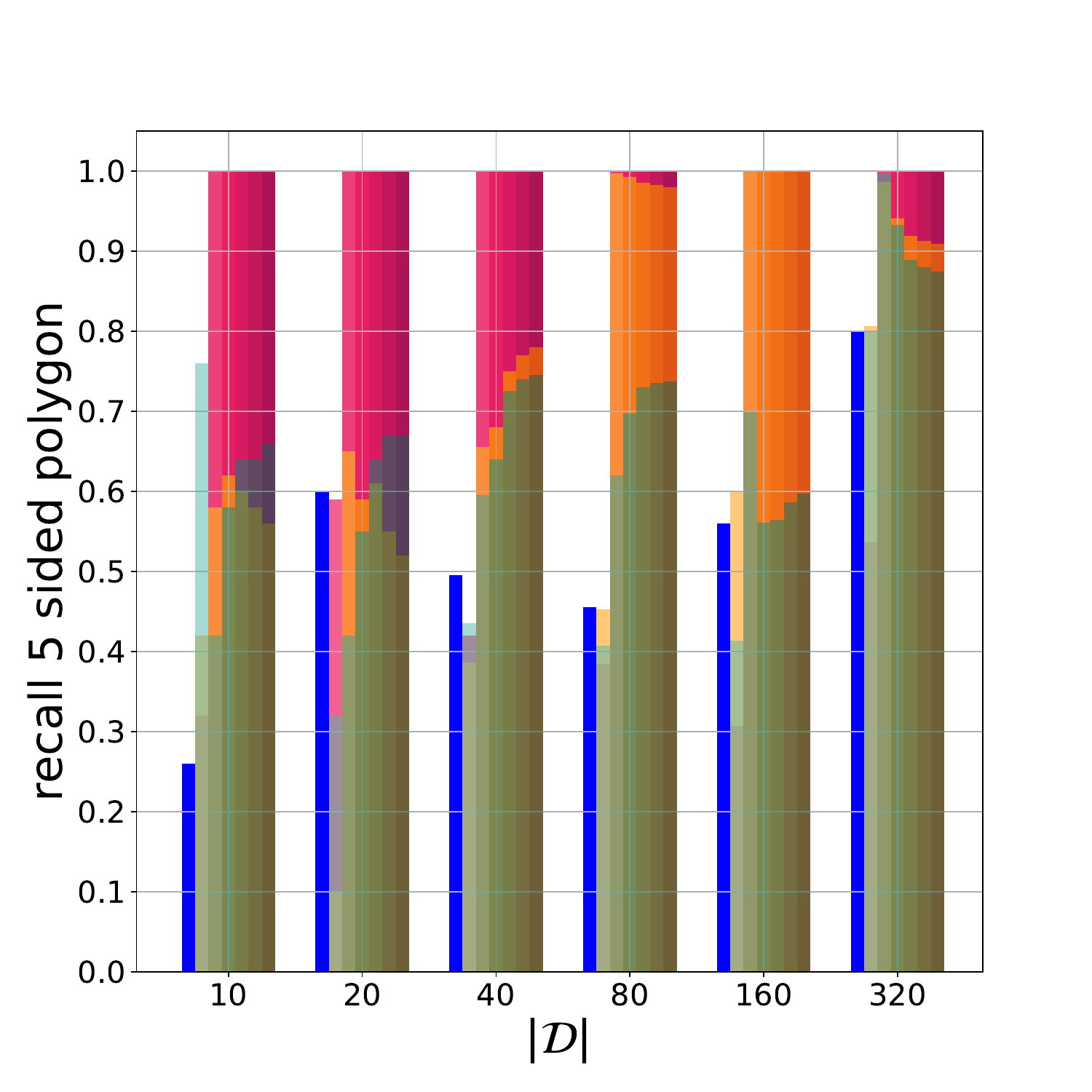}};
        \end{tikzpicture}
    \end{subfigure}
    \begin{subfigure}[b]{0.24\textwidth}
        \begin{tikzpicture}
            \node (figure) at (0,0) {\includegraphics[width=\textwidth]{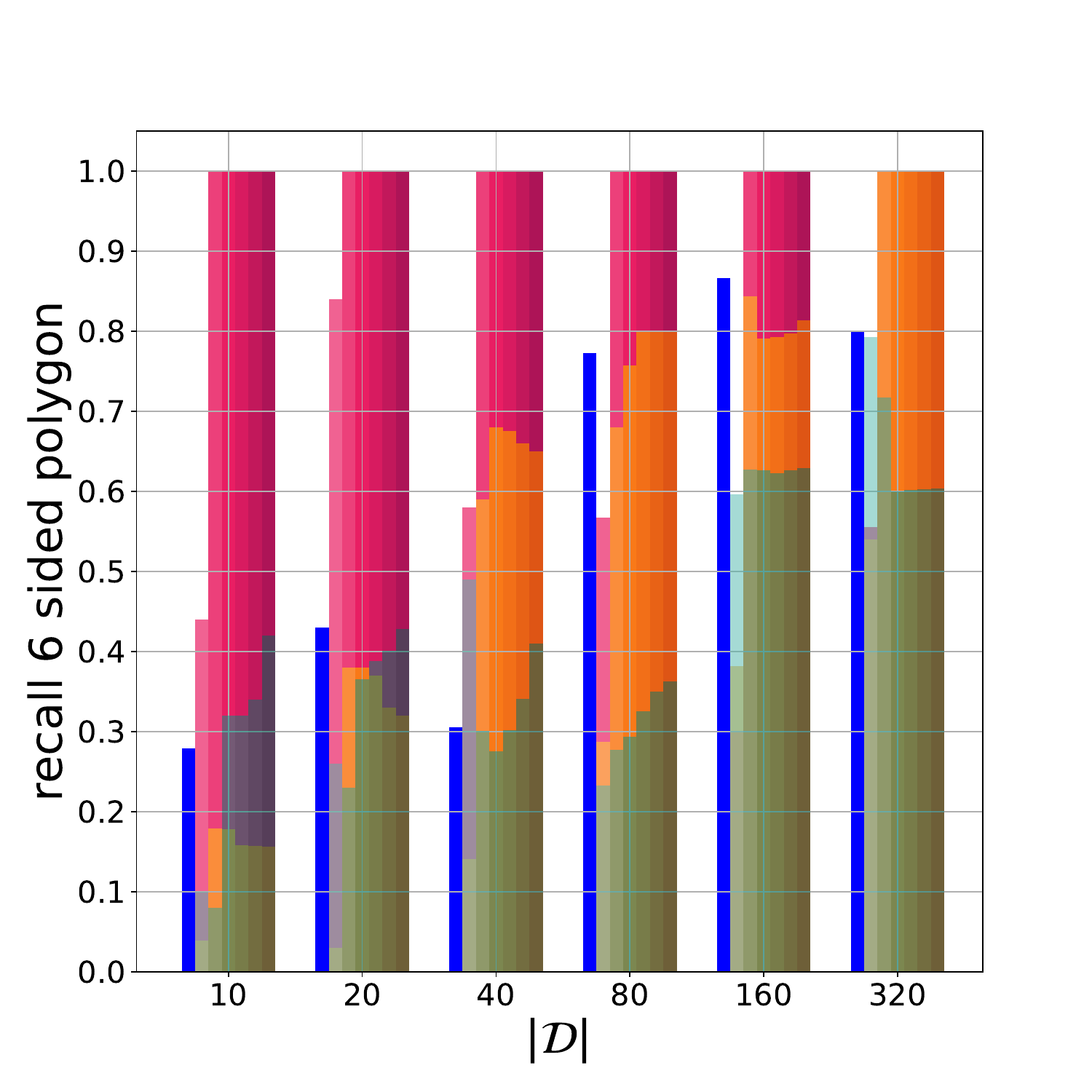}};
        \end{tikzpicture}
    \end{subfigure}

    \begin{subfigure}[b]{0.24\textwidth}
        \begin{tikzpicture}
            \node (figure) at (0,0) {\includegraphics[width=\textwidth]{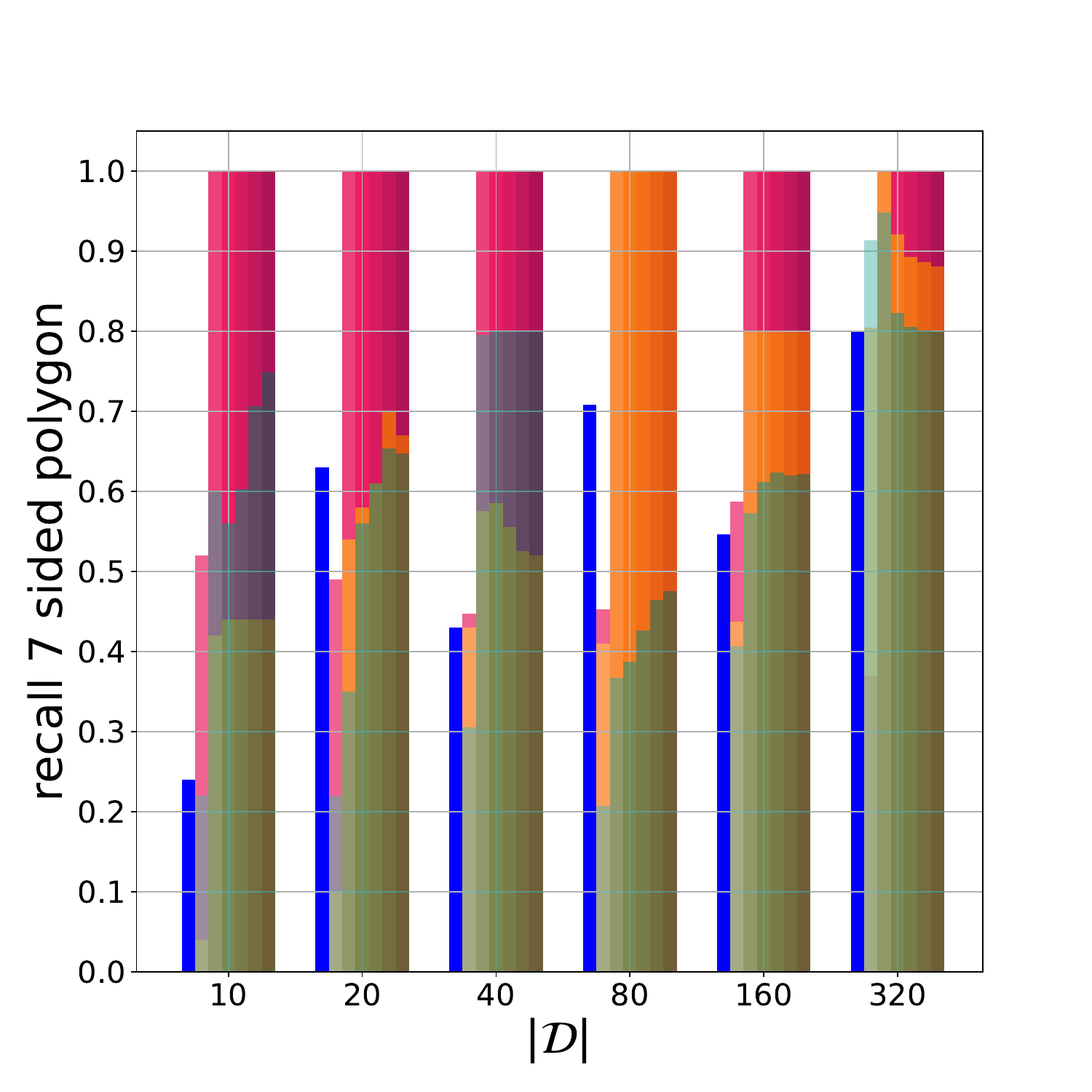}};
        \end{tikzpicture}
    \end{subfigure}
    \begin{subfigure}[b]{0.24\textwidth}
        \begin{tikzpicture}
            \node (figure) at (0,0) {\includegraphics[width=\textwidth]{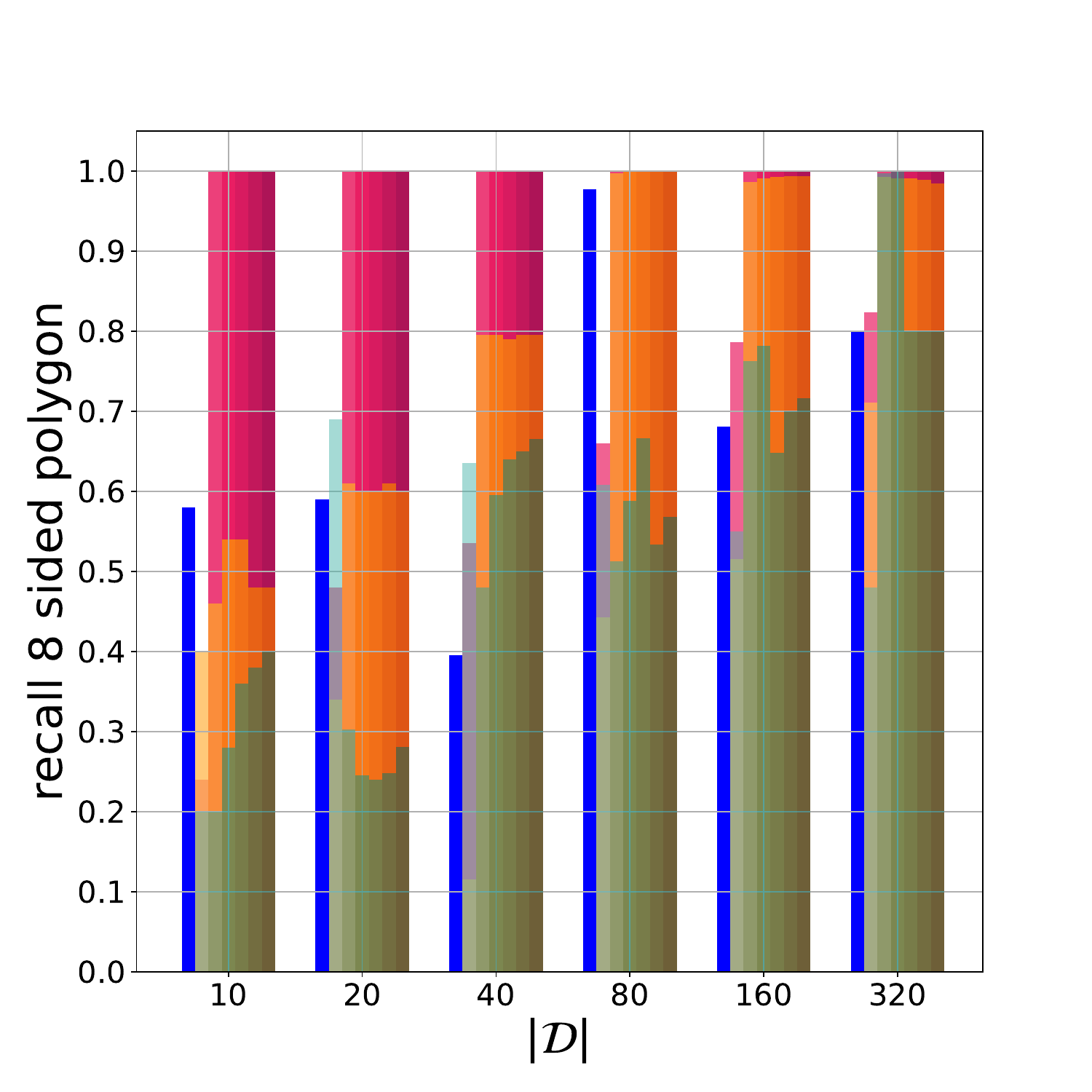}};
        \end{tikzpicture}
    \end{subfigure}
    \begin{subfigure}[b]{0.24\textwidth}
        \begin{tikzpicture}
            \node (figure) at (0,0) {\includegraphics[width=\textwidth]{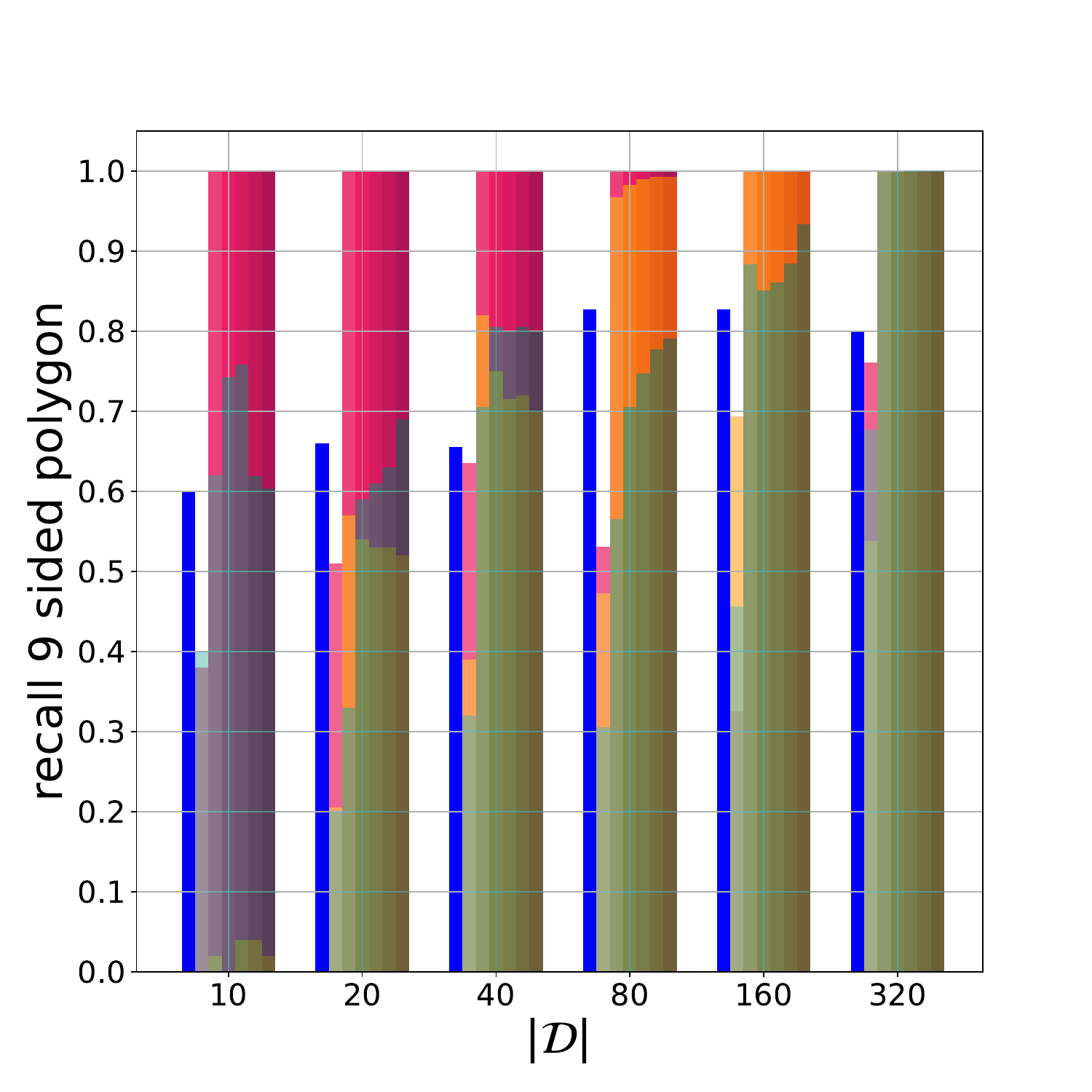}};
        \end{tikzpicture}
    \end{subfigure}
    \begin{subfigure}[b]{0.24\textwidth}
        \begin{tikzpicture}
            \node (figure) at (0,0) {\includegraphics[width=\textwidth]{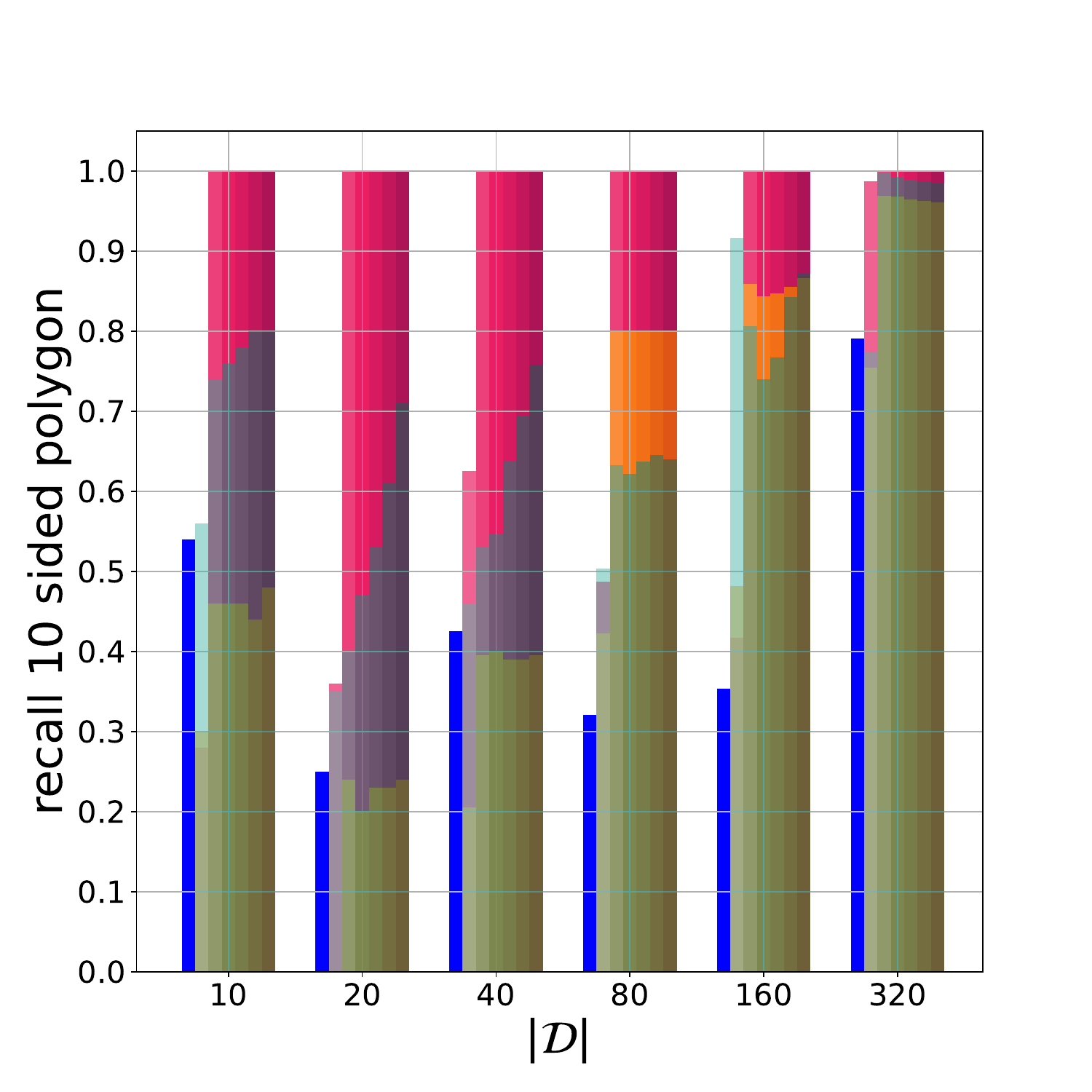}};
        \end{tikzpicture}
    \end{subfigure}

    \begin{subfigure}[b]{0.24\textwidth}
        \begin{tikzpicture}
            \node (figure) at (0,0) {\includegraphics[width=\textwidth]{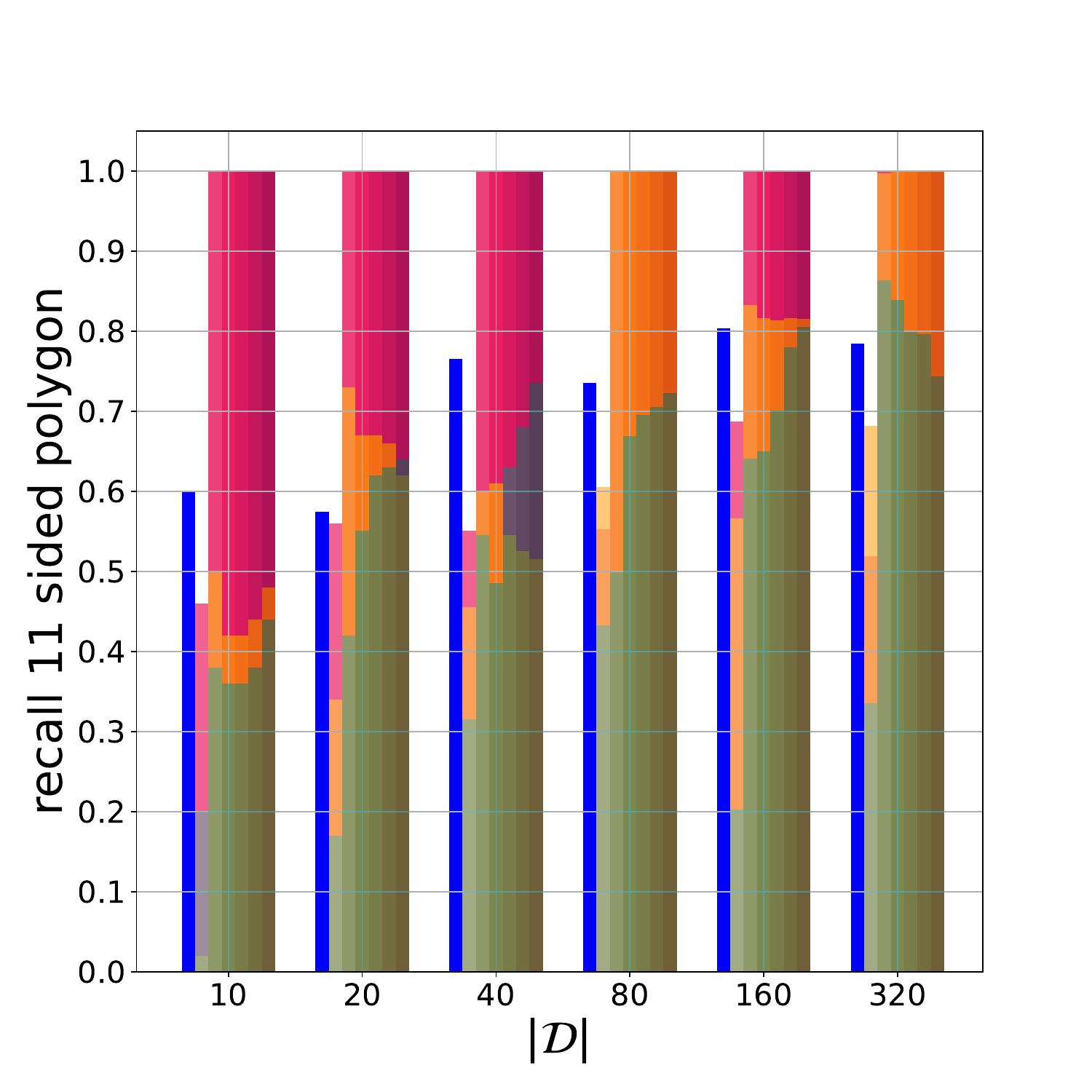}};
        \end{tikzpicture}
    \end{subfigure}
    \begin{subfigure}[b]{0.24\textwidth}
        \begin{tikzpicture}
            \node (figure) at (0,0) {\includegraphics[width=\textwidth]{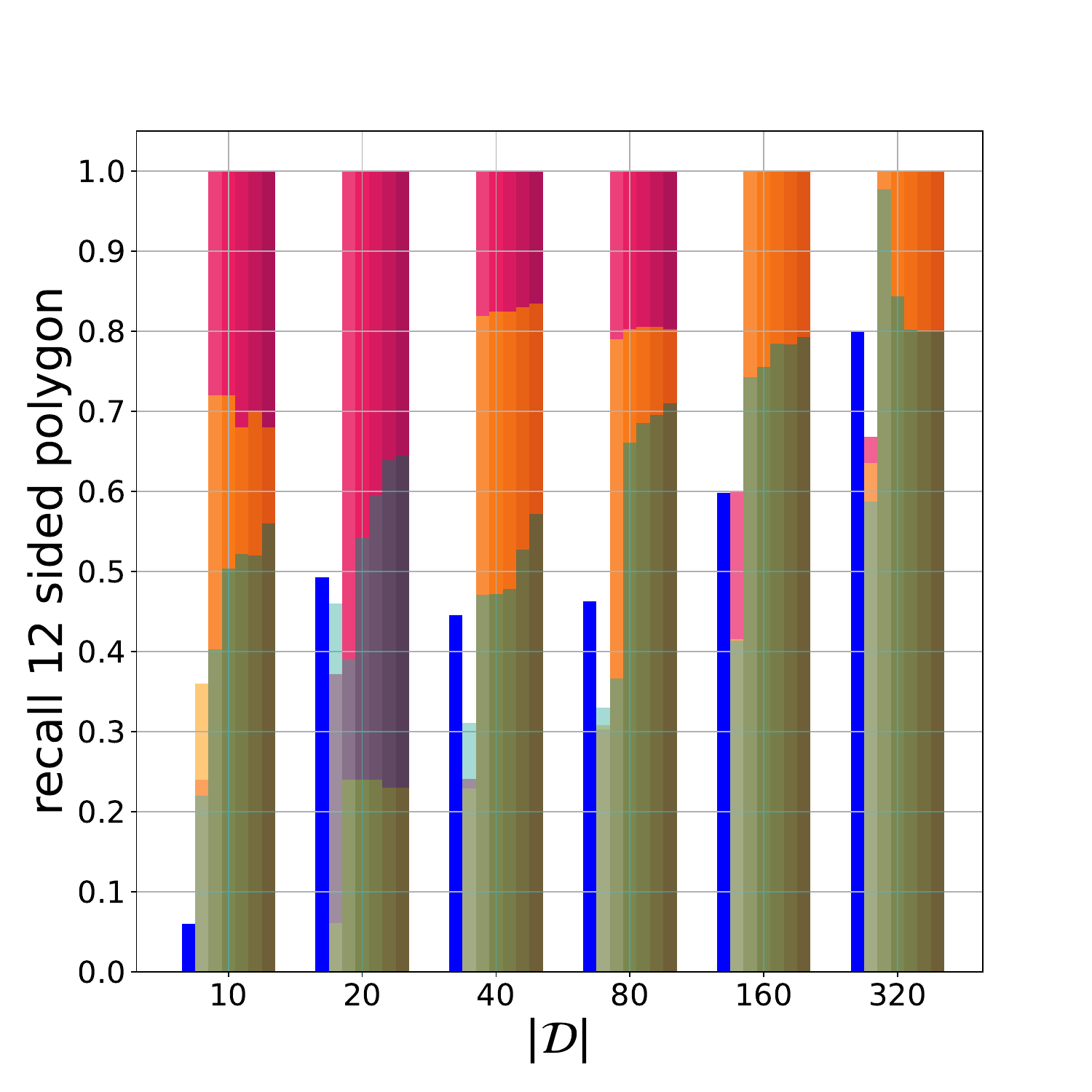}};
        \end{tikzpicture}
    \end{subfigure}
    \begin{subfigure}[b]{0.24\textwidth}
        \begin{tikzpicture}
            \node (figure) at (0,0) {\includegraphics[width=\textwidth]{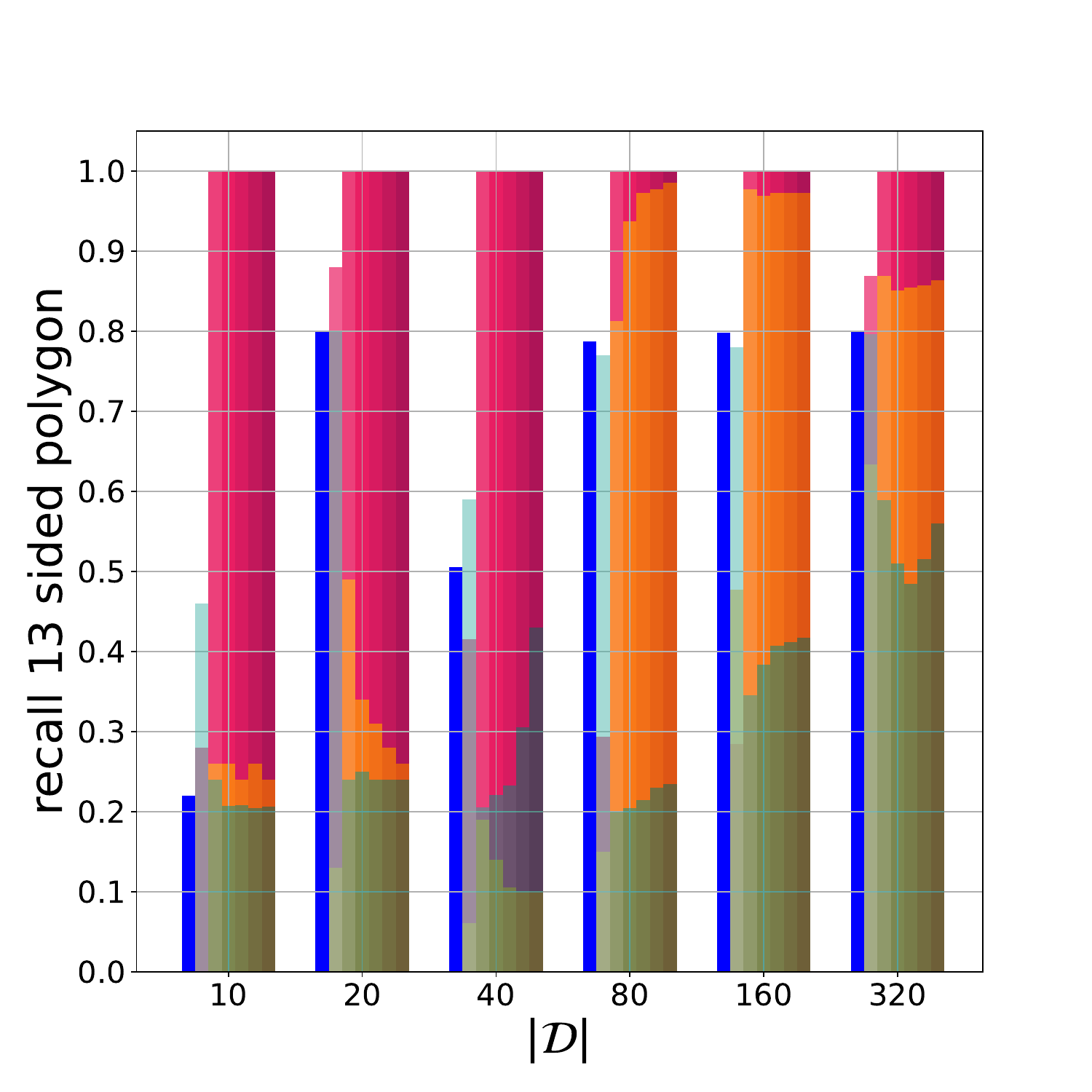}};
        \end{tikzpicture}
    \end{subfigure}
    \begin{subfigure}[b]{0.24\textwidth}
        \begin{tikzpicture}
            \node (figure) at (0,0) {\includegraphics[width=\textwidth]{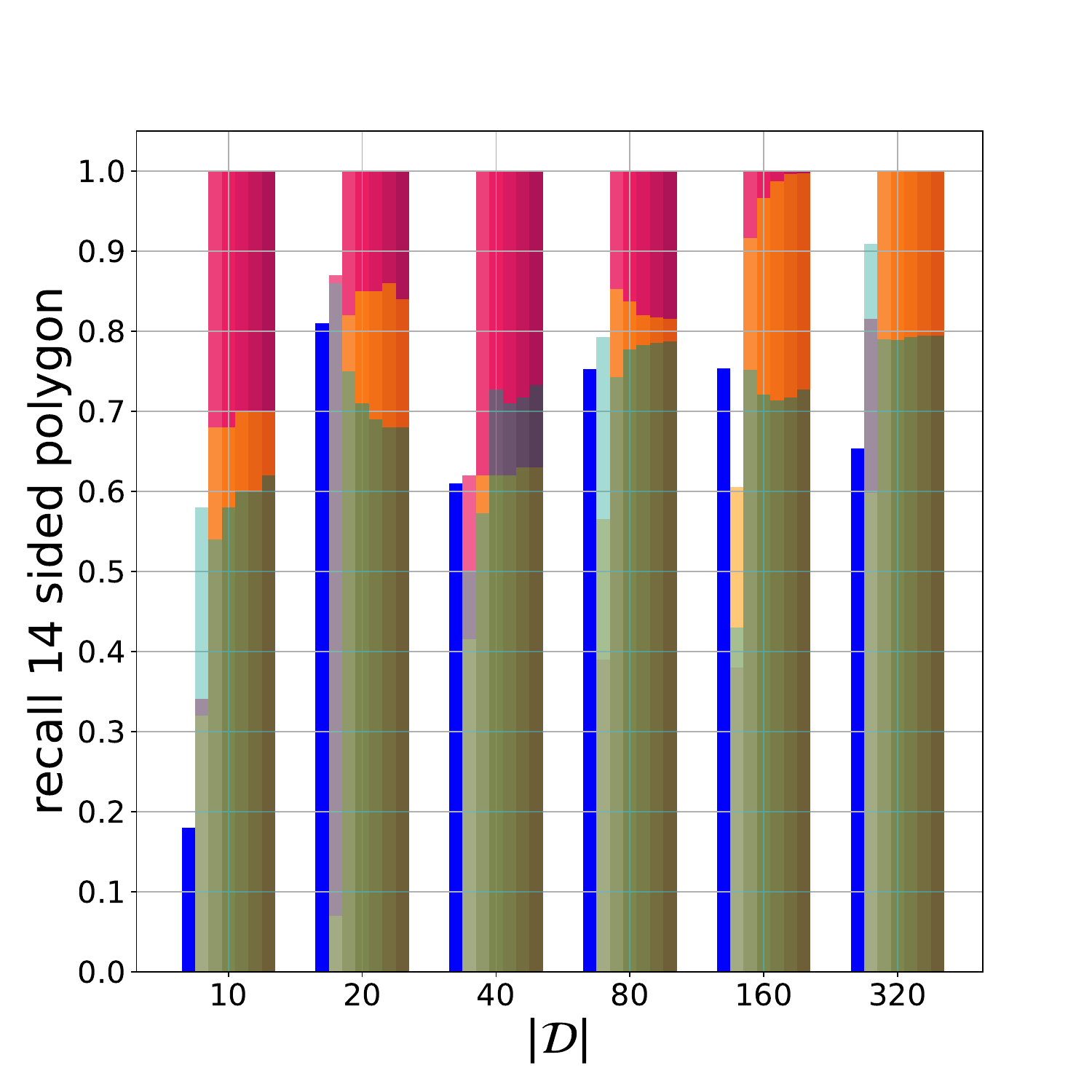}};
        \end{tikzpicture}
    \end{subfigure}

    \begin{subfigure}[b]{0.24\textwidth}
        \begin{tikzpicture}
            \node (figure) at (0,0) {\includegraphics[width=\textwidth]{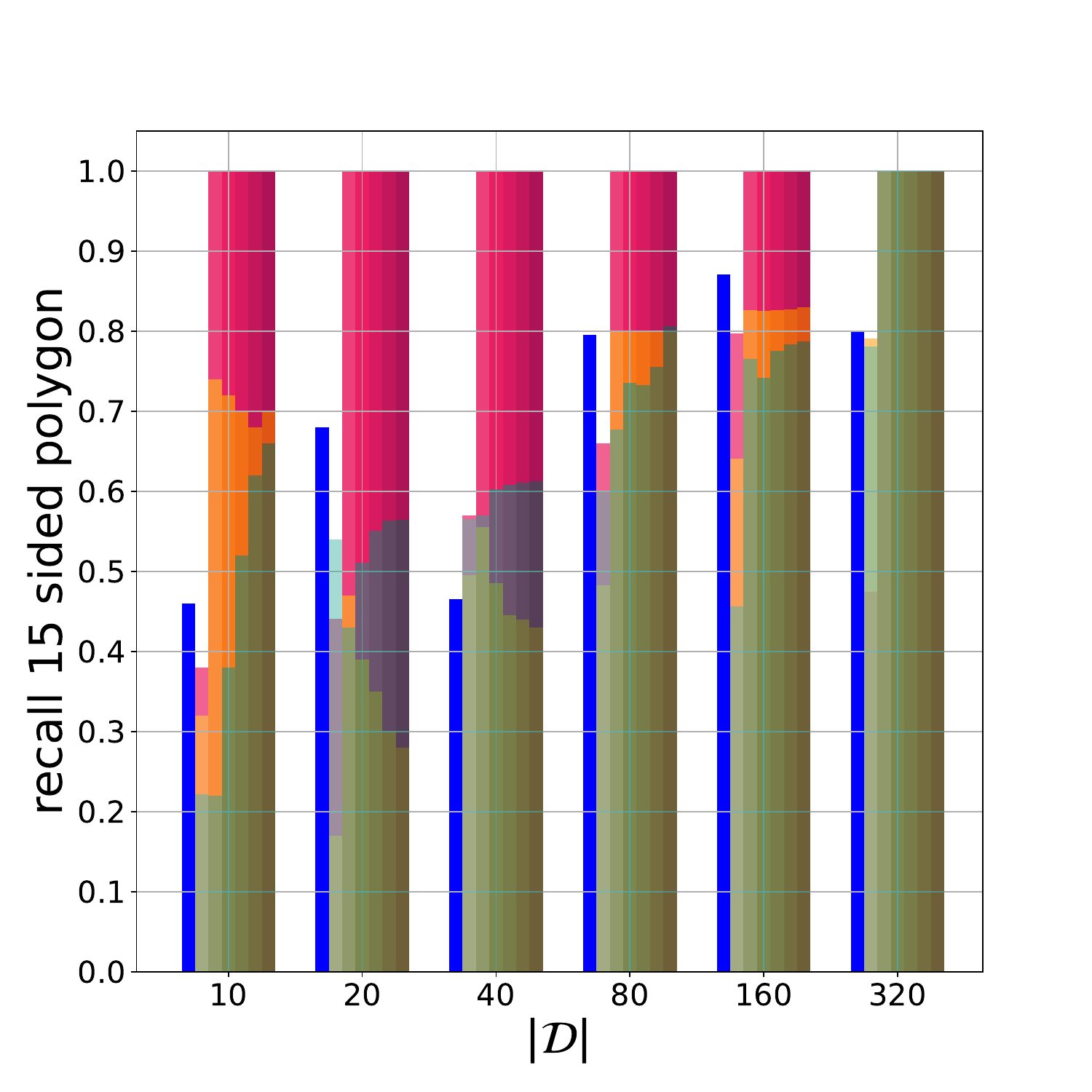}};
        \end{tikzpicture}
    \end{subfigure}
    \begin{subfigure}[b]{0.24\textwidth}
        \begin{tikzpicture}
            \node (figure) at (0,0) {\includegraphics[width=\textwidth]{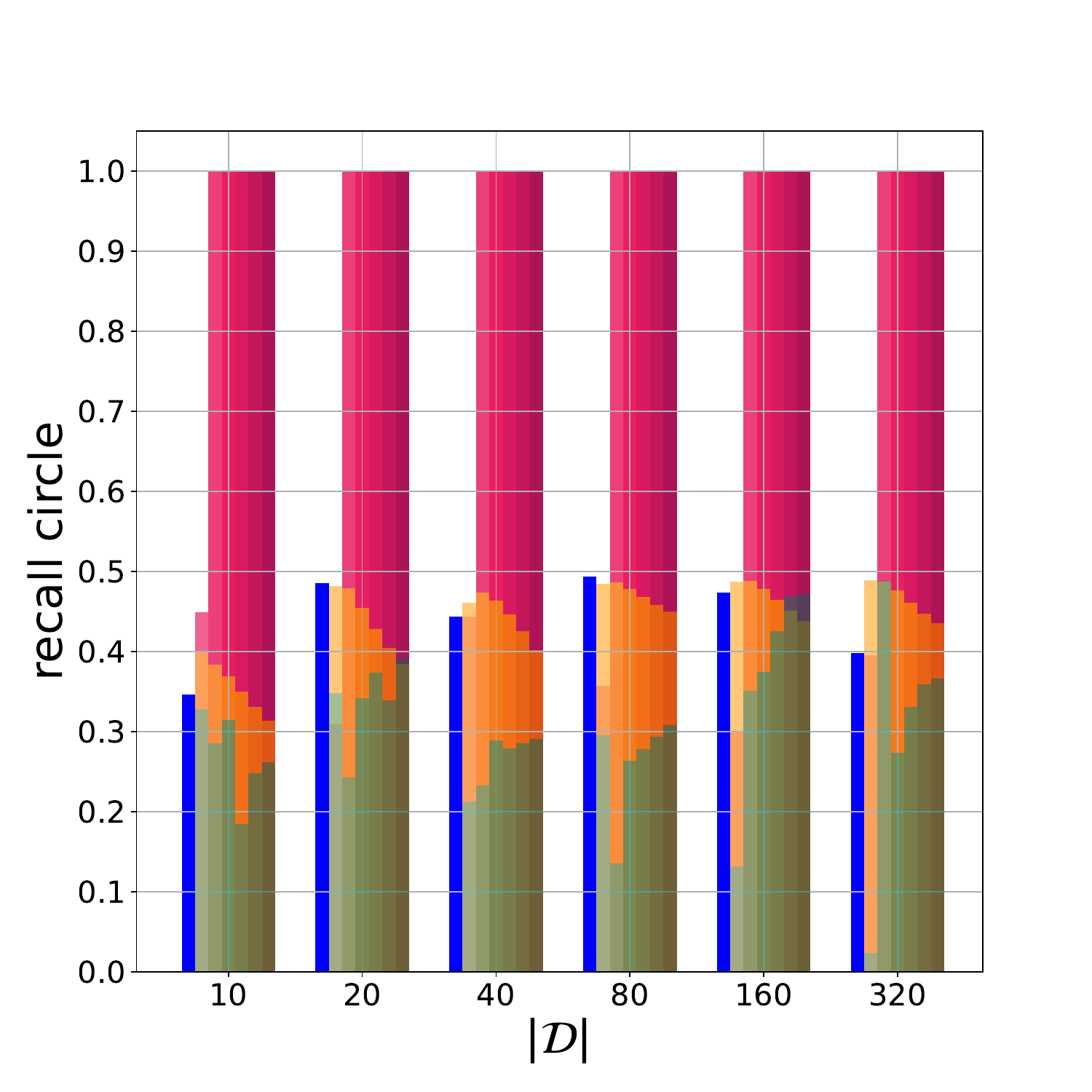}};
        \end{tikzpicture}
    \end{subfigure}
\end{figure}

%% file: results/noise/cad.tex
\begin{figure}[ht]
    \centering

    \begin{subfigure}[b]{0.24\textwidth}
        \begin{tikzpicture}
            \node (figure) at (0,0) {\includegraphics[width=\textwidth]{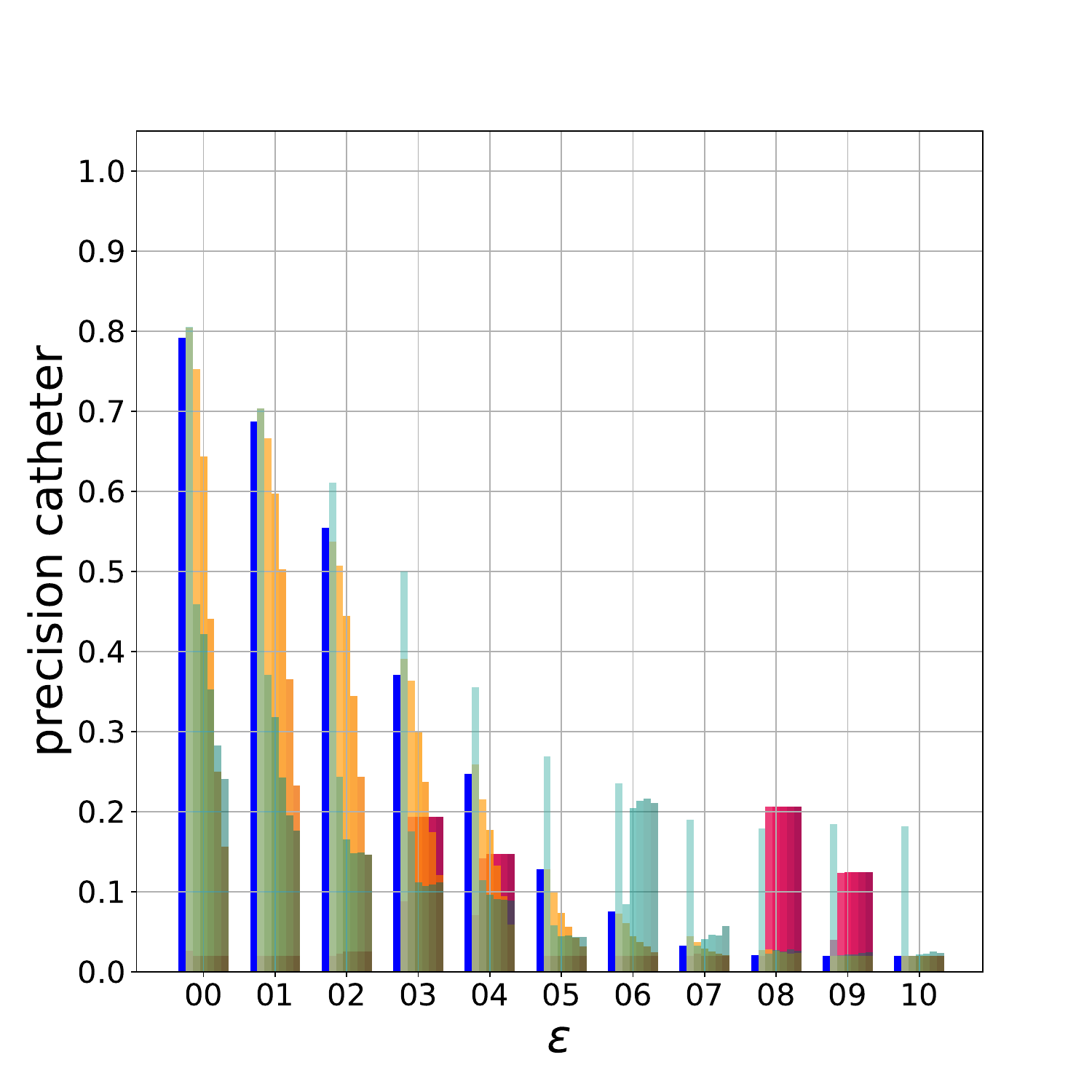}};
        \end{tikzpicture}
    \end{subfigure}
    \begin{subfigure}[b]{0.24\textwidth}
        \begin{tikzpicture}
            \node (figure) at (0,0) {\includegraphics[width=\textwidth]{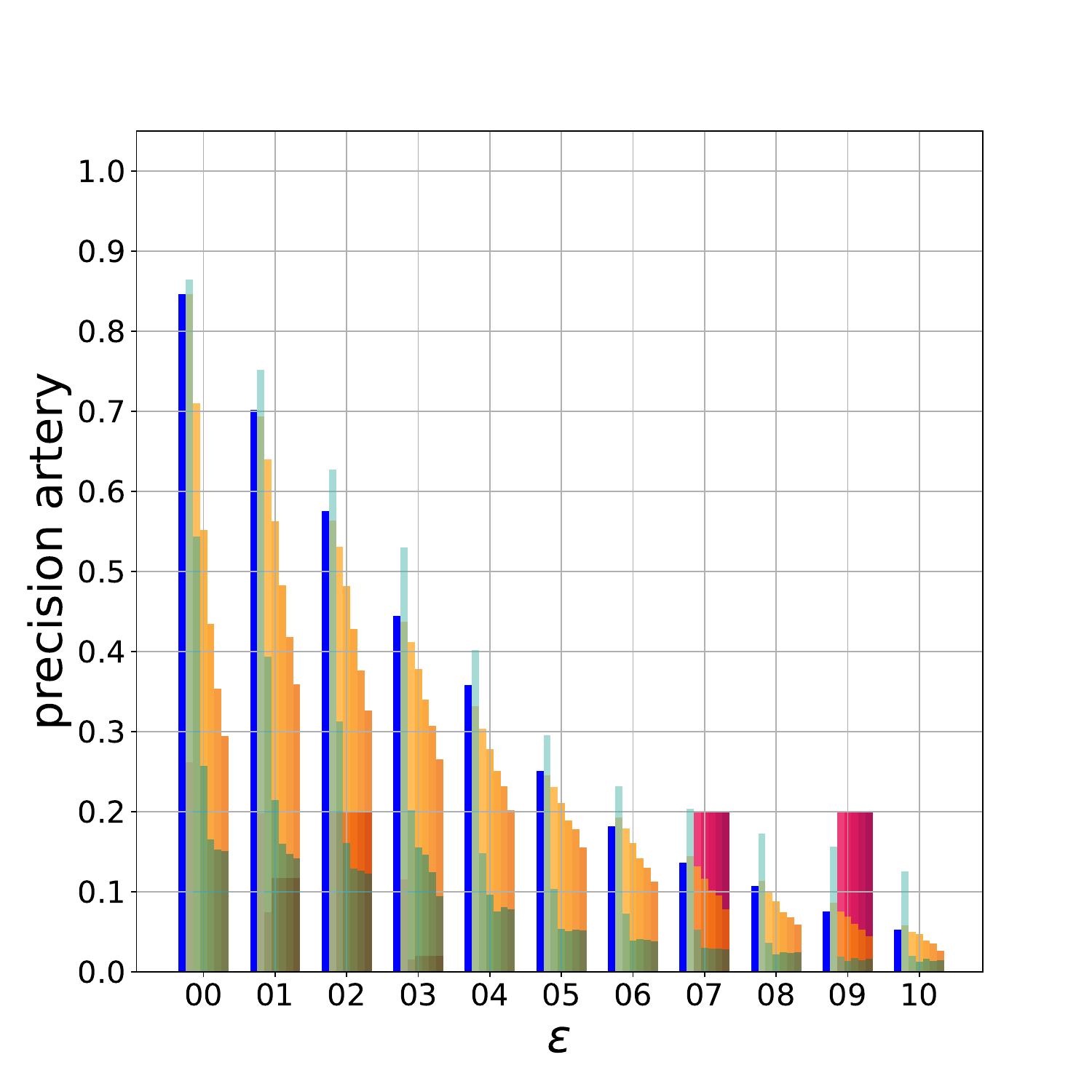}};
        \end{tikzpicture}
    \end{subfigure}

    \begin{subfigure}[b]{0.24\textwidth}
        \begin{tikzpicture}
            \node (figure) at (0,0) {\includegraphics[width=\textwidth]{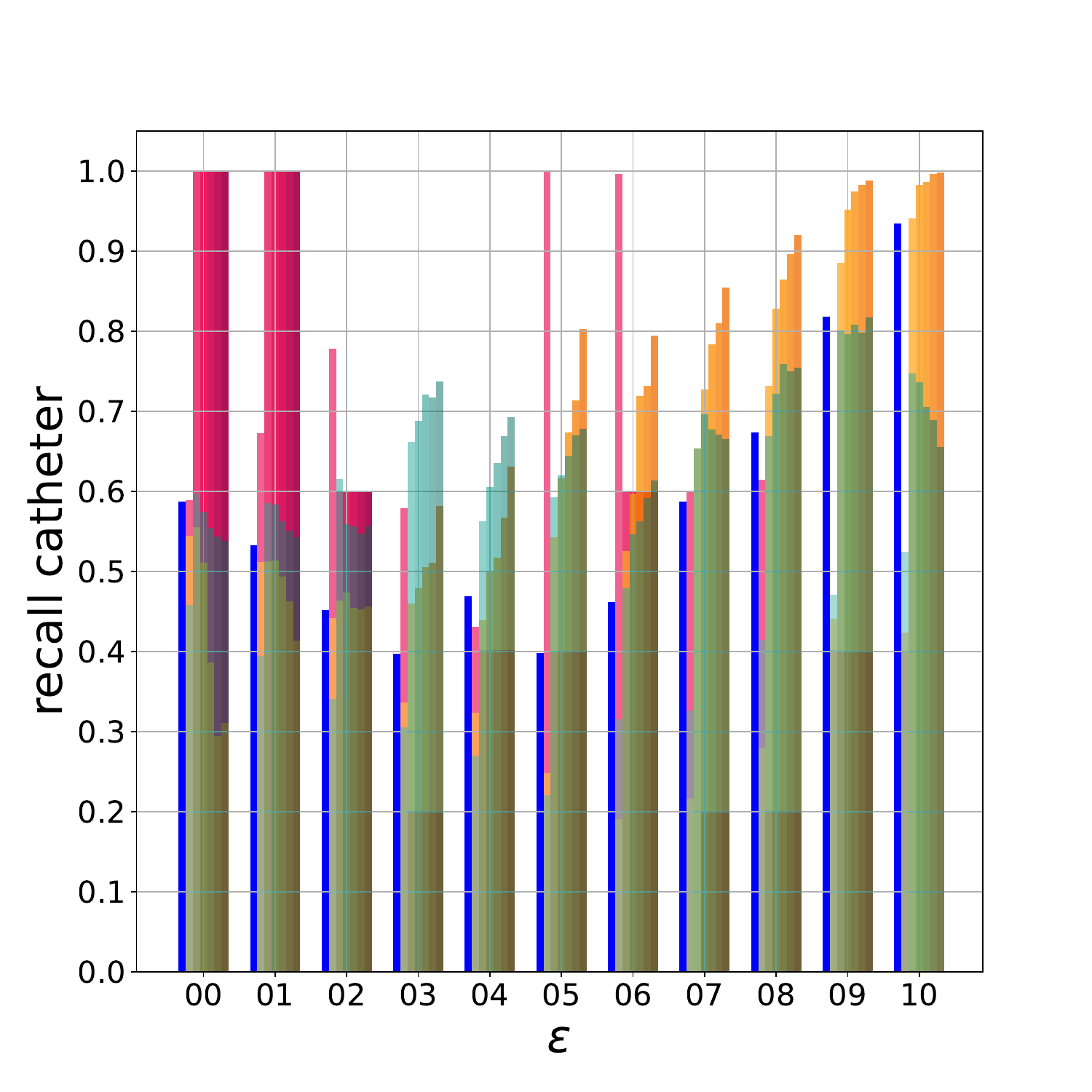}};
        \end{tikzpicture}
    \end{subfigure}
    \begin{subfigure}[b]{0.24\textwidth}
        \begin{tikzpicture}
            \node (figure) at (0,0) {\includegraphics[width=\textwidth]{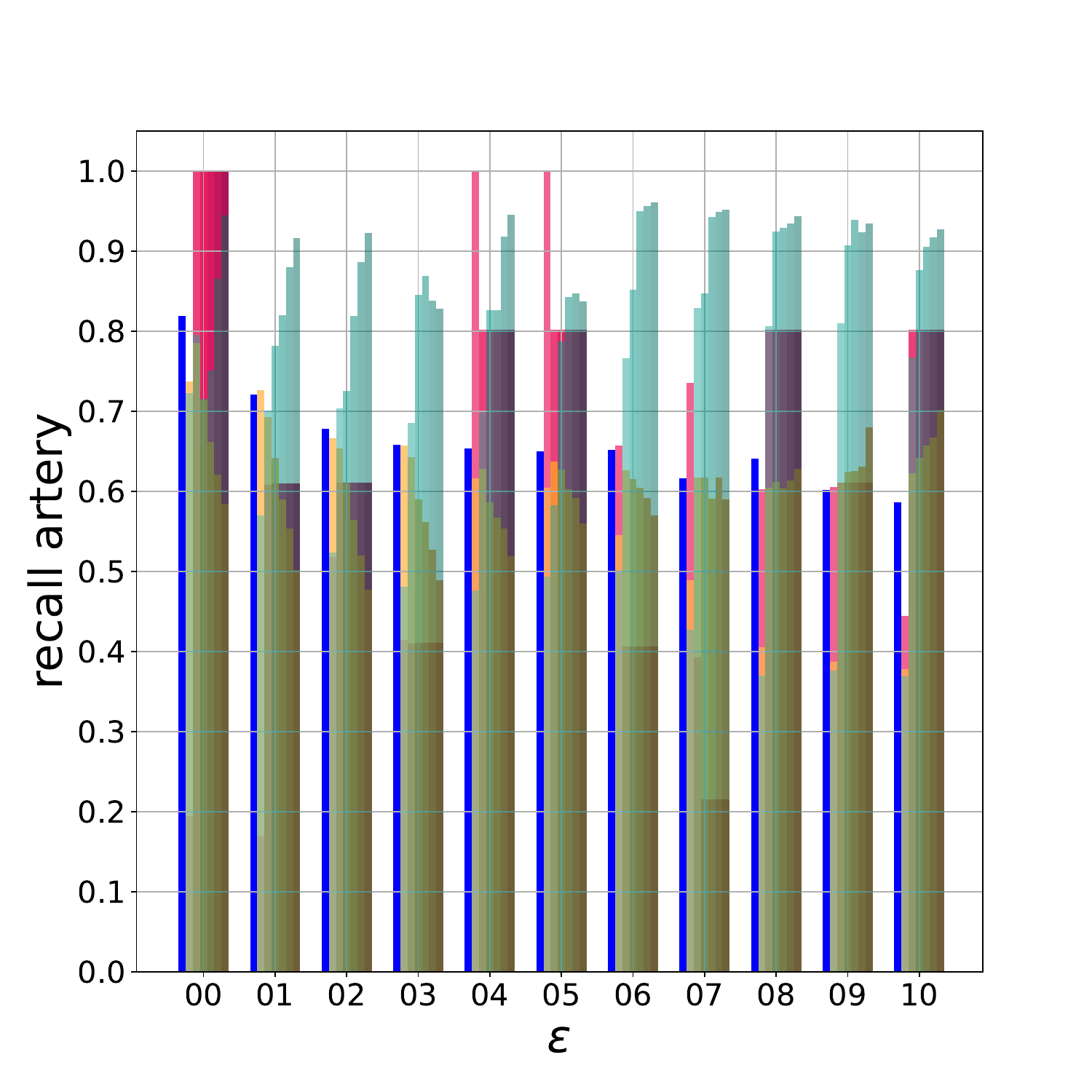}};
        \end{tikzpicture}
    \end{subfigure}
\end{figure}

%% file: results/examples/cad.tex
\begin{figure}[ht]
    \centering

    \begin{subfigure}[b]{0.24\textwidth}
        \begin{tikzpicture}
            \node (figure) at (0,0) {\includegraphics[width=\textwidth]{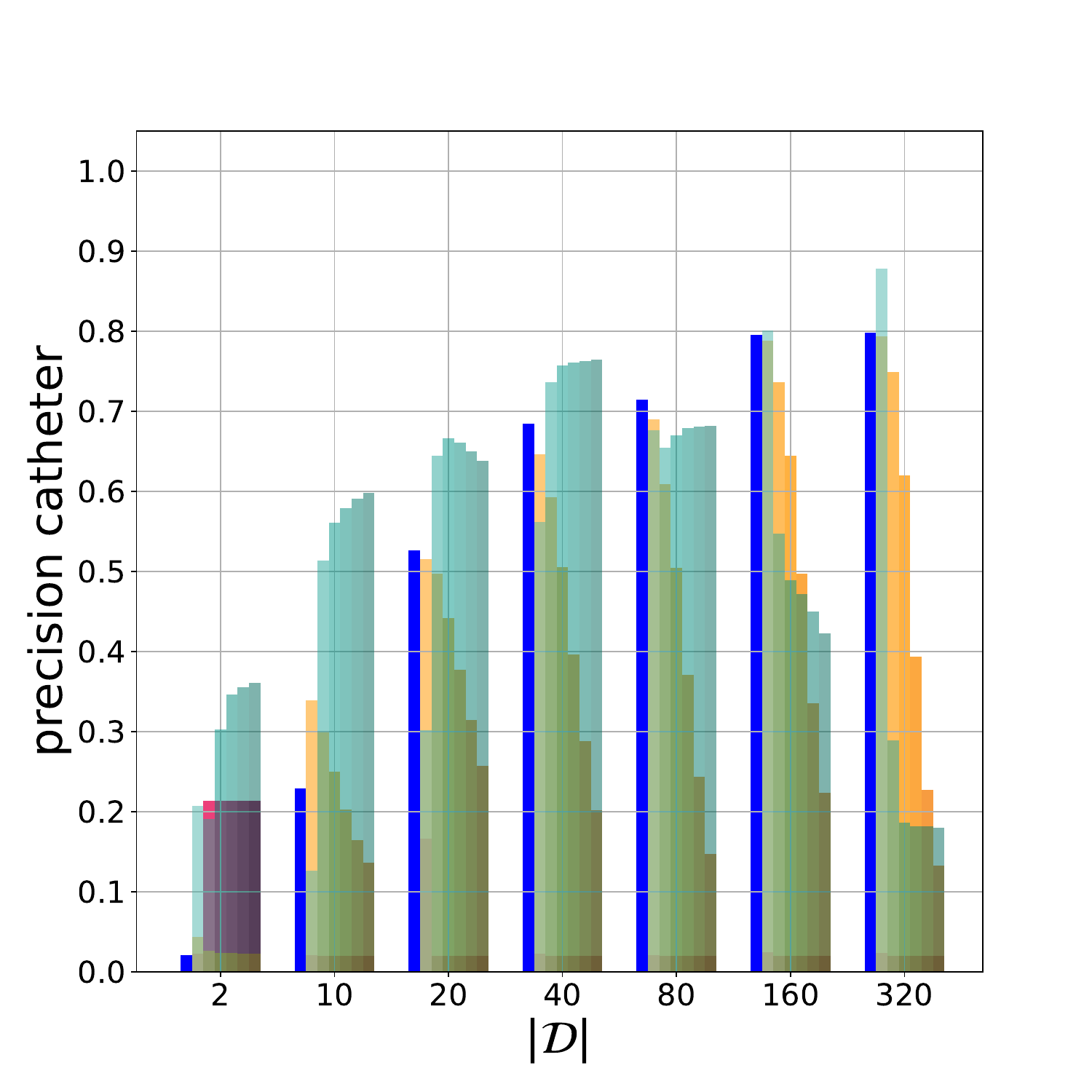}};
        \end{tikzpicture}
    \end{subfigure}
    \begin{subfigure}[b]{0.24\textwidth}
        \begin{tikzpicture}
            \node (figure) at (0,0) {\includegraphics[width=\textwidth]{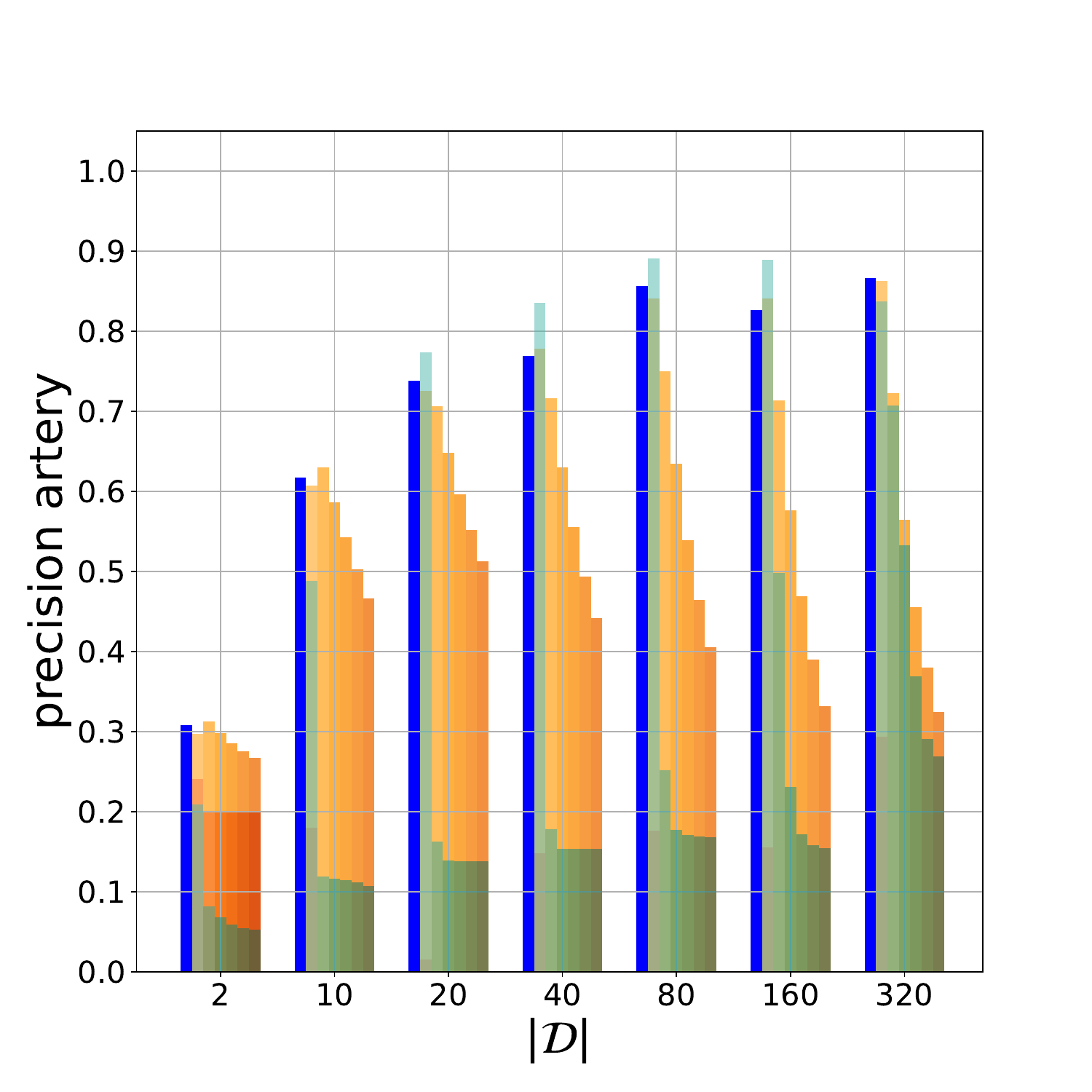}};
        \end{tikzpicture}
    \end{subfigure}

    \begin{subfigure}[b]{0.24\textwidth}
        \begin{tikzpicture}
            \node (figure) at (0,0) {\includegraphics[width=\textwidth]{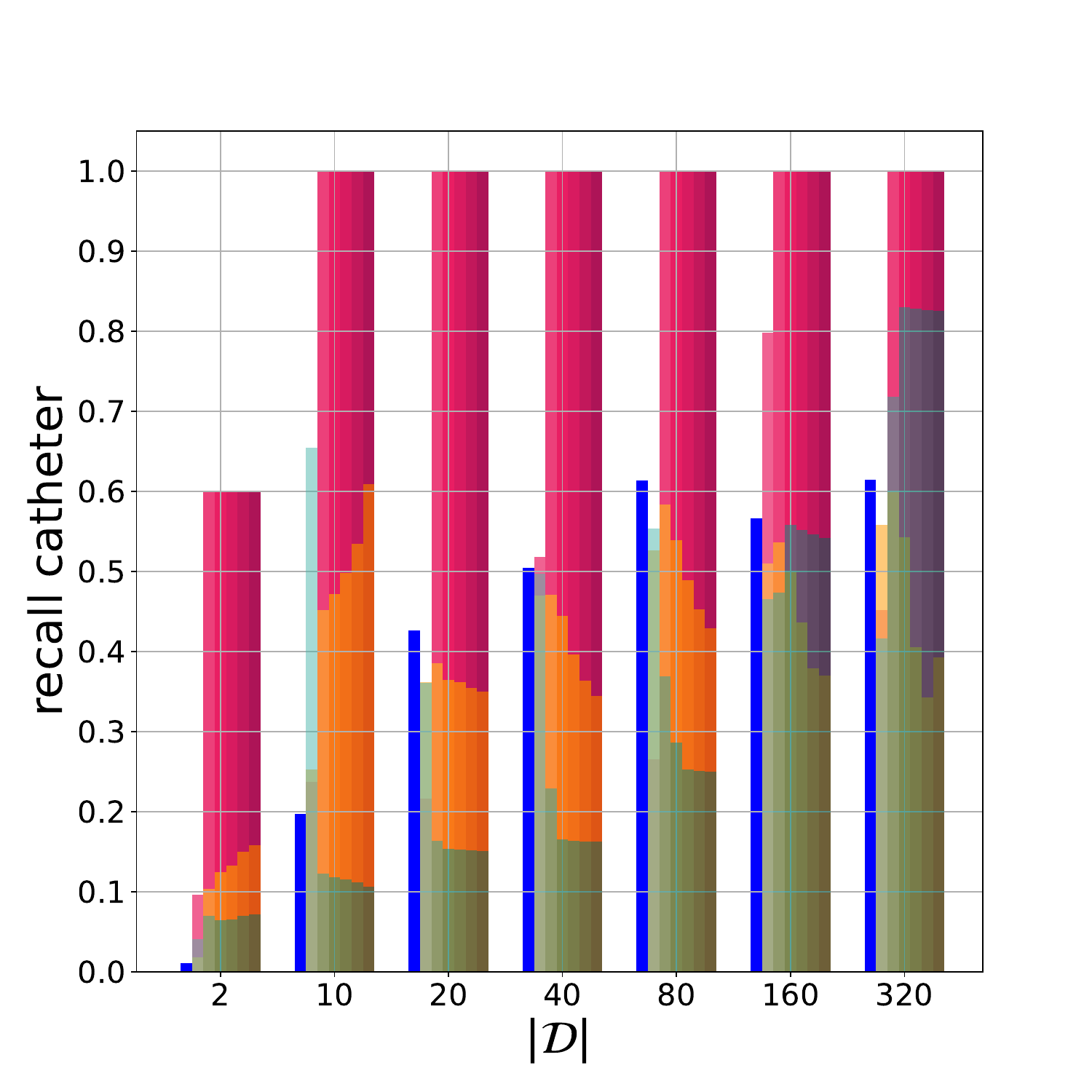}};
        \end{tikzpicture}
    \end{subfigure}
    \begin{subfigure}[b]{0.24\textwidth}
        \begin{tikzpicture}
            \node (figure) at (0,0) {\includegraphics[width=\textwidth]{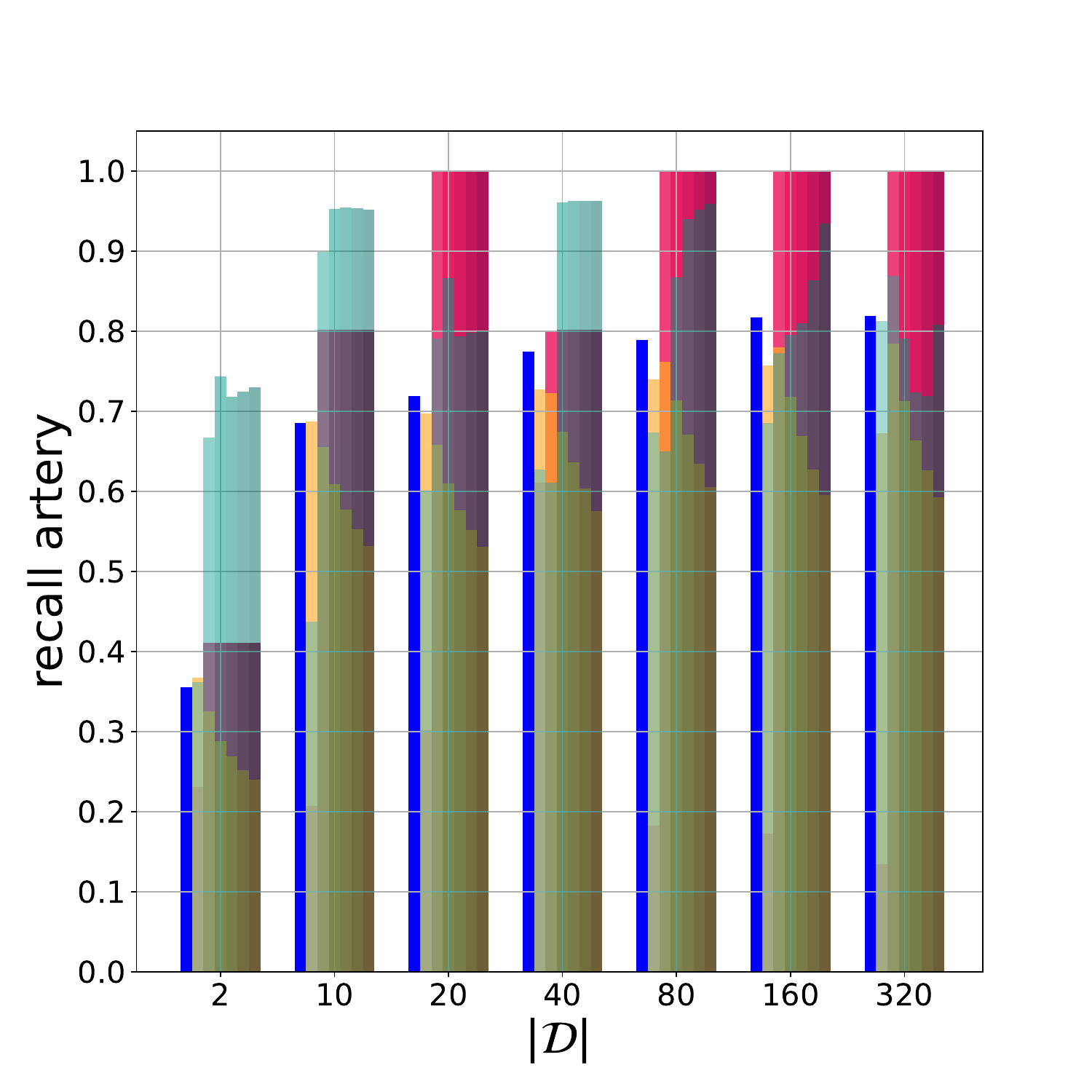}};
        \end{tikzpicture}
    \end{subfigure}
\end{figure}

%% file: main.bbl
\begin{thebibliography}{31}
\providecommand{\natexlab}[1]{#1}
\providecommand{\url}[1]{\texttt{#1}}
\expandafter\ifx\csname urlstyle\endcsname\relax
  \providecommand{\doi}[1]{doi: #1}\else
  \providecommand{\doi}{doi: \begingroup \urlstyle{rm}\Url}\fi

\bibitem[Ronneberger et~al.(2015)Ronneberger, Fischer, and Brox]{ronneberger2015u}
Olaf Ronneberger, Philipp Fischer, and Thomas Brox.
\newblock U-net: Convolutional networks for biomedical image segmentation.
\newblock In \emph{Medical image computing and computer-assisted intervention--MICCAI 2015: 18th international conference, Munich, Germany, October 5-9, 2015, proceedings, part III 18}, pages 234--241. Springer, 2015.

\bibitem[Ho et~al.(2020)Ho, Jain, and Abbeel]{ho2020denoising}
Jonathan Ho, Ajay Jain, and Pieter Abbeel.
\newblock Denoising diffusion probabilistic models.
\newblock \emph{Advances in neural information processing systems}, 33:\penalty0 6840--6851, 2020.

\bibitem[Zhou et~al.(2018)Zhou, Rahman~Siddiquee, Tajbakhsh, and Liang]{zhou2018unet}
Zongwei Zhou, Md~Mahfuzur Rahman~Siddiquee, Nima Tajbakhsh, and Jianming Liang.
\newblock Unet++: A nested {U}-net architecture for medical image segmentation.
\newblock In \emph{Deep Learning in Medical Image Analysis and Multimodal Learning for Clinical Decision Support: 4th International Workshop, DLMIA 2018, ML-CDS 2018, Granada, Spain, September 20, 2018, Proceedings 4}, pages 3--11. Springer, 2018.

\bibitem[Serre(2014)]{serre2014hierarchical}
Thomas Serre.
\newblock Hierarchical models of the visual system.
\newblock \emph{Encyclopedia of computational neuroscience}, 6:\penalty0 1--12, 2014.

\bibitem[Spoerer et~al.(2020)Spoerer, Kietzmann, Mehrer, Charest, and Kriegeskorte]{spoerer2020recurrent}
Courtney~J Spoerer, Tim~C Kietzmann, Johannes Mehrer, Ian Charest, and Nikolaus Kriegeskorte.
\newblock Recurrent neural networks can explain flexible trading of speed and accuracy in biological vision.
\newblock \emph{PLoS computational biology}, 16\penalty0 (10):\penalty0 e1008215, 2020.

\bibitem[Flavell et~al.(2022)Flavell, Gogolla, Lovett-Barron, and Zelikowsky]{flavell2022emergence}
Steven~W Flavell, Nadine Gogolla, Matthew Lovett-Barron, and Moriel Zelikowsky.
\newblock The emergence and influence of internal states.
\newblock \emph{Neuron}, 110\penalty0 (16):\penalty0 2545--2570, 2022.

\bibitem[Rao and Ballard(1999)]{rao1999predictive}
Rajesh~PN Rao and Dana~H Ballard.
\newblock Predictive coding in the visual cortex: a functional interpretation of some extra-classical receptive-field effects.
\newblock \emph{Nature neuroscience}, 2\penalty0 (1):\penalty0 79--87, 1999.

\bibitem[Louren{\c{c}}o-Silva et~al.(2021)Louren{\c{c}}o-Silva, Menezes, Rodrigues, Silva, Pinto, and Oliveira]{lourencco2021encoder}
Jo{\~a}o Louren{\c{c}}o-Silva, Miguel~Nobre Menezes, Tiago Rodrigues, Beatriz Silva, Fausto~J Pinto, and Arlindo~L Oliveira.
\newblock Encoder-decoder architectures for clinically relevant coronary artery segmentation.
\newblock In \emph{International Conference on Computational Advances in Bio and Medical Sciences}, pages 63--78. Springer, 2021.

\bibitem[Tan and Le(2019)]{tan2019efficientnet}
Mingxing Tan and Quoc Le.
\newblock Efficientnet: Rethinking model scaling for convolutional neural networks.
\newblock In \emph{International conference on machine learning}, pages 6105--6114. PMLR, 2019.

\bibitem[Kohonen(1982)]{kohonen1982self}
Teuvo Kohonen.
\newblock Self-organized formation of topologically correct feature maps.
\newblock \emph{Biological cybernetics}, 43\penalty0 (1):\penalty0 59--69, 1982.

\bibitem[Zheng et~al.(2015)Zheng, Jayasumana, Romera-Paredes, Vineet, Su, Du, Huang, and Torr]{zheng2015conditional}
Shuai Zheng, Sadeep Jayasumana, Bernardino Romera-Paredes, Vibhav Vineet, Zhizhong Su, Dalong Du, Chang Huang, and Philip~HS Torr.
\newblock Conditional random fields as recurrent neural networks.
\newblock In \emph{Proceedings of the IEEE international conference on computer vision}, pages 1529--1537, 2015.

\bibitem[Hoover et~al.(2024)Hoover, Liang, Pham, Panda, Strobelt, Chau, Zaki, and Krotov]{hoover2024energy}
Benjamin Hoover, Yuchen Liang, Bao Pham, Rameswar Panda, Hendrik Strobelt, Duen~Horng Chau, Mohammed Zaki, and Dmitry Krotov.
\newblock Energy transformer.
\newblock \emph{Advances in Neural Information Processing Systems}, 36, 2024.

\bibitem[Von~der Malsburg(1973)]{von1973self}
Chr Von~der Malsburg.
\newblock Self-organization of orientation sensitive cells in the striate cortex.
\newblock \emph{Kybernetik}, 14\penalty0 (2):\penalty0 85--100, 1973.

\bibitem[Frangi et~al.(1998)Frangi, Niessen, Vincken, and Viergever]{frangi1998multiscale}
Alejandro~F Frangi, Wiro~J Niessen, Koen~L Vincken, and Max~A Viergever.
\newblock Multiscale vessel enhancement filtering.
\newblock In \emph{Medical Image Computing and Computer-Assisted Intervention—MICCAI’98, Cambridge, MA, USA, October 11--13, 1998 Proceedings 1}, pages 130--137. Springer, 1998.

\bibitem[Li et~al.(2024)Li, Wen, Li, and Lee]{li2024emergence}
Tianqin Li, Ziqi Wen, Yangfan Li, and Tai~Sing Lee.
\newblock Emergence of shape bias in convolutional neural networks through activation sparsity.
\newblock \emph{Advances in Neural Information Processing Systems}, 36, 2024.

\bibitem[Hopfield(1982)]{hopfield1982neural}
John~J Hopfield.
\newblock Neural networks and physical systems with emergent collective computational abilities.
\newblock \emph{Proceedings of the national academy of sciences}, 79\penalty0 (8):\penalty0 2554--2558, 1982.

\bibitem[Krotov and Hopfield(2016)]{krotov2016dense}
Dmitry Krotov and John~J Hopfield.
\newblock Dense associative memory for pattern recognition.
\newblock \emph{Advances in neural information processing systems}, 29, 2016.

\bibitem[Martins and Astudillo(2016)]{martins2016softmax}
Andre Martins and Ramon Astudillo.
\newblock From softmax to sparsemax: A sparse model of attention and multi-label classification.
\newblock In \emph{International conference on machine learning}, pages 1614--1623. PMLR, 2016.

\bibitem[Martins et~al.(2023)Martins, Niculae, and McNamee]{martins2023sparse}
Andre Martins, Vlad Niculae, and Daniel~C McNamee.
\newblock Sparse modern {Hopfield} networks.
\newblock In \emph{Associative Memory \& Hopfield Networks in 2023}, 2023.

\bibitem[Kingma and Ba(2014)]{kingma2014adam}
Diederik~P Kingma and Jimmy Ba.
\newblock Adam: A method for stochastic optimization.
\newblock \emph{arXiv preprint arXiv:1412.6980}, 2014.

\bibitem[Kuck et~al.(2020)Kuck, Chakraborty, Tang, Luo, Song, Sabharwal, and Ermon]{kuck2020belief}
Jonathan Kuck, Shuvam Chakraborty, Hao Tang, Rachel Luo, Jiaming Song, Ashish Sabharwal, and Stefano Ermon.
\newblock Belief propagation neural networks.
\newblock \emph{Advances in Neural Information Processing Systems}, 33:\penalty0 667--678, 2020.

\bibitem[Satorras and Welling(2021)]{satorras2021neural}
Victor~Garcia Satorras and Max Welling.
\newblock Neural enhanced belief propagation on factor graphs.
\newblock In \emph{International Conference on Artificial Intelligence and Statistics}, pages 685--693. PMLR, 2021.

\bibitem[Kennedy and Eberhart(1995)]{kennedy1995particle}
James Kennedy and Russell Eberhart.
\newblock Particle swarm optimization.
\newblock In \emph{Proceedings of ICNN'95-international conference on neural networks}, volume~4, pages 1942--1948. ieee, 1995.

\bibitem[Ramsauer et~al.(2020)Ramsauer, Sch{\"a}fl, Lehner, Seidl, Widrich, Adler, Gruber, Holzleitner, Pavlovi{\'c}, Sandve, et~al.]{ramsauer2020hopfield}
Hubert Ramsauer, Bernhard Sch{\"a}fl, Johannes Lehner, Philipp Seidl, Michael Widrich, Thomas Adler, Lukas Gruber, Markus Holzleitner, Milena Pavlovi{\'c}, Geir~Kjetil Sandve, et~al.
\newblock Hopfield networks is all you need.
\newblock \emph{arXiv preprint arXiv:2008.02217}, 2020.

\bibitem[Grushin(2023)]{grushin2023training}
Alexander Grushin.
\newblock Training neural networks with internal state, unconstrained connectivity, and discrete activations.
\newblock \emph{arXiv preprint arXiv:2312.14359}, 2023.

\bibitem[Demircigil et~al.(2017)Demircigil, Heusel, L{\"o}we, Upgang, and Vermet]{demircigil2017model}
Mete Demircigil, Judith Heusel, Matthias L{\"o}we, Sven Upgang, and Franck Vermet.
\newblock On a model of associative memory with huge storage capacity.
\newblock \emph{Journal of Statistical Physics}, 168:\penalty0 288--299, 2017.

\bibitem[Vaswani et~al.(2017)Vaswani, Shazeer, Parmar, Uszkoreit, Jones, Gomez, Kaiser, and Polosukhin]{vaswani2017attention}
Ashish Vaswani, Noam Shazeer, Niki Parmar, Jakob Uszkoreit, Llion Jones, Aidan~N Gomez, {\L}ukasz Kaiser, and Illia Polosukhin.
\newblock Attention is all you need.
\newblock \emph{Advances in neural information processing systems}, 30, 2017.

\bibitem[Krotov(2021)]{krotov2021hierarchical}
Dmitry Krotov.
\newblock Hierarchical associative memory.
\newblock \emph{arXiv preprint arXiv:2107.06446}, 2021.

\bibitem[Elman(1990)]{elman1990finding}
Jeffrey~L Elman.
\newblock Finding structure in time.
\newblock \emph{Cognitive science}, 14\penalty0 (2):\penalty0 179--211, 1990.

\bibitem[Goetschalckx et~al.(2024)Goetschalckx, Govindarajan, Karkada~Ashok, Ahuja, Sheinberg, and Serre]{goetschalckx2024computing}
Lore Goetschalckx, Lakshmi~Narasimhan Govindarajan, Alekh Karkada~Ashok, Aarit Ahuja, David Sheinberg, and Thomas Serre.
\newblock Computing a human-like reaction time metric from stable recurrent vision models.
\newblock \emph{Advances in Neural Information Processing Systems}, 36, 2024.

\bibitem[Linsley et~al.(2020)Linsley, Ashok, Govindarajan, Liu, and Serre]{linsley2005stable}
D~Linsley, AK~Ashok, LN~Govindarajan, R~Liu, and T~Serre.
\newblock Stable and expressive recurrent vision models.
\newblock \emph{arXiv preprint arXiv:2005.11362}, 2020.

\end{thebibliography}
